\RequirePackage{fix-cm}
\documentclass[twocolumn,natbib]{svjour3}
\smartqed
\usepackage{url}
\usepackage{graphicx}
\usepackage[colorlinks=true,linkcolor=blue,citecolor=blue]{hyperref}
\usepackage{multirow}
\usepackage{amsmath,amssymb,amsfonts}
\usepackage{mathrsfs}
\usepackage[title]{appendix}
\usepackage[dvipsnames]{xcolor}
\usepackage{textcomp}
\usepackage{manyfoot}
\usepackage{booktabs}
\usepackage{algorithm}
\usepackage{algorithmicx}
\usepackage{algpseudocode}
\usepackage{listings}
\usepackage{comment}
\usepackage{pgffor}
\usepackage[T1]{fontenc}
\usepackage{microtype}
\usepackage{diagbox}
\usepackage{slashbox}
\usepackage{soul}
\usepackage{gensymb}

\pdfmapline{+futura < Futura.ttf <T1-WGL4.enc}

\newcommand\futurafont[1]{{\usefont{T1}{futura}{m}{n} #1 }}

\newlength{\dpcircle}
\newlength{\rcircle}
\newlength{\dcircle}
\newcommand{\docircle}[4]{%
  \setlength{\dpcircle}{\dp\strutbox}%
  \setlength{\dcircle}{\dpcircle}%
  \addtolength{\dcircle}{\ht\strutbox}%
  \setlength{\rcircle}{0.5\dcircle}%
  \setlength{\unitlength}{1sp}%
  \begin{picture}(\number\dcircle,0)
    \color{#1}
    \put(\number\rcircle,\number\dpcircle){\circle*{\number\dcircle}}
    \color{#2}
    \put(\number\rcircle,\number\dpcircle){\circle{\number\dcircle}}
    \put(\number\rcircle,0.08\dcircle){\makebox[0pt]{\hspace{1.8pt}\textcolor{#3}{\futurafont{#4}}}}
  \end{picture}%
}

\usepackage{xspace}
\newcommand{\egox}{EgoAI\xspace}

\definecolor{ourblue}{HTML}{3dc1c1}
\definecolor{ourcoral}{HTML}{c13d57}
\definecolor{darkred}{HTML}{F53A3A}

\DeclareRobustCommand{\storyref}[1]{%
  \foreach \num in {#1}{%
    \docircle{ourblue}{white}{white}{\fontsize{6.5}{0}\selectfont \textls[-100]{\num}}%
  }%
}

\newcommand\taskref[2][]{\setlength\fboxsep{1.8pt} \colorbox{ourcoral!70}{\textcolor{white}{\fontsize{6.5}{0}\selectfont\textbf{\usefont{T1}{futura}{m}{n}{#2}}}}}
\begin{document}

\title{An Outlook into the Future of Egocentric Vision}

\author{Chiara Plizzari$^\star$         \and
Gabriele Goletto$^\star$ \and
Antonino Furnari$^\star$ \and \\
Siddhant Bansal$^\star$ \and
Francesco Ragusa$^\star$ \and
Giovanni Maria Farinella$^\dagger$ \and\\
Dima Damen$^\dagger$ \and
Tatiana Tommasi$^\dagger$
}

\institute{$^\star$: Equal Contribution/First Author\\
$^\dagger$: Equal Senior Author\\
C. Plizzari, G. Goletto and T. Tommasi, Politecnico di Torino, Italy\ \and A. Furnari, F. Ragusa and G. M. Farinella, University of Catania, Italy \and S. Bansal and D. Damen, University of Bristol, UK. 
\email{Tatiana.Tommasi@polito.it}
}

\date{Received: date / Accepted: date}

\maketitle
\begin{abstract}
\textit{What will the future be? We wonder!}\\
In this survey, we explore the gap between current research in egocentric vision and the ever-anticipated future, where wearable computing, with outward facing cameras and digital overlays, is expected to be integrated in our every day lives.
To understand this gap, the article starts by envisaging the future through character-based stories, showcasing through examples the limitations of current technology.
We then provide a mapping between this future and previously defined research tasks.
For each task, we survey its seminal works, current state-of-the-art methodologies and available datasets, then reflect on shortcomings that limit its applicability to future research.
Note that this survey focuses on software models for egocentric vision, independent of any specific hardware.
The paper concludes with recommendations for areas of immediate explorations so as to unlock our path to the future always-on, personalised and life-enhancing egocentric vision.

\end{abstract}

\sloppy
\vspace{-1.75\baselineskip}
\setcounter{tocdepth}{2}
\tableofcontents

\section{Introduction}\label{sec1}
\vspace*{-8pt}
Designing and building tools able to support human activities, improve quality of life, and enhance individuals' abilities to achieve their goals is the ever-lasting aspiration of our species.
Among all inventions, digital computing has already had a revolutionary effect on human history.
Of particular note is mobile technology, currently integrated in our lives through hand-held devices, i.e. \textit{mobile smart phones}.
These are nowadays the de facto for outdoor navigation, capturing static and moving footage of our everyday and connecting us to both familiar and novel connections and experiences.

However, humans have been dreaming about the next-version of such mobile technology --- wearable computing, for a considerable amount of time.
Imaginations are present in movies, fictional novels and pop culture\footnote{Few examples: (1) Molly's Vision-Enhancing Lenses from the \textit{Neuromancer} novel, William Gibson, 1984. (2) JVC Personal Video Glasses from the \textit{Back to the Future II} movie, 1989. (3) Iron Man Suits with J.A.R.V.I.S. AI system from Marvel movies 2008-2015. (4) AI Earbuds and smartphone in shirt pocket from the \textit{Her} movie, 2013. (5) E.D.I.T.H. smart glasses from the \textit{Spider-Man: Far From Home} movie, 2019.}.
Notwithstanding the fast progress of Artificial Intelligence, and the hardware advances of the last ten years, our ability to fulfil this dream is lagging behind.

In computer vision, research papers on egocentric vision have instead limited their focus to a handful of applications, where current technology can already make a difference.
These are: training or monitoring in industrial settings, performing adhoc and infrequent tasks such as assembling a piece of furniture, preparing a new recipe, or playing a group game in a social setting.
These showcase egocentric wearables as niche devices very distant from everyone's everyday needs.
This perspective has not only limited our chances to convince others that egocentric vision is a key technology of our future, but it also restricted our ability to push the boundaries and remove obstacles to the integration of egocentric devices as the ultimate replacement of the \textit{mobile phone} with unlocking of additional capabilities.

To make a difference, we choose a future-to-present perspective in this paper, where we start from the envisaged future then analyse the fundamental tasks that are required. 
This approach allows us to take a more systemic and informed perspective, highlight the gap between the expected applications and the current technological status, and provide insights into promising future research directions.
While technology forecasting is not a very common approach to research review, \cite{firat2008technological} note its value in prioritising R\&D.
We take a scenario-based approach, amongst the options proposed in \cite{firat2008technological} for future forecasting.

Our work is related to previous surveys in egocentric vision. \cite{betancourt2015evolution} summarised the evolution of the state of the art in egocentric vision analysis from 1997 to 2014, the year of writing of the survey. 
\cite{nguyen2016recognition} reviewed algorithms for the recognition of activities of daily living from egocentric vision.
\cite{bolanos2017storytelling} surveyed approaches for visual storytelling from the analysis of egocentric photo-streams.
\cite{delmolino2017summarization} provided a survey of techniques for used to summarise egocentric video.
\cite{rodin2021predicting} analysed algorithms, datasets and tasks for action anticipation in egocentric vision.
\cite{nunez-marcos2022egocentric} summarised works in egocentric action recognition.
\cite{bandini2023analysis} considered works based on the analysis of hands in egocentric vision.

All previous survey papers, with the exception of \cite{betancourt2015evolution}, addressed specific topics in egocentric vision. In contrast, this paper offers a holistic overview. We also offer a comprehensive and updated view of the current status of egocentric vision, covering topics of localisation (Section \ref{sec:localisation}), scene understanding (Section \ref{sec:3dunderstanding}), recognition (Section \ref{sec:recognition}), anticipation (Section \ref{sec:anticipation}), gaze understanding and prediction~(Section \ref{sec:gaze}), social behaviour understanding (Section \ref{sec:social}), full-body pose estimation (Section \ref{sec:fullbody_egopose}), hand and hand-object interactions~(Section \ref{sec:hand-object}), person identification~(Section \ref{sec:personid}), summarisation~(Section \ref{sec:video_summarisation}), dialogue (Section \ref{sec:dialogue}), and privacy~(Section \ref{sec:privacy}). 

The remainder of this paper is organised as follows. In Section~\ref{sec:stories}, we present our vision of the future of egocentric vision through character-based stories and associated visuals.
Section~\ref{sec:stories_tasks} relates these stories to research tasks, structuring these into familiar research questions. 
In Section~\ref{sec:tasks}, we survey each task with subsections dedicated to seminal works, current state-of-the-art, dedicated datasets to these tasks and limitations to future applications.
In Section~\ref{sec:datasets}, we present general datasets frequently used in egocentric vision beyond a single task.
Finally, in Section~\ref{sec:conclusion} we conclude by providing a perspective to key questions that need to be unlocked soon for a step-change in egocentric vision.

\section{Imagining the Future}\label{sec:stories}
With the aim of performing an inspirational review of the current status of egocentric vision, we look into how research outputs are expected to impact our everyday life in the near future and investigate the gap still existing towards those results.  
The envisaged future takes the shape of 
\textit{five  distinctive use cases that are grounded in either a location or an occupation}.
In presenting each use case, we first summarise the existing relevant technology and then we introduce future narratives in the form of brief character-based stories, supporting the readers' imagination through artist-drawn sketches.
The protagonists of the plots use \emph{\egox, a wearable device that enables in-situ multimodal sensing from the wearer's perspective and provides ego-based assistance}. 
We associate story \storyref{p}arts with research tasks (marked by \taskref{section number}) and later revisit the link between these use-cases and research tasks in Section~\ref{sec:stories_tasks}.

\subsection{EGO-Home}
\label{sec:ego-home}
\textbf{Presently}, smart home technology encompasses a range of Internet of Things (IoT) devices. They either control specific domestic environmental variables (e.g. light, temperature, humidity, CO2 level, energy consumption) or manage the operation of electrical appliances on the basis of occupants' preferences. Surveillance cameras are increasingly being installed both indoors and outdoors to ensure safety and enable remote monitoring of pets, children, and the elderly. Furthermore, there has been a recent surge in the introduction of speakers assistive devices like Amazon Echo, Google Nest, and Apple HomePod, which mainly rely on audio input for interaction and event tracking. All these tools, though empowered by machine learning techniques, are dedicated to a few specific tasks and they are static in nature, covering only limited areas of the home.
\egox will replace the set of heterogeneous sensing tools currently in operation, but also provide much more. 

\vspace*{6pt}

\textit{Sam is finally at home after a hard-working day. A good dinner is certainly needed.
When Sam opens the fridge, \egox automatically analyses present stock in view and suggests a tomato soup as the tomatoes look perfectly ripe. Moreover, \egox has kept track of Sam's food intake for that day and the soup sounds like the best complementary nutrition. Sam does not enjoy cooking much, but \egox switches on the 3D projection of the Remy from the movie Ratatouille to help him through the soup prep. Remy jumps around his kitchen efficiently avoiding obstacles and appears to hold the knife while he chops the tomato encouraging Sam to slice his tomatoes thinner. Remy says, “this way the tomato will cook evenly”. The audio appears to come from the direction of the chopping board, where now Remy is comfortably sitting. Sam is continuously impressed by how fun it is to cook with his 3D projected friend. Sam is in doubt about the amount of spice he has added and whether more is recommended. \egox keeps track of ingredients, recommends more spice to be added and reminds Sam about the bread slice he’s nearly forgotten in the toaster.}

\begin{figure}[bp]
    \centering
    \includegraphics[width=\columnwidth]{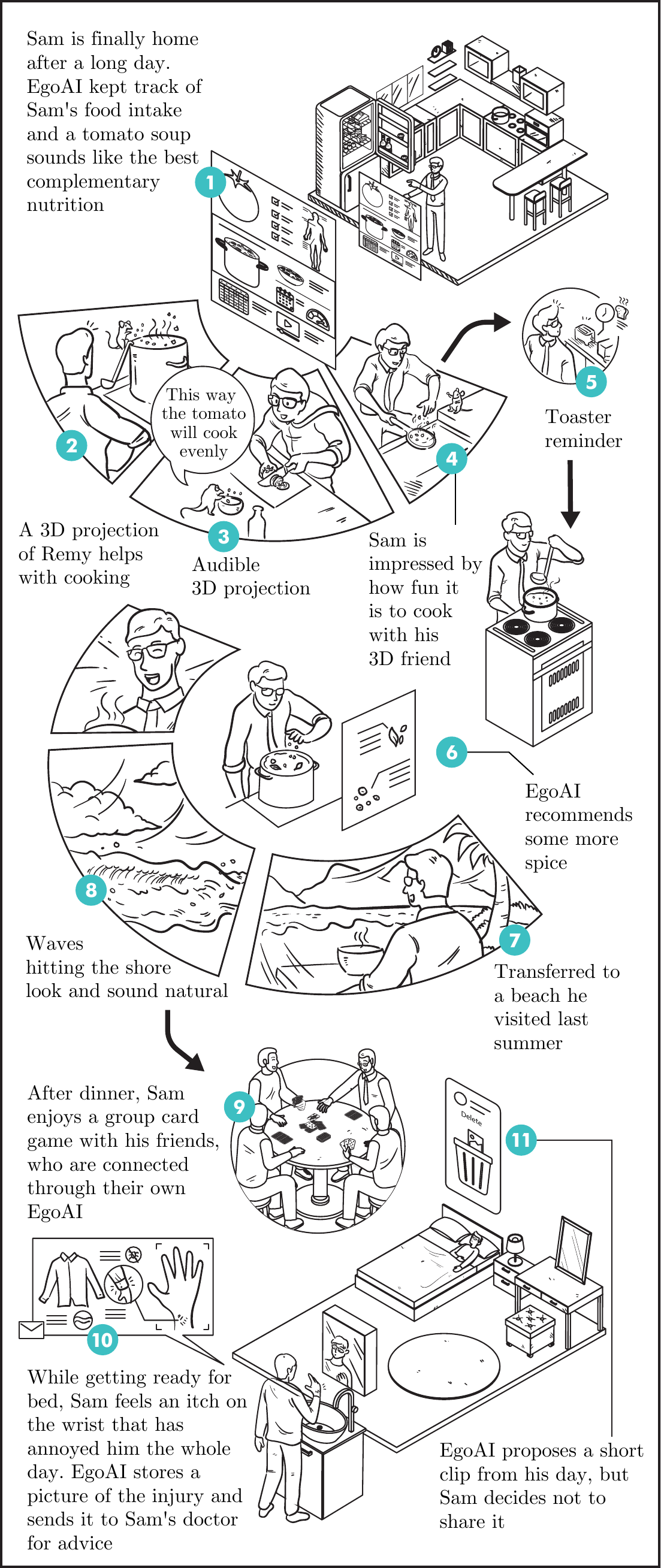}
    \caption{\textbf{EGO-Home}. Character-based story envisaging the future of egocentric vision at home. Illustration of the story from Section~\ref{sec:ego-home}. \egox{} assists Sam during dinner preparation and keeps him entertained with interactive and immersive experiences. \taskref{4.2} 3D Scene Understanding \storyref{1,2,3,4,7,8,9}.  \taskref{4.3}~Object and Action Recognition~\storyref{1,5,6,10}. Measuring System \storyref{6}. \taskref{4.11} Dialogue \storyref{6}. \taskref{4.10} Summarisation and Retrieval \storyref{7,11}. \taskref{4.7}~Full-body Pose, \taskref{4.8}~Hand Pose and \taskref{4.6}~Social Interaction \storyref{9}. Medical Imaging \storyref{10}. Messaging \storyref{10,11}.}
    \label{fig:home}
\end{figure}

\textit{While enjoying his warm soup, Sam asks \egox to take him back to that beach he visited last summer. Sam is virtually transformed to that same view he captured many months ago, and relaxes by listening to the waves hitting the shore while eating his hot soup. He laughs at the absurdity of the soup at the beach. Before heading to bed, Sam enjoys a group card game with his friends who recently moved to Australia. They are connected though their own \egox, which makes Sam feel as they are all physically present with him. He can hear the sound of the cards shuffling as his two friends appear seated around his table. \egox is a great game companion and points out a strategic move to make when he's about to play a suboptimal card.}

\textit{While getting ready for the night, Sam feels again that itch on the wrist that has annoyed him the whole day. \egox assures him that with high probability it is just the cuff of his new shirt that irritated the skin, but also offers to take care of this by sending a picture to his physician for advice. As Sam heads to bed, \egox proposes a short clip from his day that could be shared on social media, but Sam thinks ``not today'' and asks \egox to delete the post draft.}

\subsection{EGO-Worker}
\label{sec:ego-worker}
\begin{figure}
    \centering
    \includegraphics[width=\columnwidth]{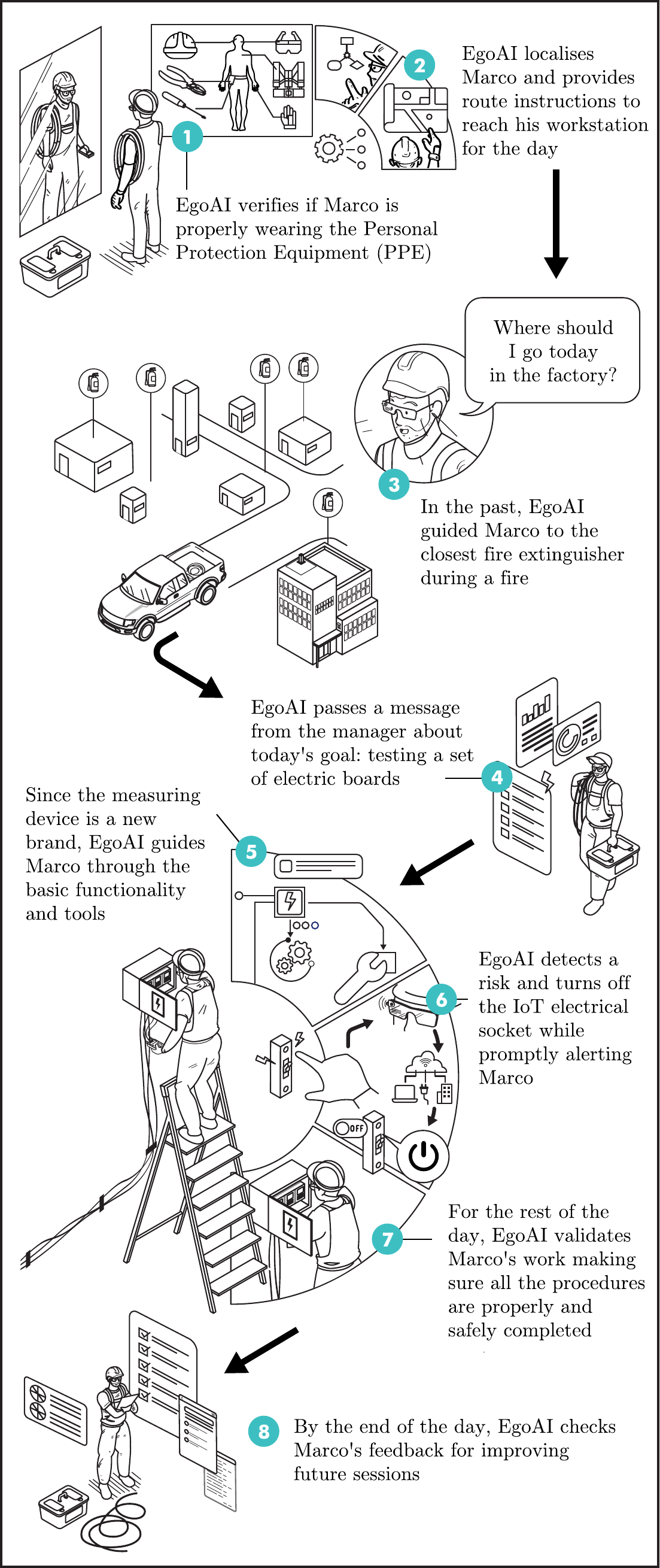}
    \caption{Character-based story envisaging the future of egocentric vision in industrial settings. Illustration of the story from Section~\ref{sec:ego-worker}.
    \egox assists Marco from the start of his day until its conclusion. Safety Compliance Assessment~\storyref{1}. \taskref{4.1}~Localisation and Navigation \storyref{2,5}. Messaging \storyref{4}. \taskref{4.8} Hand-Object Interaction \storyref{5}. \taskref{4.4} Action Anticipation \storyref{6}. Skill Assessment \storyref{7}.\taskref{4.11}~Visual Question Answering~\storyref{8}, \taskref{4.10}~Summarisation~\storyref{8}.}
    \label{fig:worker}
\end{figure}

\textbf{Current} vision-based systems are being integrated in large scale workshops and factories, but these mainly rely on fixed cameras, which need to be installed in all the areas of interest and can only perceive a limited view of the scene, hence restricting their usefulness.
Training and monitoring of workers is mostly offline through recorded material or over-the-shoulder advice from experienced workers. Often the knowledge is lost as one worker changes job.
Feedback to workers about their performance is based on heuristic automatic or manual calculations and often does not correspond to actual performance. 
This is often disconnected from training and advice for how to improve performance.
While technology is employed for workers' safety increasingly, this is below the levels expected with most technological advances focusing on improved productivity.
\egox will fill this gap and make the lives of workers safer and more comfortable.

\vspace*{6pt}
\emph{As with every morning, Marco begins his shift by looking at himself in the mirror: in this way \egox can verify if he is properly wearing the Personal Protection Equipment (PPE) which will guarantee his safety. After this check, Marco asks \egox where in the factory he is needed today. \egox  localises Marco and provides route instructions to reach his workstation for the day, avoiding dangerous areas with suspended loads and the paths reserved for the transit of vehicles. Marco trusts \egox navigation abilities and always remembers that day when \egox swiftly guided him to the closest fire extinguisher to avoid flames spreading.} 

\emph{As Marco reaches his workstation, \egox passes a message from the manager about today's goal: testing a set of electrical boards. Since the hand-held measuring tool is a new brand, \egox guides Marco through the basic functionality useful to correctly test the boards. Unfortunately, Marco gets distracted and is about to probe the electrical board while it is still plugged in. \egox detects the risk and turns off the IoT electrical socket on which the board is connected while promptly alerting Marco.}

\emph{For the rest of the day, \egox validates Marco's work making sure that all the procedures are properly and safely completed, answering his questions in case of doubts, and estimating his stress level to make sure that he takes breaks when needed.}

\emph{By the end of the day, \egox thanks Marco for his very hard work particularly with all new procedures involved and checks on his feedback for better training. \egox sends this feedback on any misunderstandings and obstacles automatically to future training sessions and planning.}

\begin{figure}
    \centering
    \includegraphics[width=\columnwidth]{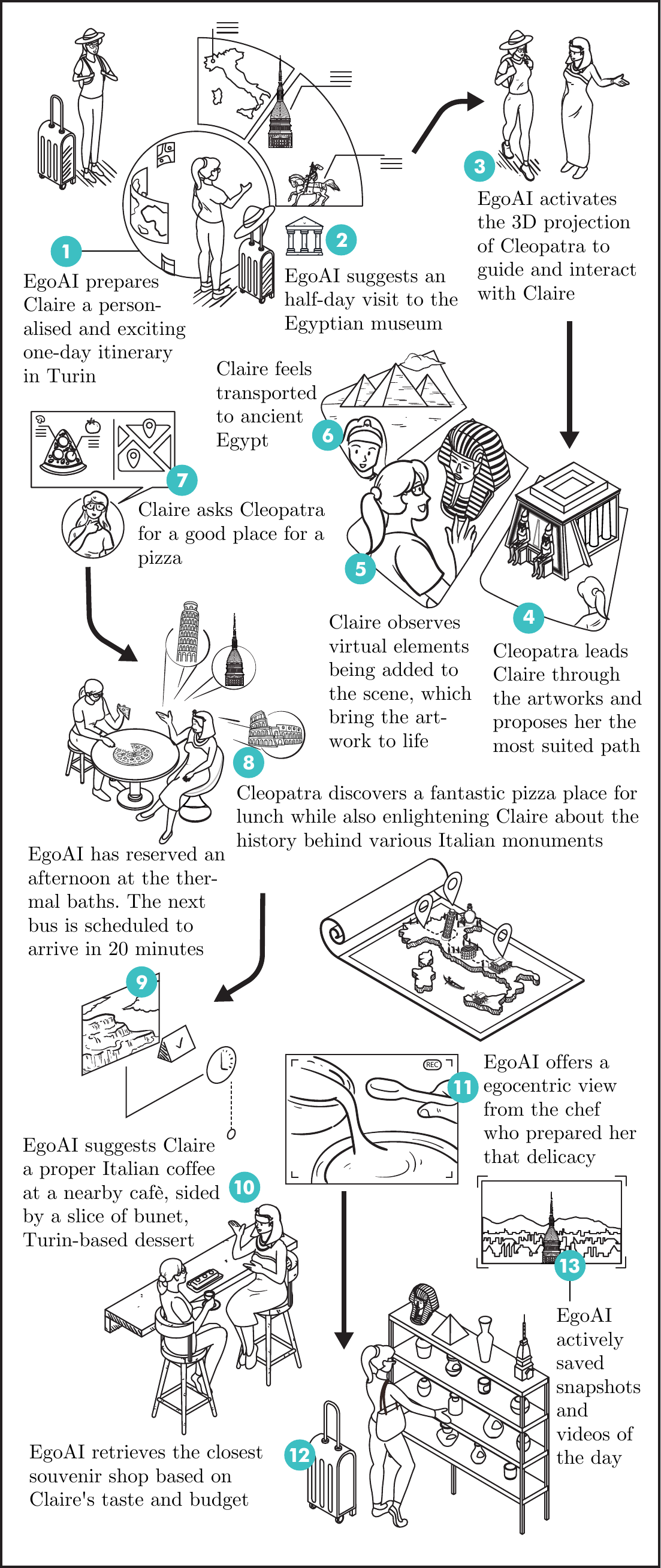}
    \caption{Character-based story envisaging the future of egocentric vision in tourism. Illustration of the story from Section~\ref{sec:ego-tourist}.
    \egox accompanies Claire throughout her itinerary in Turin. Recommendation and Personalisation \storyref{1,2,8,9,10,11}. \taskref{4.2}~3D Scene Understanding \storyref{2,3,4,5,6}. \taskref{4.5} Gaze Prediction~\storyref{5}. \taskref{4.1} Localisation and Navigation \storyref{3,4,8,12}. Messaging \storyref{7}. \taskref{4.11} Dialogue \storyref{8}. \taskref{4.3} Action Recognition and Retrieval \storyref{11}. \taskref{4.10} Summarisation \storyref{13}.}
    \label{fig:tourist}
\end{figure}
\subsection{EGO-Tourist}
\label{sec:ego-tourist}
\textbf{Today,} travelling abroad for tourism and vacations has more than doubled in the past 20 years\footnote{\url{https://ourworldindata.org/tourism}}.
Technology and art, both ancient and modern, are becoming increasingly bounded, with the former increasing the spread and the possibility of interaction with the latter. Indeed, the use of technological tools such as digital audio guides or virtual tours is becoming predominant in museums and touristic sites with engagement being crucial to increasing the visitor's interest. Despite modern tools, the visitor experience still lacks a form of personalisation and necessitates active interaction from the user. 
\egox on the other hand, fills these gaps and makes travelling a fun and interactive experience.

\vspace*{6pt}

\emph{Claire has just reached Turin as the last stop of her Italian holidays. She is thrilled to start her visit but does not know much about the city. Luckily, \egox is already tuned on Claire's tastes and prepared her a personalised and exciting one-day itinerary. \egox knows Claire is a big fan of museums so suggests half a day to visit the famous local Egyptian one. During the visit, \egox activates the 3D projection of Cleopatra to guide and interact with Claire. Cleopatra leads Claire through the artworks and proposes her the most suited path. 
While Claire is asking Cleopatra information about a sarcophagus, she observes virtual elements being added to the scene which bring the artwork to life. Claire feels transported to ancient Egypt where she can manipulate and use the pieces as it was intended.}

\emph{At the end of the visit, Claire decides to keep Cleopatra as her AR guide for lunch and asks her for a good pizza place. While enjoying her meal, Claire asks Cleopatra questions about famous Italian monuments she visited along her tour, augmenting her understand of the history behind them.}

\emph{\egox has booked an afternoon at the thermal baths. As the next bus is not due for another 20 minutes, \egox suggests Claire a proper Italian coffee at a nearby cafè sided by a slice of bunet, a popular Turin-based dessert. Claire would like to learn the recipe, so \egox offers a first-person view from the chef who prepared that delicacy earlier in the day.} 

\emph{After the thermal baths, \egox checks whether Claire is interested in buying some souvenirs for her family. 
\egox then retrieves the closest souvenir shop based on each relative's taste and the budget set by Claire.}

\emph{Claire was engaged during her one-day itinerary and did not worry about taking pictures. \egox actively saved relevant snapshots of the day, and videos of her favourite moments.}

\subsection{EGO-Police}
\label{sec:ego-police}
\textbf{In 2023}, it has been almost ten years since the adoption of body-worn cameras by several police departments around the world. Practical experience showed that they have a large potential in enhancing transparency and facilitating investigations, besides increasing officers' accountability and safety. Still, cameras provide only passive support to law enforcement, with data storage and post-processing analysis requiring a consistent time and cost effort. We can easily imagine how constables would benefit from AI-empowered wearable vision devices. 

\begin{figure}
    \centering
    \includegraphics[width=\columnwidth]{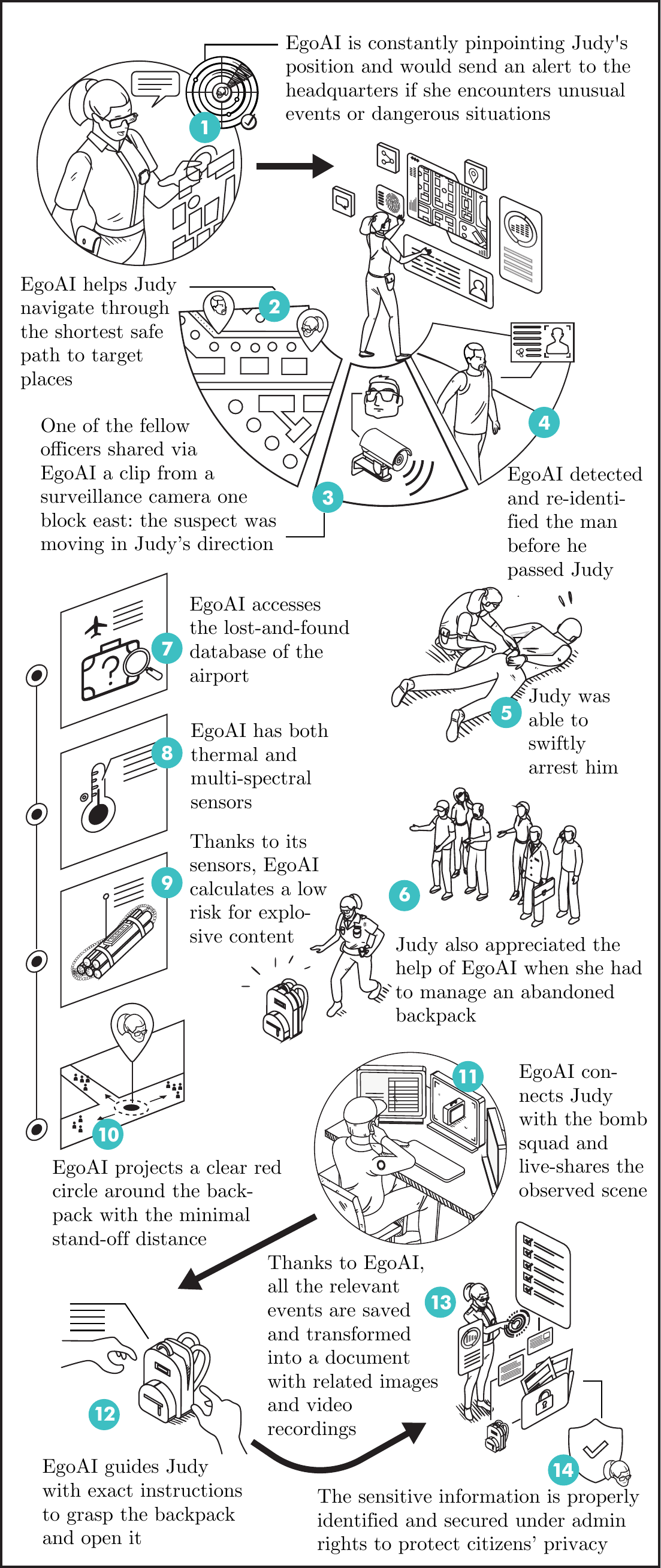}
    \caption{Character-based story envisaging the future of egocentric vision within the police force. Illustration of the story from Section~\ref{sec:ego-police}. \egox helps Judy, a police officer, during her day keeping her city safe. \taskref{4.1} Localisation and Navigation~\storyref{1,2}. Messaging \storyref{1,3,11}. \taskref{4.3} Action Recognition \storyref{2,13}. \taskref{4.9} Person Re-ID \storyref{2,4}. \taskref{4.3} Object Detection and Retrieval~\storyref{7}. Measuring System \storyref{8,9}. Decision Making \storyref{9}. \taskref{4.2}~3D Scene Understanding \storyref{10}. \taskref{4.8}~Hand-Object Interaction \storyref{12}. \taskref{4.10}~Summarisation \storyref{13}. \taskref{4.12} Privacy \storyref{14}. }
    \label{fig:police}
\end{figure}

\vspace*{6pt}
\emph{Judy is a police officer who uses \egox every day of her service.
She finds it highly convenient: the device is much lighter than the usual equipment and serves as body camera, radio, phone and flashlight. Moreover, it makes her feel safe as she knows that \egox is constantly pinpointing her position and would send an alert to headquarters if she encounters unusual events or dangerous situations.}
\emph{For instance, last month Judy was assigned to a high-crime zone while searching for a suspect. \egox helped Judy navigate through the shortest safe path to several target places reported as possible hideouts. While patrolling the streets, one of the fellow officers shared via \egox a clip from a surveillance camera one block east: the suspect was moving in Judy's direction. Despite the crowds, \egox detected and re-identified the man before he passed Judy. She was able to swiftly arrest him.}

\emph{Judy also appreciated the help of \egox when she had to manage an abandoned backpack at the airport. \egox accessed the lost-and-found database of the airport but no match was found. Then, from thermal and multi-spectral sensors, it calculated a low risk for explosive content and projected a clear red circle around the backpack with the minimal stand-off distance. \egox connected Judy with the bomb squad and live-shared the observed scene: the experts agreed with the initial evaluation and excluded any risk that the backpack could contain an explosive. Then, \egox 
guided Judy with exact instructions to grasp the backpack and open it. Luckily it was only containing a pair of old tennis shoes.}

\emph{At the end of every working day, Judy does not need to fill out any form or detailed reports. Thanks to \egox, relevant events are saved and transformed into a document with related images and video recordings. Importantly, the sensitive information possibly captured by \egox during Judy's work is properly identified and secured under admin rights to protect citizens’ privacy.}

\begin{figure}
    \centering
    \includegraphics[width=\columnwidth]{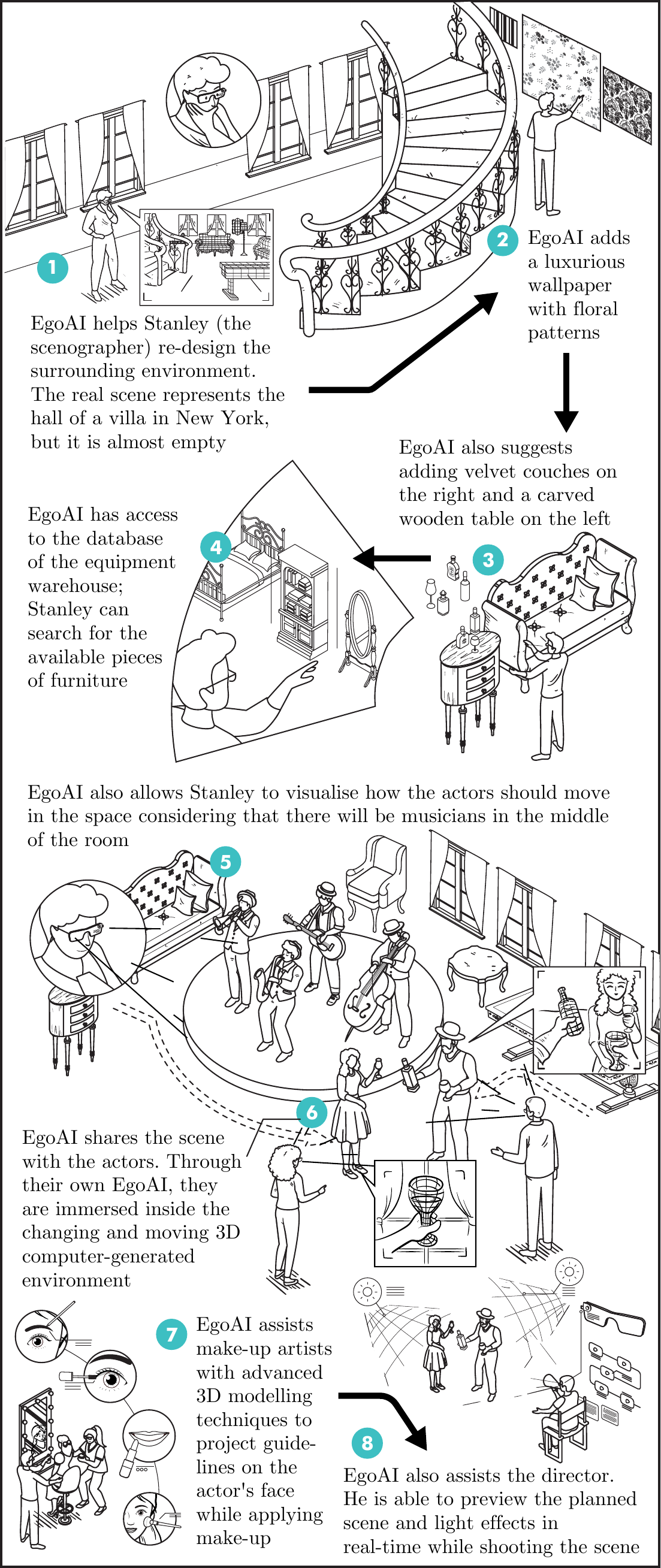}
    \caption{Character-based story envisaging the future of egocentric vision in the entertainment industry, focusing on the perspective of scene and makeup designers. Illustration of the story from Section~\ref{sec:ego-designer}. \egox helps Stanley, the scenographer, and all the crew during movie production. \taskref{4.2}~3D Scene Understanding \storyref{1,2,3,4,5,6,7,8}. Recommendation~\storyref{3}. \taskref{4.3}~Object Recognition and Retrieval \storyref{3,4}. \taskref{4.7} Full-body Pose Estimation \storyref{5,6}. \taskref{4.6} Social Interaction~\storyref{6}. \taskref{4.5} Gaze Prediction \storyref{6}. \taskref{4.8} Hand-Object Interaction \storyref{7}. Messaging \storyref{6, 8}.}
    \label{fig:artist}
\end{figure}

\subsection{EGO-Designer}
\label{sec:ego-designer}
\textbf{Nowadays,} films are full of digital artefacts, not only sci-fi ones but also realistic dramas. These include fantasy environments as well as scenes that cannot simply be shot on-site. Current technology makes use of neon-green screens which are then removed using video editing software in post-production. However, this makes it difficult for the scenographer to visualise the final effects while shooting, and the actors must perform without having a full perception of the 3D scene around them. A movie production crew may largely benefit from egocentric devices for augmented reality (AR), digital rendering, and 3D modelling leading to a completely innovative way to experience the movie creation processes.

\vspace*{6pt}
\textit{It is another hot day in Hollywood and Stanley has promised the movie director that the scenography will be ready first thing tomorrow. He is at the studios wearing \egox which is augmenting the surrounding environment: the real scene he is looking at is the reconstructed hall of a villa in New York during the 1920s. There is a fancy large spiral staircase but besides that, it is almost empty and should be designed to host a glamorous party.}

\textit{\egox helps Stanley to virtually add a luxurious wallpaper with floral patterns and a ceiling adorned with intricate moldings. He adjusts the position of two digital chandeliers so that they cast a warm, golden glow across the room. \egox also suggests adding velvet couches on the right and a carved wooden table on the left with crystal decanters, champagne flutes, and a variety of liquor bottles. As \egox has access to the database of the equipment warehouse, Stanley can search for the available pieces of furniture which are most similar to what he has in mind so that the production assistants can position them in the scene. \egox also allows Stanley to visualise how the actors should move in the space around the musicians in the middle of the hall.}

\textit{The scene is promptly shared with the actors. Through their \egox, actors are immersed inside the changing and moving 3D computer-generated environment so that they can visually engage with elements present in front of them and rehearse. Their natural acting is enhanced by their familiarity with the scene before shooting starts.}

\textit{Stanley has also some suggestions for the make-up artists about the colour palettes that would stand out with the chosen lights. It will be very easy to share information with them as they are also using \egox with advanced 3D modelling techniques to project guidelines on the actor's face while applying the make-up.}

\textit{At the end of the day Stanley feels satisfied and he is sure that his work will be appreciated: through  \egox the director will be able to preview the planned scene and light effects in real-time while shooting the scene, without having to wait for playback. \egox has saved the industry millions of dollars, with repetitions of scenes dropping to one fourth compared to movies captured in the ancient era before \egox was introduced.}

\section{From Narratives to Research Tasks}
\label{sec:stories_tasks}
Various research tasks can be identified in the above character-based narratives/stories. 
While some are only part of the future (e.g. particularly those related to augmented reality (AR)), others are currently achieved either via remote cameras (e.g. person identification) and Internet-of-Things devices (e.g. scene monitoring), or via smartphones (e.g. navigation). Despite their connectivity, local devices are typically restricted in coverage depending on where they are originally installed, while smartphones inevitably hinder interaction with the environment as they involve manual handling.
It is our vision that most of the mentioned tasks will be seamlessly integrated into one egocentric device that we refer to as \egox in our stories. It will be person-centric, thus wearer-focused, and will also travel \textit{anywhere} with the wearer.

In this section, we provide a mapping from the narratives above to research tasks as currently understood by the research community. We also examine whether these tasks can be performed using existing wearable devices or if new, more advanced and powerful ones are required to overcome the limitations of those currently available on the market.
This sets the scene for the literature survey of the research tasks in Section~\ref{sec:tasks}.

For any task that involves \emph{AR} technology, the need arises for in-depth \taskref{4.2} \emph{3D scene understanding}. This is exemplified by the EGO-Home's augmented reality guides for cooking, EGO-Tourists' immersive museum experiences, and EGO-Designer's creation of imaginary scenes. 
Our envisaged AR is also endowed with directional audio synthesis, where auditory feedback enhances the realism of the augmented environment, as in the case of the mouse or the sounds of cards being shuffled in EGO-Home.
To move within the 3D scene, \taskref{4.1} \emph{localisation} and \emph{navigation} emerge as recurrent tasks, both in the case of constrained spaces such as the factory in EGO-Worker as well as in open areas, as evident in EGO-Police's use of city maps.
The abilities of current egocentric devices to perceive 3D scenes is continuously evolving due to the integration of additional environment cameras (e.g., Microsoft HoloLens 2\footnote{\url{https://www.microsoft.com/en-us/hololens}}, Xreal Light\footnote{\url{https://www.xreal.com/light/}}, Magic Leap 2\footnote{\url{https://www.magicleap.com/magic-leap-2}}, Project Aria Glasses\footnote{\url{https://about.meta.com/realitylabs/projectaria/}}). These devices can scan and create a 3D model of the static environment to localise the wearer and allow them to navigate more easily.
Dynamic as well as outdoor scenes still challenge these setups and this remains an active area of research for a realistic integration of 3D understanding in the future.

Inside the scene, high-level understanding of actions is carried out. 
Tasks like \taskref{4.3} \emph{action recognition} undergo a paradigm shift 
with a transition in perspective from third-person to first-person view. 
In EGO-Worker, the device validates the user's actions in a workplace setting. Particularly noteworthy is \taskref{4.4} \textit{action anticipation}, where the device can promptly prevent dangerous situations. 
There are currently no smart glasses on the market that are able to robustly recognise human actions in real time. Usually, data from the RGB camera and depth sensor of the glasses are collected and processed offline due to hardware limitations.

\begin{figure*}[t]
    \centering
    \includegraphics[width=\textwidth]{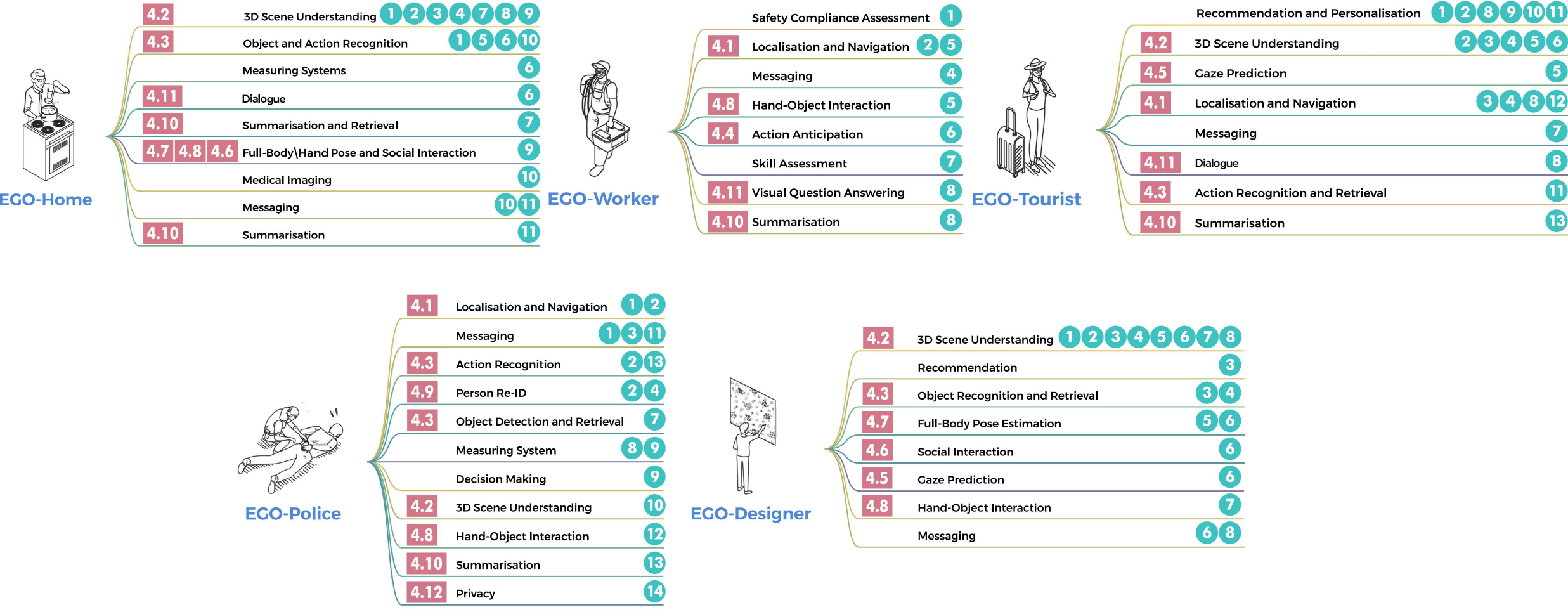}
    \vspace{-10pt}
    \caption{Illustration of the connections between our narratives and the research tasks. For each of the use cases presented in Section \ref{sec:stories}, we show the corresponding research tasks, along with the specific part of the story where the tasks are occurring, indicated by the numbers corresponding to those representing sub-stories in Figures \ref{fig:home}, \ref{fig:worker}, \ref{fig:tourist}, \ref{fig:police}, and \ref{fig:artist}, respectively. }
    \label{fig:stories_task}
\end{figure*}

Equipped with \taskref{4.5} \emph{gaze prediction}, \egox{} can track the user's eye movements and attend to objects seamlessly with their gaze. This capability is noted in both EGO-Tourist, where the user manipulates the artworks in the museum, and EGO-Designer, where the user virtually re-positions the objects in the virtual set. Nowadays, gaze tracking is a relatively stable feature but requires an eye calibration step before use and there remains the potential for drift over time. It has been implemented in wearable devices such as Microsoft Hololens2, Magic Leap 2, Project Aria Glasses and Apple Vision Pro\footnote{\url{https://www.apple.com/apple-vision-pro/}}.

Analysing the social context of the camera wearer through \taskref{4.6} \emph{social behaviour understanding} is also of significant importance. 
Social interactions are explored in EGO-Home, where users engage in interactive games with others connected through their devices. By employing \taskref{4.7} \emph{body pose estimation} techniques from a first-person perspective, each user's pose is accurately reconstructed and seamlessly integrated into AR.
Hands in particular are actively interacting with the environment and other individuals.
\taskref{4.3} \emph{action and object recognition},  \taskref{4.8} \emph{hand-pose estimation} and \emph{hand-object interactions} are key to \egox{}.
In EGO-Police, the device intelligently comprehends user actions, providing precise instructions, like how to open the suspicious backpack. In EGO-Worker, \egox helps to operate a new measuring tool.  

Recognising the user's identity and those of the bystanders plays a crucial role in social relationships. It has also a relevant role in security, going beyond what can be done with fixed cameras as in the case of \taskref{4.9}~\emph{person re-identification} described in the EGO-Police scenario. 
Of course, the identity, as well as users' data, should be properly safeguarded to ensure responsible use of the technology. As the wearable device can be in an ``always-on'' mode, it becomes imperative to address \taskref{4.12} \emph{privacy} concerns and establish robust protection measures. Different laws regulate data protection and privacy in different countries, such as the General Data Protection Regulation (GDPR) in Europe, the California Consumer Privacy Act (CCPA), and the China Cyber Security Law (CCSL). However, the wearable glasses that are currently available in the market do not have strategies for compliance and it's left to the user to regulate the device by interactive privacy switches. 

A related further issue is how to manage the ongoing and abundant stream of captured data that would be extremely costly to store in raw form. 
An efficient \taskref{4.10}~\emph{summarisation} and reporting process is clearly needed in multiple application scenarios. 
In both EGO-Home and EGO-Police, all the relevant events are saved and transformed into a report with images and video recordings. Identifying interesting events to memorise is also noted in EGO-Tourist. Thanks to \egox, it is also possible to \emph{retrieve} relevant data or objects within a database by exploiting visual cues in both EGO-Police and EGO-Designer. 
In EGO-Worker, \egox{} can conduct \emph{skill assessments} by monitoring whether the user correctly executes all the required procedures during their workday. Support in skill training is provided by the \taskref{4.11} \emph{Visual Question Answering} application (VQA) that replies to Marco's questions by translating instructional videos into a step-by-step guide in the users' view. 

By analysing the user's past data, the egocentric device can extract their preferences and offer \emph{personal recommendations}. For instance, in EGO-Home, \egox proposes dinner recipes according to the user's preferences and eating history. Similarly, in EGO-Tourist, it suggests fitting lunch and shopping destinations based on the individual's taste.

The ability to solve several other side tasks will contribute to the success of the \egox device that we foresee. 
\emph{Messaging} capability is recurrent throughout the stories. In EGO-Home, the user can send a picture of his wrist to the doctor, in EGO-Worker, the user received a message from the manager about his daily tasks, and in EGO-Police, \egox is capable of sending alerts to headquarters.  
This hands-free convenience is further enhanced by voice commands, allowing seamless interaction, as in EGO-Tourist when the tourist asks for additional information on the artwork. Some wearable glasses integrate vocal assistants such as Cortana\footnote{\url{https://www.microsoft.com/en-us/cortana}}, Siri\footnote{\url{https://www.apple.com/siri/}}, or Google Assistant\footnote{\url{https://assistant.google.com/}}, which can interact with the device to open applications, take photos, send messages, and more, clearly improving the user experience. 

\egox also functions as a \emph{measuring} instrument. In EGO-Home, \egox{} can quantify the amount of spice in the soup by leveraging its memory of the quantity previously added to the pan, or measuring the thickness of the soup from its visual appearance. Thanks to the possibility of integrating multiple sensors, such as thermal and multi-spectral cameras, it can also compute the risk of explosive content in EGO-Police by a \emph{decision making} process.
Wearable devices can also be integrated with advanced \emph{medical imaging} techniques, enabling \egox to assess the severity of the condition from a picture in EGO-Home. Another assessment expected of \egox{} is related to \emph{Safety Compliance Verification}. In EGO-Worker, \egox{} assesses whether the worker is correctly wearing a Personal Protection Equipment (PPE) through sophisticated recognition and identification techniques. 

Currently, there are no devices in the market that can match the advanced features and capabilities of \egox. They also have strong hardware limitations that do not allow prolonged use.
Even though some advanced wearable glasses provide complex and highly accurate features, the battery life is often only a few hours. This is even shorter when videos are captured constantly, as required to enhance the potential of summarisation techniques which also require a lot of computational power and memory, which can quickly drain the battery.

In this paper, we do not consider AR-specific approaches, which have their base in the computer graphics literature, but point the interested readers to recent surveys on the topic \citep{devagiri2022augmented,dargan2023augmented} as well as a structured literature survey by~\cite{cipresso2018past} and a survey of AR usability studies by ~\cite{dey2018systematic}.
We also exclude tasks that require perception or synthesis of audio, independent of the video -- this includes speech and audio-only event perception. 
We are not aware of a recent survey on the topic and encourage researchers with relevant expertise to further explore this crucial modality. 
Moreover, we do not review system-based tasks such as personalised recommendations or measuring devices, as well as tasks related to assessment, whether for medical purposes or skill. 
We refer the reader to works on action assessment~\citep{doughty2019pros,Parmar_2019_CVPR,Li_2019_ICCV,Yu_2021_ICCV} and risk warning~\citep{de2022attention}.

Instead, in Section~\ref{sec:tasks}, we focus on a subset of all the aforementioned tasks -- those that require visual understanding. 
We order the considered computer vision tasks from the most static to ones that respond to user engagement, primarily: scene-level understanding tasks -- \emph{localisation} and \emph{3D understanding}, followed by tasks at the action level -- \emph{action anticipation}, \emph{action and object recognition} and \emph{gaze understanding and prediction}. 
Then, we review tasks around understanding people, particularly, \emph{social behaviour understanding}, \emph{full-body pose}, \emph{hand and hand-object interaction}, and \emph{person identification}. 
Moreover, we note two user engagement tasks that are recurring frequently in our narrative stories -- \emph{summarisation} and \emph{dialogue}.
Finally, we introduce \emph{privacy} and the related approaches to preserve sensitive content captured by wearable devices.
Overall, given the multi-modality nature of these tasks, we will also discuss how vision can be integrated with cues from other sensors.
We visualise the connections between our use cases and these tasks in Figure \ref{fig:stories_task}.

\section{Research Tasks and Capabilities}\label{sec:tasks}
For each of the egocentric vision tasks identified in Section~\ref{sec:stories_tasks}, we now provide a structured literature review with dedicated subsections.
Rather than covering the full progress of the field, we find it most informative to focus on \textit{seminal works} that initially defined the task or changed its course as well as \textit{state-of-the-art methods} that are currently achieving best performance. 
We acknowledge there are tens of works that paved the path from those seminal works to current methods but opt for not including them in this paper. 
We encourage interested readers to explore these intermediate works to understand the full progress of the field in each research task.
Additionally, we note \textit{datasets} specifically designed to advance the research in each of these tasks.
We leave the review of more general datasets to Section~\ref{sec:datasets}.
We conclude each subsection with a short reflection on the gap between current state-of-the-art and anticipated future. 

\subsection{Localisation}
\label{sec:localisation}
We divide localisation works into two categories: \textit{visual place recognition} (Section~\ref{sec:localisation1}) and \textit{visual localisation} (Section~\ref{sec:localisation2}). Both contribute to the broader goal of positioning the camera wearer within the surrounding environment using visual data for scene understanding and navigation, but they differ in their primary objectives. Place recognition gives a coarse estimate of 2D coordinates, whilst visual localisation determines the 6-DoF (Degrees of Freedom) of the camera pose. We also review Simultaneous Localisation and Mapping (SLAM) techniques (Section~\ref{sec:localisation3}) -- simultaneously building a map of unknown indoor or outdoor environments and tracking the position or trajectory of the camera.

Note that this task only differs marginally between wearable cameras, hand-held cameras and remote cameras (third-person). 
Additionally, cameras mounted on vehicles share similarities with wearable devices that lie in the viewpoint and perspective from which visual information is captured. These analogies allow us to broaden the scope of existing approaches beyond those exclusive to wearable devices.

\subsubsection{Visual Place Recognition}
\label{sec:localisation1}
Visual place recognition analyses visual cues, from either a single image or a sequence of images, to determine the place or area being observed. In egocentric vision, this relates to ``contextual awareness'', i.e., extracting knowledge of the user’s surrounding. The most commonly used metric for evaluation is the $Recall@N$, which calculates the percentage of relevant or true positive places that are among the top $N$ retrieved results. In other words, it measures how many of the correct places were successfully recognised within the top $N$ ranked places. 

\paragraph{Seminal works}
The early investigations of the problem of recognising the user's location from wearable devices date back to the late 90s, when image-based localisation has been mostly studied as a
classification problem. \cite{starner1998visual} proposed a context-aware system for assisting users while playing the ``patrol'' game, by recognising the room in which the player is operating. 
\cite{aoki1998recognizing} presented an image matching technique for the recognition of previously visited places. 
\cite{torralba2003context} introduced a wearable system capable of recognising familiar locations and categorising new environments into high-level classes such as offices and corridors. They proposed to use that information as priors for object recognition (e.g., tables are more likely to exist in an office). 
\cite{furnari2016temporal} performed temporal segmentation of egocentric videos to highlight the continuous presence of the wearer in pre-defined personal locations. The work uses personal locations as cues for identifying activities. 

Related to visual place recognition is the problem of visual geolocalisation -- estimating the position where a given image or frame in a video was taken by comparing it with a large database of images from known locations. 
Visual geolocalisation is commonly approached as an image retrieval problem, with a retrieved image deemed correct if it is within a predefined range from the query’s ground truth position. 
\cite{jegou2010aggregating} proposed VLAD (vector of locally aggregated descriptors), an image descriptor derived from sift descriptors, bag of works and fisher kernels.
\cite{galvez2012bags} presented a fast and efficient approach for place recognition using binary descriptors. 
A few years later, \cite{arandjelovic2016netvlad} offered the first CNN-based approach for place recognition with weak supervision. From that work on, all methods have been using learned embeddings with some form of aggregation or pooling. 

The combination of GPS and visual information to localise users in an environment has also been investigated. \cite{capi2014guide} proposed an assistive system able to guide
visually impaired people in urban environments, and \cite{ahmetovic2016navcog} proposed a smartphone app which can perform accurate and real-time localisation over large spaces. 

\paragraph{State-of-the-art papers}
Current literature has shifted focus towards developing methods specifically tailored for visual geolocalisation. Most recent works aim at better training time scalability to exploit large-scale data. \cite{berton2022rethinking} introduced CosPlace, a method that uses a classification task as a proxy to train the model that is used at inference to extract discriminative descriptors for retrieval. \cite{zhu2023r2former} proposed R$^2$Former, a place recognition architecture that builds on the success of vision transformers and fuses multi-level attention information to generate global and local descriptors which are used for re-ranking. MixVPR by \cite{ali2023mixvpr}, is a new feature aggregation technique that takes in input feature maps from pretrained networks, and iteratively combines them using a stack of multi-layer perceptrons in a cascade of feature mixing. 

\subsubsection{Visual localisation}
\label{sec:localisation2}
Visual localisation refers to the process of determining the pose (position and orientation) of a camera with respect to a known 3D scene or environment, based on visual information. Approaches for visual localisation divide into hierarchical localisation pipelines, consisting of image retrieval, local feature extraction and matching. These are followed by 2D-3D correspondence mapping and pose estimation, and absolute pose regressors, that estimate the camera pose with a single forward pass, using only the query image. A commonly used metric for evaluating visual localisation tasks is the average of median position and orientation errors in meters and degrees, respectively.

\paragraph{Seminal works}
The work of \cite{irschara2009structure} explored the transition from point cloud-based reconstruction to efficient feature-based localisation via Structure-from-Motion (SfM). After computing a representative set of 3D point fragments that cover a 3D scene from arbitrary viewpoints, they matched directly the pose of the query image. The last stage uses the resulting 2D-3D matches for pose estimation using Random Sample Consensus (RANSAC) algorithm. 
\cite{sattler2011fast} made significant contributions by introducing an efficient and direct matching approach between 2D query images and 3D reference data. \cite{kendall2015posenet} presented a deep learning-based approach for camera localisation. Their Convolutional Neural Network (CNN) architecture, called PoseNet, enabled real-time and accurate estimation of camera poses in 6-DOF, by regressing the 6-DoF camera pose from a single RGB image in an end-to-end manner with no need for additional engineering. 
\cite{blanton2020extending} extended pose regression to multiple scenes by proposing the Multi-Scene PoseNet, where the network first classifies the particular scene related to the input image, and then uses it to index a set of scene-specific weights for regressing the pose. Also, the work of \cite{sattler2016efficient} contributed to large-scale image-based localisation by introducing an efficient and effective prioritised matching algorithm. 

\paragraph{State-of-the-art papers}
\cite{shavit2021learning} presented a novel approach using transformers for multi-scene pose regression. The approach uses encoders to focus on
pose-informative features and decoders to transform encoded scene identifiers to latent pose representations. Generally, algorithms for visual localisation mostly rely on complex 3D point clouds that are expensive to build, store, and maintain over time. \cite{do2022learning} trained a CNN to detect the appearance of a sparse set of 3D scene points (scene landmarks), and showed that those predicted landmarks can yield accurate pose estimates, while being privacy preserving and requiring low data storage. \cite{panek2022meshloc} explored dense 3D scene models as an alternative to the sparse Structure-from-Motion point clouds as they are more flexible than SfM-based representations and can be rather compact. Moreover, storing the original images and extracting features when needed takes up less memory than storing the features. 

\subsubsection{Simultaneous Localisation and Mapping (SLAM)}
\label{sec:localisation3}
SLAM is a technique used to build a map of an unknown environment while simultaneously estimating the camera pose within that environment. In this section, we focus on Vision SLAM (V-SLAM), which refers to those SLAM systems which use cameras as the main input sensors. In general, V-SLAM algorithms have three steps: initialisation, tracking, and mapping. The initialisation determines the global coordinates and builds an initial map. The tracking step involves the continuous estimation of the camera pose. In general, during this stage the algorithm extracts 2D–3D correspondences between the current frame and the map. Finally, the mapping step results in a sparse, semi-dense, or dense 3D reconstruction.
SLAM algorithms can be mainly classified into two categories: feature-based and direct. Feature-based methods rely on sparse features for tracking, with the correspondences being used to refine poses through Structure-from-Motion techniques. Direct methods use the sensor data without pre-processing, estimating camera poses within an expectation maximisation framework. The most commonly used metric is the Root Mean Square Error (RMSE), which measures the difference between estimated and ground truth camera poses and map points, providing an overall indication of accuracy.

\paragraph{Seminal works}
The first applications of SLAM to wearable cameras are from \cite{davison2003real} and \cite{mayol2005applying}. \cite{davison2003real} proposed a general method for real-time, single-camera V-SLAM and studied its application to the localisation of a wearable robot with active vision. The approach proposed by \cite{mayol2005applying} enables prolonged periods of focused attention on specific areas of interest, followed by deliberate and controlled redirection of gaze to different parts of the scene. This reduces the need for frequent feature initialisation, and enhances overall system robustness. \cite{castle2010combining} used monocular SLAM and object recognition for AR.  
\cite{badino2011head} introduced a head-wearable stereo system for structure and motion estimation.
\cite{alcantarilla2012combining} developed a wearable stereo system that combines SLAM with dense scene flow estimation to segment moving objects in the scene.
\cite{murillo2012wearable} proposed to use wearable omnidirectional vision systems to augment people's navigation and recognition capabilities. Their approach involves accurate ego-motion estimation and topological/semantic localisation, enabling precise user guidance. 

One problem of monocular SLAM is \textit{scale drift}. It occurs during the initialisation of monocular SLAM, where the scale is initially set to a real or arbitrary value. However, as the camera moves and old landmarks are lost while new ones are initialised, the scale of the scene changes continuously. To address this issue in large environments, \cite{gutierrez2016true} proposed an approach that computes the true scale dynamically using visual odometry estimates from wearable single cameras. Their method relies on the characteristic oscillatory movement of the human body during walking to extract scale information, making it particularly suitable for wearable systems.

The nature of egocentric videos, characterised by sharp head rotations and predominantly forward motion, leads to rapid changes in the camera view, resulting in short and noisy feature tracks. Additionally, the dominant 3D rotation caused by natural head motion further reduces parallax, leading to triangulation errors.
To address these issues, \cite{patra2017computing} proposed a fast and robust egomotion estimation method for egocentric videos, using a local loop closure technique aligned with the wearer’s head motion. 

\cite{suveges2021egomap} proposed a semantic, non-geometric, human-centred form of SLAM, by constructing a representation of a user’s everyday environment in terms of locations that they frequent and their patterns of transition between, and their behaviours within those locations. 

\paragraph{State-of-the-art papers}
With the rise of AR applications, achieving precise alignment of virtual content with the user's physical surroundings has become crucial. To accomplish this, modern devices are equipped with a range of sensors. One notable example is the HoloLens, which incorporates four tracking cameras and a time-of-flight range camera. One of its key features is spatial mapping, which allows the device to create a detailed map of its surrounding environment \citep{hubner2020evaluation}. Using spatial mapping, the HoloLens scans the area within a 70-degree cone, capturing depth information from distances between 0.8 and 3.3 meters. Based on the data, it reconstructs a mesh representation of the observed scene, which serves as a foundation for accurately placing virtual objects within the real world.
Meta's Aria glasses have been also recently released with multiple sensors such as stereo cameras, dual inertial measurement units, spatialised microphones, eye tracking cameras and more. They make use of localisation and mapping techniques to build ``LiveMaps'', a virtual 3D representation of the world.

The combination of neural radiance fields (NeRF, \cite{mildenhall2021nerf}) and SLAM has also emerged as a recent trend. By utilising the capabilities of SLAM for accurate pose estimation and dense depth maps, together with the power of NeRF, it is possible to generate real-time neural scene representations \citep{rosinol2022nerf}.  
\cite{haitz2023combining} proposed an acquisition pipeline that enables real-time image and pose streaming through a TCP client-server application, allowing simultaneous training of Instant-NeRF. The HoloLens acts as the image and pose server, while the client application receives the images and writes them into a GPU image buffer. Instant-NeRF model is incrementally trained using the incoming image stream. Additionally, a fast geometric reconstruction of the scene is performed by querying the trained network based on sample rays from the training poses.

\paragraph{Datasets}
For \textit{visual place recognition}, \cite{furnari2016temporal} collected a dataset of egocentric videos containing 10 personal locations of interest. More recently, \cite{milotta2019egocentric} collected and publicly released a dataset of egocentric videos asking 12 subjects to freely visit a natural site with a total of 6 hours of recording. \cite{ragusa2020ego} proposed a dataset of egocentric videos for visitor behaviour understanding, including 27 hours of video acquired by 70 subjects, with labels for 26 environments which allow room-level localisation.

In \textit{visual geo-localisation}, all previous datasets capture an autonomous driving viewpoint which is very different from that of a wearable camera or are collected using hand-held devices. This lacks characteristic head-mounted motion patterns.  Up to our knowledge, no dataset is available for \textit{visual geo-localisation} from a body-worn camera.

For \textit{visual localisation}, \cite{sarlin2023orienternet} created a dataset using Meta's Aria glasses at 3 locations in Seattle (Downtown, Pike Place Market, Westlake). In each location, they recorded 3 to 5 sequences following the same trajectories, for a duration of 5 to 25 minutes varying by location, and a total of 3 hours of recordings. Each device is equipped with a GPS sensor, IMUs, grayscale SLAM cameras, and a front-facing RGB camera. 

\cite{suveges2021egomap} proposed the first dataset specifically designed for SLAM applications on egocentric vision. Five videos were recorded using a head-mounted GoPro Hero 4, for a total of four hours of videos including transition segments between locations, repeated visits by a user to multiple distinct locations, and unique labels for all visited locations. 

Multiple sensors data, such as depth images, hand and eye tracking data, are essential for accurate spatial mapping and scene understanding for XR applications. \cite{chandio2022holoset} proposed HoloSet, a dataset captured using Microsoft Hololens 2, that contains the raw synchronised data streams from the following sensors: depth, RGB, four grayscale visible light tracking (VLC) cameras, and an IMU, along with the ground truth pose trajectory. It contains 29 sequences and 78.5k samples that cover more than 6200 meters. \cite{sarlin2022lamar} introduced a large-scale dataset of over 100 hours and covering 45’000 square meters of multi-sensor data streams (images, depth, tracking, IMU, BT, WiFi) captured using HoloLens 2 and iPhone/iPad devices in diverse environments, including a historical building, a multi-story office building, and part of a city center. Data include indoor and outdoor images with varying illumination, semantic changes, and dynamic objects. 

Importantly, all previous datasets were collected specifically for localisation purposes. In these recordings, the camera wearer is only navigating the scene to capture these sequences and is not carrying out any of their daily tasks necessarily.
It is acknowledged to be challenging to perform visual localisation from unscripted egocentric footage of actual activities \citep{suveges2021egomap}.
Recently, \cite{EPICFields2023} provided 6 DoF camera positions for 99 hours of the EPIC-KITCHENS dataset \citep{Damen2021rescaling} in 45 home kitchens.
Camera estimates are achieved through intelligent sampling without any additional sensors or sequences specific for localisation.
However, no ground truth is available for this dataset and these camera estimates are only qualitatively evaluated.  

\paragraph{For the future} 
Despite progress made in recent localisation techniques for robotics and autonomous vehicles applications \citep{kazerouni2022survey,cheng2022review}, the robustness of these algorithms in dynamic and changing environments as the ones captured by wearable devices require further development. For visual place recognition, current state-of-the-art performance are $64.0\%$ recall on the Mapillary challenge. On LaMAR \citep{sarlin2022lamar}, the recent benchmark for localisation and mapping in the context of AR, results on single-frame localisation only achieve 45.6\% / 61.3\% recall at (1\degree, 10cm)/(5\degree, 1m).
Additionally, wearable devices often have limited computational resources, which can limit the complexity and accuracy of localisation algorithms. The most attractive application for localisation in head-mounted devices is AR, where the objective is to place virtual content in the physical 3D world, persisting it over time, and sharing it with other users. 

Common benchmarks over the last years often rely on limited datasets with minimal scene diversity and sensors. 
These datasets also are typically collected specifically for localisation, through navigation-only sequences rather than capturing individuals engaged in actual activities.
However, ongoing research efforts and advancements in computer vision, sensor technologies, and wearable computing such as Meta's Project Aria glasses and Microsoft HoloLens are paving the way for future applications of localisation on wearable devices, enabling promising use cases such as indoor navigation and AR experiences. 
\subsection{3D Scene Understanding}
\label{sec:3dunderstanding}
The goal of 3D scene understanding is to build an AI agent able to interpret the surrounding environment and explore possible interactions with it.  
This also involves identifying relevant objects in the scene and reasoning on their locations. 
The complexity of the field has attracted attention over the last few years, leading to the proposal of numerous sub-tasks and datasets. 
Their diversity underscores the multifaceted nature of 3D scene understanding, prompting researchers to explore various evaluation measures tailored to specific challenges. 

\paragraph{Seminal works}
The first work to explore task-relevant objects in 3D is that of \cite{you-do_damen}.
Given a mapped environment, gaze estimation is used to cluster interaction regions into task-relevant objects and their modes of interaction.
For studying human-centric interactions with the environment, \cite{bertasius2015exploiting} proposed to utilise egocentric stereo cameras to establish an egocentric object prior within a first-person view RGBD frame, which could then be employed for 3D saliency detection. Through observations, it was discovered that humans possess a fixed size prior to salient objects, indicating that salient objects in 3D undergo consistent transformations, enabling people's visual system to perceive them with an approximately constant size. This insight led to the identification of a consistent egocentric object prior that can be characterised by its shape, size, depth, and location within the first-person view. 
\cite{rhinehart2016learning} focused on learning and predicting ``Action Maps'' that encode the user's ability to perform activities at various locations. By mapping actions to specific regions within a scene, this technique enables the understanding and prediction of human activities in a given environment.
\cite{li2022egocentric} focused on anticipating as early as possible the target location of a person's object manipulation action in a 3D workspace. While this is a special case of trajectory forecasting, the latter is infeasible in manipulation scenarios and the hands often are located outside the field of view. Therefore, focusing on predicting the 3D target location gives a better understanding of possible interactions with objects, useful for applications such as robot planning and control.
Recently, \cite{grauman2022ego4d} proposed the task of Visual Queries with 3D Localisation (VQ3D), which focuses on retrieving the relative 3D localisation of a query object with respect to a current query frame. 
Another interesting problem has been proposed by \cite{majumder2023chat2map}: building the map of a previously unseen 3D environment by exploiting shared information in the egocentric audio-visual observations of participants in a natural conversation. Finally, \cite{pan2023copilot} introduced the task of collision prediction and localisation from unposed egocentric videos, which aims at predicting \textit{when} and \textit{where} a collision with the environment might occur.

\paragraph{State-of-the-art papers}
\cite{nagarajan2020learning} introduced a reinforcement learning approach where an embodied agent autonomously discovers the affordance landscape in new, unmapped, 3D environments, enabling interaction exploration. They rewarded the agent for quickly interacting with all objects in an environment and trained an affordance model online to segment images according to the likelihood of each of the agent’s actions succeeding. \cite{do2022egocentric} focused on predicting depths and surface normals of the surrounding environment from a single view egocentric image. They addressed challenges derived from the use of wearable devices such as tilted images and the presence of dynamic foreground objects by proposing an image stabilisation method which transforms titled images to a canonical orientation for better learning. \cite{nagarajan2022egocentric} proposed learning environment-aware video representations that encode the surrounding physical space, facilitating the prediction of local environment states at different time-steps. 
These states are used to train a transformer-based video encoder model, which gathers visual information from the entire video and constructs an environment memory. This memory can then be accessed to predict the local state at any specific point in the video. 

\cite{liu2022egocentric} proposed the task of jointly recognising and localising actions of a user on a known 3D map from egocentric videos. They proposed a novel deep probabilistic model that utilised a Hierarchical Volumetric Representation (HVR) of the 3D environment and an egocentric video to infer the 3D action location and recognise the action based on contextual cues.
Other works focused on object visual query localisation in the 3D space. \cite{xu2023my} proposed a transformer-based module that incorporates object-proposal set context while considering query information. \cite{mai2023egoloc} formalised a pipeline that better integrates 3D multiview geometry with 2D object retrieval from egocentric videos, leading to improved camera pose estimation and substantially improved VQ3D performance. The process involves three main steps: first, a sparse 3D reconstruction is performed using Structure from Motion (SfM) to estimate 3D poses and create a sparse 3D map. Second, the frames of an egocentric video and a visual crop of a query object are fed into a model that retrieves response frames and their corresponding 2D bounding boxes. Third, for each response frame, the depth is estimated and the object centroid is back-projected to 3D using the corresponding camera pose. \cite{qian2023understanding} addressed the task of predicting the 3D location, physical properties and affordance of objects from single images. Given a set of query points, the output includes the potential 3D interaction, in terms of movable, location, rigidity, articulation, action and affordance. They achieve that using a transformer-based model which builds on a detection backbone. 

\paragraph{Datasets}
General-purpose egocentric datasets such as EPIC-KITCHENS \citep{Damen2021rescaling} and Ego4D \citep{grauman2022ego4d}, which are reviewed in Section~\ref{sec:datasets}, can be used for scene understanding. Additionally, other task-specific datasets have been proposed. 
The Egocentric Depth on everyday INdoor Activities (EDINA) dataset by~\cite{do2022egocentric} has the goal of facilitating learning the visual representation of dynamic egocentric scenes. It comprises more than 500K synchronised RGBD frames and gravity directions captured from an egocentric viewpoint with diverse daily activities, for a total of 16 hours RGBD recording. EgoPAT3D \citep{li2022egocentric} is a large multimodality dataset of more than a million frames of RGB-D and IMU streams, which has been designed for the task of anticipating the target location of a person’s object manipulation action in a 3D workspace. The total collection contains 150 recordings, 15 household scene point clouds, 15,000 hand-object actions, 600 minutes of raw RGB-D/IMU data, 0.9 million hand-object action frames, and 1 million RGB-D frames for the entire dataset. \cite{qian2023understanding} introduced the 3D Object Interaction Dataset (3DOI), with Internet videos, egocentric videos and indoor images. For the egocentric part, they sampled 2K images from EPIC-KITCHENS~\citep{Damen2021rescaling}. Images come with 3D ground truth, including depth and surface normals, and 5 interactable query points, including both large and small objects. For each of them, they annotated whether the object is movable, its location, its rigidity, its articulation, the potential action that can be done with it, and its affordance (where it is possible to interact with the object). 

The Aria Digital Twin \citep{pan2023aria} is an egocentric dataset captured using the Aria glasses that contains 200 sequences of real-world activities conducted by Aria wearers in two real indoor scenes with 398 object instances (324 stationary and 74 dynamic).  Each sequence includes raw data of two monochrome camera streams, one RGB camera stream, two IMU streams, complete sensor calibration, ground truth data including continuous 6-degree-of-freedom (6DoF) poses of the Aria devices, object 6DoF poses, 3D eye gaze vectors, 3D human poses, 2D image segmentations, image depth maps and photo-realistic synthetic renderings. \cite{ravi2023odin} proposed ODIN (the OmniDirectional INdoor dataset), a large-scale dataset of more than 300K omnidirectional images capturing a diverse range of activities of daily living. This includes scans of the recording environments from a 3D scanner and camera-frame 3D human pose estimates, enabling its use for scene understanding purposes. 
Recently, \cite{EPICFields2023} released EPIC Fields, an augmentation of EPIC-KITCHENS with 3D camera poses. It reconstructs 96\% of videos in EPIC-KITCHENS, registering 19M frames in 99 hours recorded in 45 kitchens, creating an opportunity to bring 3D geometry and video understanding closer together. \cite{mur2023multi} built a dataset on affordances based on the EPIC-KITCHENS dataset, EPIC-Aff, which provides interaction-grounded, multi label, metric and spatial affordance annotations. Finally, \cite{shapovalov2023replay} introduced Replay, a collection of multi-view, multimodal videos of humans interacting socially. It contains long scenes in an indoor environment, each captured in 4K resolution using 8 static DSLR cameras and 3 head-mounted GoPro cameras, along with a comprehensive microphone array. In total, it contains 66 hours of footage. It is suitable for a series of tasks, such as novel-view audio/visual synthesis and 3D reconstruction.

\paragraph{For the future}
Egocentric videos provide a natural connection between the activities of the camera wearer and the surrounding 3D spatial context. Although this is an intrinsic characteristic of egocentric vision,  
However, motion blur, and unusual viewpoints caused by how egocentric videos are captured introduce overwhelming challenges, causing  3D reconstruction to struggle with dynamic content. 
As a result, much work remains before we can have a 3D understanding of dynamic phenomena, such as actions and activities. Another promising future direction is working with both egocentric and exocentric views. 
By combining the insights gained from both perspectives, researchers could potentially unlock a more comprehensive understanding of complex scenes and human interactions. This approach however is limited in its applicability for our anticipated \egox{} future, where exo views are unlikely to be part of our everyday lives. 

\subsection{Recognition}
\label{sec:recognition}
Recognition in egocentric vision is crucial as it involves understanding interactions as well as the objects the wearer interacts with and their actions. This dual focus on both actions and objects enables a comprehensive understanding of the wearer's environment and activities. We divide the works into \textit{action} (Section \ref{sec:recognition_actions}) and  \textit{object} (Section \ref{sec:recognition_objs}) respectively.

\subsubsection{Action Recognition}\label{sec:recognition_actions}
The goal of egocentric action recognition is to classify human actions from the egocentric point of view, i.e., the person wearing the camera is carrying out the action. In areas such as robotics and AR, egocentric action recognition is critical to enable downstream applications, such as contextual recommendations or reminders.
The egocentric point of view and a wearable, hence moving in dynamic and often unpredictable ways, camera presents an higher level of complexity when compared to standard action recognition from a fixed and remote cameras. 
Moreover, as the camera wearer themselves are largely out of the field of view, several challenges come from the partial observability of the main actor.

One possible way to address this is to leverage complementary cues to support the visual modality. 
Audio, gaze and temporal dynamics via optical flow are examples of information that play a relevant role in understanding the performed actions.
As managing multiple modalities may be costly, recent advancements are focusing on low-energy consumption architectures and higher-level action understanding.
The task is formalised as a classification problem and generally evaluated with top-1 and top-5 accuracy.

\paragraph{Seminal works}
Early works considered the egocentric perspective to improve action recognition for robots ~\citep{johnson2005perceptual} and humans \citep{surie2007activity}.
\cite{5204354} explored action recognition for egocentric vision with Inertial Measurement Units (IMUs) used for temporally identifying the actions.
\cite{kitani2011fast} authored a pioneering work about tackling action recognition from egocentric sports videos in an unsupervised manner.
The research field gained large momentum after the introduction of the dataset collecting activities of daily living (ADL)~\citep{pirsiavash2012detecting}, particularly thanks to its large set of annotations on activities, object tracks, hand positions, and interaction events.  
To deal with complex object interactions and long-range temporal activity structures, the authors also introduced tailored representations that included temporal pyramids and composite object models. 

The work by \cite{fathi2012learning} was the first to highlight the utility of gaze: it presented a probabilistic generative model for simultaneously recognising daily actions and predicting gaze locations from egocentric videos.
\cite{li2015delving} proposed to combine features encoding hand pose, head motion and gaze direction together with motion and object features coming from local descriptors. 

In the last years, deep learning has alleviated the burden of manually defining features. \cite{singh2016first} was the first work to use CNNs for end-to-end learning and classification of the wearer’s actions. Since then, the attention moved to learning architectures with novel pooling mechanisms \citep{ryoo2015pooled} or temporal convolutions on motion fields for long-term activity recognition \citep{poleg2016compact}. 

Techniques that use recurrent neural networks such as Long Short-Term Memory (LSTM) \citep{cao2017egocentric,verma2018making} and Convolutional Long Short-Term Memory (ConvLSTM) \citep{sudhakaran2017convolutional,sudhakaran2018attention} have been proposed to better encode temporal information. \cite{sudhakaran2019lsta} proposed a new recurrent neural unit that augments LSTM with built-in spatial attention and a revised output gating. This allows to focus on features from relevant spatial parts while attention is being tracked smoothly across the video sequence. 
\cite{tang2017action} added an additional stream to take as input depth maps enabling the model to encode 3D information present in the scene.
\cite{kazakos2019TBN} proposed an end-to-end trainable mid-level fusion Temporal Binding Network~(TBN) on top of a convolutional network to asynchronously fuse audio, RGB and optical flow across multiple temporal windows. 

The success of the transformer architecture has also given rise to a new line of works that employ transformers as a backbone for processing videos, with the most popular ones being those by \cite{patrick2021keeping} and \cite{arnab2021vivit}. These works extend the vision transformer to operate on multiple frames within videos. However, they were not developed specifically for egocentric videos, and report results on both third-person and egocentric videos using the same architecture.

Still, training a deep model is data and energy intensive and several works have been focusing on reducing the related costs. \cite{possas2018egocentric} defined a reinforcement learning based technique for understanding actions using less energy. 
\cite{8578870} proposed to jointly learn from first- and third-person videos using weak supervision. 
Similarly, \cite{li2021ego} introduced  an approach for pretraining egocentric video models using large-scale third-person video datasets. 
\cite{min2020integrating} presented a probabilistic approach to estimate the gaze and utilise it for action recognition, avoiding the need for expensive gaze recording equipment.  
\cite{9879783} showed that the visual information collected by event cameras is suited for egocentric action recognition thanks to the lack of motion blur, high temporal resolution, and reduced power consumption. 

Aiming to reduce the burden and uncertainty involved in  the annotations of temporal bounds, a different line of works considered the problem of recognising actions using a single timestamp originating from narrations as supervision rather than temporal bounds  \citep{moltisanti2019action}.

Another approach to egocentric action recognition is to consider it as a procedural problem and learn the key steps required to perform a task upon observing multiple egocentric videos as done in \cite{EgoProceLECCV2022}. 
This work is restricted to procedural tasks but is a venue for exploration as opposed to recognising isolated actions. 

\paragraph{State-of-the-art papers}
\cite{kazakos2021MTCN} developed an approach specific to egocentric videos using an audio-visual transformer with the visual features from \cite{patrick2021keeping}. Importantly, in this work, the action is not seen in isolation: the untrimmed video and context are explored along with a language model providing action sequencing to enhance the predictions. 
This approach reported significant performance improvement over prior works, with action recognition reaching 49.6\% on the validation set of EPIC-KITCHENS-100.

Following the trend of Transformers, \cite{wu2022memvit} proposed a memory-based approach for efficient long-term video understanding. It uses the ``keys'' and ``values'' of a transformer as memory. The queries attend to an extended set of keys and values, which come from both the current time and the past. Each layer attends further down into the past, resulting in a significantly longer receptive field. They achieve 48.4\% of action recognition accuracy on the EPIC-KITCHENS-100 dataset with much less model parameters (0.5$\times$ the parameters of~\cite{patrick2021keeping}). 

Recent works focused on designing new multi-modal integration strategies, in order to build models that work well across modalities, instead of being over-optimised for each modality. \cite{girdhar2022omnivore} proposed a transformer-based model which, leveraging the flexibility of transformers, is trained jointly on classification tasks from different modalities -- 2D images, 3D images and videos. They achieve an impressive top-1 performance of 47.4\% on EPIC-KITCHENS-100 validation set using their largest Swin-B transformer model. \cite{yan2022multiview} adapted a multi-view transformer to multi-modal inputs: they created multiple representations or “views” by tokenising spectrogram, optical flow, and RGB using tubelets of different sizes. These tokens are fed into separate encoders and further fused through a fusion module, and aggregated by a global encoder. 
Using both modalities, the approach achieves action recognition of 47.2\% on EPIC-KITCHENS-100 validation set. They outperform the approach from \cite{yan2022multiview} by 1\% but are still below other approaches previously published such as \cite{girdhar2022omnivore,kazakos2021MTCN}. 
All the prior work, except \cite{kazakos2021MTCN} have been developed for general action recognition and the architectures are not optimised for egocentric vision specifically.

\cite{gong2023mmg} studied the problem of generalisation when data from certain modalities is limited or even completely missing during inference. They proposed a method for multi-modal generalisation based on a fusion module with modality dropout training, a cross-modal contrastive alignment loss, and a cross-modal prototypical loss for better few-shot performance. To further improve the efficiency, they jointly trained a memory compression module for reducing the memory footprint. \cite{radevski2023multimodal} proposed to distill knowledge from a high-performing but impractical multimodal ensemble into a light-weight RGB-based model. \cite{tan2023egodistill} achieved efficient recognition by  combining RGB with the head motion information from IMUs.

\cite{wang2023learning} addressed the task of unpaired multi-view video learning. To this purpose, they introduced a method that aligns multi-view pseudo-pairs with high similarities in a semantics-aware manner. They allow first-person videos to gain insights from samples of varying views or modalities. Similarly, \cite{xue2023learning} learn fine-grained frame-wise video features that are invariant to both the ego and exo views from unpaired data. They achieved that through a self-supervised contrastive-based temporal alignment objective.

Another recent trend is to leverage over Large Language Models (LLMs), to obtain stronger representations. \cite{zhao2023learning} used LLMs to automatically generate text pairing for videos, by densely annotating rich textual descriptions. 
When using those to learn video-text embeddings contrastively, and then evaluating on action recognition as a downstream task, results outperformed previous state-of-the-art on EPIC-KITCHENS-100, with 51.0\% accuracy on action recognition. This sets the current state-of-the-art performance on the validation set of this dataset. 
Language has also been used in \cite{plizzari2023can} as a robust modality for improving domain generalisation to multiple domains. Starting from the rich diversity of Ego4D in terms of both scenarios and geographical locations, they proposed to represent each video as a cross-instance reconstruction of videos from other domains. Reconstructions are paired with text narrations to guide the learning of a domain generalisable representation. 

\cite{Shah_2023_ICCV} considers learning keysetps for procedural problem using multiple modalities. To enable AR applications, \cite{Shah_2023_ICCV} utilise optical flow, depth, or gaze and propose the BMC2 loss to force modalities from multiple datasets to be close in the representation space. The work improves the F1 Score on the EgoProceL dataset by 14\% and achieves state-of-the-art results.

Before concluding this section, we note the relevant tasks of action segmentation and action detection. The latter is distinct from action recognition as it aims to detect the start and end of action instances in long untrimmed videos as well as predicting the action categories. Previous works on action segmentation considered graph-based temporal reasoning \citep{huang2020improving} and segmentation from single timestamps \citep{li2021temporal}, while, up to our knowledge, \cite{Wang2023EgoOnlyEA} offer the first approach to action detection specifically for the egocentric domain without exocentric pretraining. This topic requires further exploration in untrimmed egocentric videos.
We refer those interested in exploring action detection to \cite{vahdani2022deep} for up-to-date methods. 

\paragraph{Datasets}
The most popular datasets for action recognition, EPIC-KITCHENS-100, Ego4D and EGTEA are detailed in Section~\ref{sec:datasets}. Additionally, several specialised datasets have been proposed to explore different aspects of action recognition in egocentric videos. 
\cite{kitani2011fast} proposed a dataset made of videos both recorded in-house and sourced from YouTube. The first video, recorded on a QUAD, consists of 124 video splices (a video splice contains 60 frames) and contains 11 ego-actions. The second video, recorded in a park, is a 25 minute workout video which contains 766 video splices and contains 29 different ego-action categories. Six egocentric YouTube sports videos have also been annotated to understand actions in outdoor sports videos.

DataEgo~\citep{possas2018egocentric} and Multimodal Egocentric Activity~\citep{7789544} datasets have been used for evaluating methods that focus on activity recognition with limited resources or on a budget. In DataEgo~\citep{possas2018egocentric},  Images from the camera have been synchronised with readings from the accelerometer and gyroscope. In total, it contains approximately 4 hours of continuous activity while its multi-modal subset has only 50 minutes of separate activities. The Multimodal Egocentric Activity dataset~\citep{7789544} contains 20 distinct life-logging activities, which are recorded both indoor and outdoor with significant changes in the illumination conditions. The dataset has 200 sequences in total and each activity category has 10 sequences of 15 seconds each. It also includes other synchronised sensor data: accelerometer, gravity, gyroscope, linear acceleration, magnetic field and rotation vector. Charades-Ego~\citep{8578870} aims to bridge the gap between egocentric and third-person videos, providing a dataset with paired first-person and third-person videos involving $112$ individuals and $4000$ paired videos. \cite{bockwear} recently introduced WEAR, an outdoor sports dataset for both vision- and inertial-based human activity recognition. The dataset comprises data from 18 participants performing a total of 18 different workout activities with untrimmed inertial (acceleration) and camera (egocentric video) data recorded at 10 outdoor locations totalling 15 hours. 

\subsubsection{Recognising objects}\label{sec:recognition_objs}
Object recognition in egocentric vision is pivotal for applications in augmented reality and robotics but remains a challenging task. Videos recorded from the first-person point of view capture spontaneous, unscripted scenes, in densely packed environments, where objects of various scales are closely packed and often occluded. 

\paragraph{Seminal works} \cite{ren2010figure} introduced a figure-ground segmentation system for egocentric object manipulation videos captured from a wearable camera, in order to separate the moving hands and the objects in-hand from the background. 
\cite{kang2011discovering} tackled the problem of object instance discovery, defining a method for finding new objects that a person can encounter in their daily living. 
\cite{fathi2011learning} addressed the problem of learning object models from egocentric videos of household activities, using weak supervision. For each activity sequence, the method is merely supervised by the names of the objects which are present within it. They propose a segmentation method to partition each frame into hand, object, and background categories. \cite{bolanos2015ego} attempted object discovery -- that is detecting new object instances or concepts, and assigning them a label without prior training. \cite{damen2016you} proposed a fully unsupervised approach to discover objects and their usage from multiple users in a common environment. It consists of discovering task relevant objects, building an appearance model for each, distinguishing different ways in which each discovered object has been used and discovering the spatio-temporal dependencies between object interactions.

Combined with tracking, \cite{bertasius2017unsupervised} formulated the object detection task as an interaction between the segmentation and recognition agents. Initially, the segmentation agent generates a candidate object mask for each image, and relays this mask to the recognition agent, which then tries to learn a classifier using visual semantics and spatial cues. Other works addressed the problem of object tracking. \cite{alletto2015egocentric} developed an approach based on visual odometry and 3D localisation for tracking objects moving around a person.

With the release of the Ego4D dataset \citep{grauman2022ego4d}, new benchmarks involving object understanding have been proposed. The visual Queries Localisation (VQL) task aims to retrieve given query objects from an egocentric video. The hands and objects benchmark captures how the camera-wearer changes the state of an object by using or manipulating it – particularly capturing object state change. \cite{yu2023video} proposed a new benchmark for segmenting state-changing objects in each frame of the video, given the first frame mask as reference. \cite{zhao2023instance} proposed a new benchmark for studying instance tracking in 3D scenes from egocentric videos. Finally, \cite{herzig2022object} used object knowledge to achieve action recognition, as objects can be essential for recognising actions. They presented an object-centric approach that extends video transformer layers with a block that directly incorporates object representations.

In~\cite{VISOR2022}, semi-supervised video object segmentation is evaluated on egocentric videos, focusing on active objects using a newly annotated dataset of object segmentations.

\paragraph{State-of-the-art papers} \cite{akiva2023self} proposed a self-supervised object detection model from egocentric videos. It uses two patch-wise objectives: an objective function operates in the temporal space, enforcing similarity of multi-temporal patches, and a function in the scale space, enforcing similarity of multi-scale patches. The former captures appearance variations in time such as viewing angles and illumination conditions, and the latter captures appearance variations in scale. \cite{wu2023label} examined the problem of continual object detection in egocentric streaming videos, by a plug-and-play module inspired from the complementary learning systems theory.

On tracking, \cite{huang2023tracking} proposed DETracker, a method that jointly detects and tracks deformable objects in egocentric videos. DETracker consists of three key components: the motion disentanglement network (MDN), the patch association network~(PAN), and the patch memory network (PMN). MDN plays a crucial role in efficiently estimating the motion flow between successive frames, where it distinguishes between global camera motion and local object motion. This separation ensures the algorithm's robustness in the presence of significant ego motion. PAN is responsible for tracking deformable objects by breaking them down into patches and locating corresponding patches in future frames for each individual patch. PAN maintains and continually updates feature embeddings of tracked objects over an extended time window.

In the VQL setting (\cite{grauman2022ego4d}), due to random viewpoints and the large number of possible object classes that are exhibited in egocentric recordings, the target object is hard to discover and confused with high-confidence false positives. To tackle this issue, \cite{xu2023my} proposed the CocoFormer, a detection model that incorporates a conditional projection layer. This layer is responsible for generating a transformation matrix based on the query. Subsequently, this transformation is applied to the proposal features, resulting in query-conditioned proposal embeddings. These query-aware proposal embeddings are then inputted into a set-transformer, enabling the model to effectively leverage the global context of the associated frame. \cite{jiang2023single} proposed VQLoC, a single-stage framework. The method jointly models the query-to-frame relationship and frame-to-frame relationships across nearby video frames, and uses that information for end-to-end training. 
\cite{xue2023egocentric} achieved state-of-the-art performance on the Object State Change Classification benchmark, by means of EgoTask Translation (EgoT2), a framework that takes a collection of models optimised on separate tasks and learns to translate their outputs for improved performance on all tasks jointly.
Recently, several works have been proposed that make use of 3D information. \cite{tschernezki2021neuraldiff} proposed a three-stream neural rendering architecture, where the streams model respectively the static background, the dynamic foreground objects, and the actor. \cite{mai2023egoloc} formalised a pipeline that better entangles 3D multi-view geometry with 2D object retrieval from egocentric videos.

Recent state-of-the-art action recognition benchmarks make use of object-related information for better classifying actions. \cite{zhou2023can} proposed an object-guided token sampling strategy that allows to retain a small fraction of the input tokens with minimal impact on accuracy. Moreover, they introduced an object-aware attention module that enriches our feature representation with object information and improves overall accuracy. \cite{zhang2023helping} tasked the model to predict object bounding boxes and names of objects during training in order to learn grounded and fine-grained correspondence between vision and language modalities. 

\paragraph{Datasets}
Several egocentric datasets focused on objects have been built. TEgO (\cite{lee2019hands}) contains egocentric images of 19 distinct objects for training object recognisers. HOI4D (\cite{liu2022hoi4d}) captures videos of human-object interaction with 800 object instances from 16 categories. TREK-150 (\cite{dunnhofer2023visual}) annotated 150 videos from EPIC-KITCHENS~\citep{Damen2021rescaling} for tracking objects from 34 categories. 
VISOR \citep{VISOR2022} annotated 272K manual semantic masks of 257 object classes, 9.9M interpolated dense masks, 67K hand-object relations, covering 36 hours from EPIC-KITCHENS~\citep{Damen2021rescaling}.
EgoObjects (\cite{zhu2023egoobjects}) is a large-scale egocentric dataset for fine-grained object understanding. It contains over 9,200 videos of over 30 hours collected by 250 participants, 654K object annotations from 368 object categories and 14K unique object instances. EgoTracks (\cite{tang2023egotracks}) is a new dataset for long-term egocentric visual object tracking, with more than 22,028 tracks from 5708 average 6-minute videos from Ego4D~\citep{grauman2022ego4d}. PACO (\cite{ramanathan2023paco}) goes beyond traditional object masks and provide richer annotations such as part masks and attributes. It captures both egocentric and non-egocentric views. The PACO-Ego4D subset of egocentric images has 140K part masks annotated in 26.3K images across 75 object classes and 456 object-specific part classes. \cite{kurita2023refego} introduced RefEgo, which is also annotations of Ego4D videos, with more than 12k video clips and 41 hours for video-based object referring expression annotations.

\paragraph{For the future}
Despite the growing interest in action recognition for egocentric videos, there are several areas that warrant attention from the computer vision community. 
Firstly, there are only limited approaches developed specifically for egocentric vision. Most architectures are re-purposed from third-person videos and not optimised specifically for the ego viewpoint or the camera motion.
For instance, the main evident consequence of relying on third-person video pretraining is that the ability to recognise fine-grained actions in egocentric videos is still significantly lower than the corresponding performance in third-person. 
Secondly, even by exploiting transformer architectures and multiple modalities, state-of-the-art methods currently achieve only 51.0\% activity classification accuracy (obtained by \cite{zhao2023learning} on EPIC-KITCHENS-100). 
It is not clear whether approaches are lacking due to the size of datasets, ambiguity (or fine-grained nature) of labels, or the need for new architectures. With multiple possible explanations and avenues for exploration, the field is only progressing slowly. Sequences of papers tend to improve performance by small margins (0.5-1\%).

Thirdly, egocentric vision introduces further challenges, such as the need for modelling long temporal dependencies and learn from long-tail and class imbalanced data. 
\cite{perrett2023use} recently introduced a new benchmark for long-tail recognition in video, including egocentric video. 

Fourthly, despite the early success of integrating gaze for egocentric action recognition, subsequent datasets do not capture the rich, though expensive, egocentric gaze. A few sequences in Ego4D~\citep{grauman2022ego4d} include gaze but these are not labelled specifically with fine-grained actions. Gaze offers the ability to focus the attention on areas of the image and prime for the next actions. However, with the absence of large scale egocentric action recognition datasets that include gaze, this avenue of research is currently under explored.

Fifthly, the recent introduction of the EPIC-SOUNDS dataset~\citep{EPICSOUNDS2023} showcased the need for modality-specific annotations of action classes and temporal extents. The work showcased the disadvantage of training one modality with labels from a different modality.
This perspective on multi-modalities in egocentric video can unlock new approaches for developing multi-modal architectures.

Lastly, the heavy reliance on labeled datasets for training limits the capabilities of models. 
Not only is labelled data expensive to acquire, but the choice of the closed vocabulary of action classes and the granularity of actions remains subjective.
Progressing from a closed subset to open labels remains an open question in most machine learning tasks including egocentric action recognition.
With the advent of LLMs, this surely is the future of recognising actions, despite the lack of metrics to assess success and monitor progress.

In parallel, understanding object affordances—how objects can be used or interacted with—goes beyond basic recognition and it is crucial for applications like assistive robotics, augmented reality, and personal assistance. Such understanding will allow systems to interact more intuitively and effectively with their environment, a key advancement for \egox.
\subsection{Anticipation}
\label{sec:anticipation}
Anticipation tasks aim to predict the future state of the scene from the observation of the present. These tasks are particularly relevant in egocentric vision as they capture an uninterrupted picture of the camera wearer's interaction with the environment and objects, hence providing the opportunity to model their behaviour and understand their goals and intentions.
Following~\cite{rodin2021predicting}, we divide the works focusing on anticipation tasks into three categories, depending on the target of future prediction, namely, \textit{actions} (Section~\ref{sec:anticipation_actions}), \textit{objects} (Section~\ref{sec:anticipation_regions}), and \textit{trajectories} (Section~\ref{sec:anticipation_trajectories}). We would like to note that, while we refer to these works with ``anticipation'', the term ``forecasting'' has also been used in the literature to refer to these tasks.

\subsubsection{Anticipating Actions}
\label{sec:anticipation_actions}
Action anticipation is the task of semantically predicting the next action to take place in a video. Systems able to tackle this task can provide proactive assistance to the users and improve their safety by understanding the camera wearer's goals and future interactions. Current approaches formalise action anticipation as a video classification task which aims to predict a future action from the observation of a video segment of the past. Due to the stochastic nature of the task, methods are required to produce a ranked list of outputs and they are considered successful when the ground truth future action is in the top-k predictions.

\paragraph{Seminal works}
Action anticipation has been studied both in egocentric and third-person vision with similar approaches. The problem has been introduced by \cite{pei2011parsing}, who considered a goal-oriented scenario and used and-or-graphs to represent the different actions which might be performed by a human actor at any given point in an observed video. The feasibility of the task is demonstrated by comparing the proposed solution with human performance. \cite{lan2014hierarchical} subsequently standardised the task, evaluating it at different prediction horizons on videos from TV shows. 

\cite{koppula2015anticipating} showed the benefits of predicting future human poses, future human and object trajectories and future interacted objects for robotics applications. On the side of methodological advancements, \cite{vondrick2016anticipating} first explored the idea of predicting future actions by training deep neural networks to anticipate future representations on unlabeled videos. A similar concept has been further explored by \cite{gao2017red} and \cite{gers2000learning}, who also leveraged Long-Short-Term Memory (LSTM) networks and a reinforcement learning criterion to anticipate future actions at different prediction horizons. 

While the aforementioned works mainly considered the problem of anticipating the next action appearing in the video (short-term action anticipation), \cite{abu2018will} proposed the task of predicting a longer sequence of future actions in the case of goal-driven, structured procedures (long-term action anticipation). Recently, the problem of action anticipation has been studied also in the context of egocentric videos. In particular, \cite{damen2018scaling} first proposed an action anticipation challenge on egocentric videos, and \cite{furnari2019would} systematically tackled the anticipation problem exploring the importance of egocentric cues such as object-based features.

\paragraph{State-of-the-art papers} \cite{girdhar2021anticipative} proposed Anticipative Video Transformer (AVT), an
end-to-end attention-based video modelling architecture
that attends to the previously observed video in order to
anticipate future actions. \cite{gu2021transaction} leveraged transformer-based attention to aggregate features across temporal dimension, modalities, and symbiotic branches (\textit{verb}/\textit{noun} branches) respectively. \cite{zhong2023anticipative} extended the transformer architecture to operate on multiple modalities, by unifying multi-modal data through mid-level fusion and using the obtained representations for anticipating next actions.
\cite{roy2022action} proposed an approach that uses
learned latent goals
to anticipate the next action. Latent goals are accompanied by \textit{goal closeness} and \textit{goal consistency} losses, aiming to produce a visual representation that is closer to the latent goal and consistent throughout consecutive actions. At the moment of writing, \cite{roy2022interaction} achieve state-of-the-art performance on the EPIC-KITCHENS Action Anticipation challenge, by refining video representations using a transformer model computing the change in the appearance of objects and human hands due to the execution of the actions.
Recently, \cite{zhao2023antgpt} achieved state-of-the-art long term action anticipation performance on the Ego4D dataset with a hybrid architecture integrating vision-based action recognition to infer high level, symbolic video representations and large language models for procedure planning.

\subsubsection{Anticipating Objects}
\label{sec:anticipation_regions}
To provide assistance to the user at a more granular level it is useful to make future predictions of attended or manipulated physical regions appearing in the egocentric video, such as  objects, scene parts, or object parts. 
In this way it is possible to issue alerts when specific parts of a potentially dangerous object are going to be touched or when the camera wearer is about to interact with the wrong object in a known workflow. 
Current approaches formulate future region prediction as object detection, heat-map prediction, or semantic mask prediction tasks.
Algorithms are usually evaluated using spatial overlapping retrieval metrics such as mean Average Precision.

\paragraph{Seminal works}
\cite{furnari2017next} first introduced the problem of predicting which objects will be interacted with next. 
\cite{zhang2017deep} proposed to predict future gaze, i.e., the spatial location which will be attended to by the user in the future.
\cite{nagarajan2019grounded} investigated the anticipation of object affordances by predicting interaction hotspots in videos. \cite{liu2020forecasting}~showed how predicting interaction hotspots and future hands trajectories can support more abstract tasks such as action anticipation. 
Notably, these works have addressed their own versions of object anticipation tasks. \cite{grauman2022ego4d} worked toward standardisation of this task, termed short-term object-interaction anticipation. This predicts which of the objects in the scene will be interacted with by the camera wearer (noun of the future object), how the interaction will take place (verb denoting the interaction), and when the interaction will begin (time-to-contact in seconds).

\paragraph{State-of-the-art papers}
Among the anticipation algorithms based on regions, most state-of-the-art approaches mainly focus on the prediction of future interacted objects and in particular on the short-term object interaction anticipation task as defined in~\cite{grauman2022ego4d}.
\cite{pasca2023summarize} currently achieve state-of-the-art performance on the task with TransFusion, a multimodal transformer-based architecture that exploits language. In particular, TransFusion leverages pretrained image captioning and vision-language models to extract the action context from past video frames. This, together with the next video frame, is processed by the multi-modal fusion module to forecast the next object interaction.
Recently, \cite{lai2023listen} proposed a state-of-the-art approach to future gaze prediction in social scenes based on the analysis of audio and video. 
It models audio-video correlations with a spatial fusion and a temporal fusion branch, guided by a multi-modal contrastive loss. 
Fused embeddings are decoded jointly to predict future gaze.

\subsubsection{Anticipating Trajectories}
\label{sec:anticipation_trajectories}
Systems able to make future predictions in the form of trajectories will know in advance where the user may go, how the observed objects will move in the scene, and how the camera wearer's hands are going to move in the near future. Such information is crucial for all those applications that need to plan in advance, e.g., to suggest alternate routes (to avoid passing through dangerous zones) or detect unsafe operations involving the interaction between hands and objects. The metric most commonly used in trajectory prediction is the final displacement error (FDE), defined as the L2 distance between the predicted location and the ground truth.

\paragraph{Seminal works}
\cite{park2016egocentric} first proposed the task of predicting the possible trajectories that the camera wearer may follow from egocentric video. In a complementary way, \cite{yagi2018future} studied the problem of predicting the future trajectory of other persons observed from the egocentric point of view. \cite{liu2020forecasting} investigated how predicting hands trajectories can be beneficial for action anticipation. 
\cite{jia2022generative} proposed the task of anticipating a time series of future hand masks from an egocentric video.
\cite{bao2023uncertainty} explored the problem of predicting future hand trajectories in 3D with the aim to support the understanding of human intention and behaviour in AR/VR applications.
While trajectory prediction tasks have not been systematically studied in the egocentric perspective, a first attempt to propose standard tasks has been done by \cite{grauman2022ego4d}, where two tasks related to the prediction of future locomotion and hands trajectories are formulated.

\paragraph{State-of-the-art papers}
\cite{alikadic2022transformer} presented a new method that leverages transformers to forecast future trajectories of pedestrians from egocentric views. The model predicts the trajectories by relying on previous locations and scales, dynamics poses, and ego-motions of the camera wearer. \cite{kai2023future} designed a multi-channel tensor to represent social interaction, including pedestrian pose, depth and their relative locations. They fed this input to a novel end-to-end fully convolutional transformer (Conv-Transformer) network. 
\cite{hatano2023trajectory} recently proposed an approach that uses semantic information to connect bird’s-eye coordinates to the egocentric viewpoint.
This allows to utilise existing third-person view methods on the egocentric view, without the need to re-train.

\paragraph{Datasets}
Action anticipation and future region prediction works have often relied on action recognition datasets, namely ADL \citep{pirsiavash2012detecting}, EPIC-KITCHENS \citep{Damen2021rescaling}, EGTEA Gaze+ \citep{li2021eye}, and Ego4D \citep{grauman2022ego4d}. 
These general-purpose datasets are described in Sec~\ref{sec:datasets}. 
Since most action recognition datasets, such as ADL, EPIC-KITCHENS and EGTEA, do not  contain significant human locomotion, early trajectory prediction works have performed evaluations on specific datasets collected on purpose. 
In particular, \cite{park2016egocentric} collected the EgoMotion dataset, a set of egocentric videos acquired in various indoor and outdoor scenes using first-person GoPro Hero 3 stereo cameras. The dataset comprises 26 scenes, 65.5k frames and 9.1 hours of video covering various activities such as walking, shopping, and social interactions. 
The First-Person Locomotion Dataset proposed by \cite{yagi2018future} comprises about 4.5 hours of egocentric videos recorded by people wearing a chest-mounted camera and walking around in diverse environments. 
The Ego4D~\citep{grauman2022ego4d} is the first to offer videos suitable for anticipating actions, objects and trajectories. Limited annotations on this massive-scale dataset enable works on anticipating actions, forecasting hand and full-body trajectories. 

\paragraph{For the future}
Despite the progress of research in this field, at the moment of writing, future anticipation approaches achieve limited performance. For example, the current state-of-the-art approach for action anticipation by \cite{roy2022interaction} only achieves a mean top-5 per class recall of $18.1\%$ on the test set of the EPIC-KITCHENS-100 dataset. Similarly, the best performing approach to short-term object interaction anticipation \citep{pasca2023summarize} achieves a top-5 mAP of $24.7\%$ in next-active object prediction and $3.4\%$ when also predicting the interaction verb and time-to-contact on the test set of Ego4D.
These results highlight the very complex nature of anticipation tasks and the need for advances in this area. Current anticipation approaches also suffer from major limitations which prevent their widespread adoption.
Most approaches assume that a ``trimmed'' video is sampled at a fixed time before the beginning of the action and fed to the model, which constitutes an unrealistic scenario, given that the occurrence of future actions is unknown at test time.
Despite some recent work towards an untrimmed anticipation scenario~\citep{rodin2022untrimmed}, the trimmed setting remains the most common one.

\subsection{Gaze Understanding and Prediction}
\label{sec:gaze}
Understanding and predicting which areas of the scene the camera wearer is attending is critical for AR, assistive technologies, and human behaviour analysis. This task involves developing sophisticated algorithms that can accurately estimate the direction of a person's gaze based on the visual information captured by an eye-mounted egocentric camera. Gaze understanding enables more accurate modelling of the camera wearer's visual attention, which offers useful insights for downstream tasks including predicting which object(s) the camera wearer is focused on at a time.
Methods are evaluated by their ability to produce attention maps coherent with ground truth gaze measurements. 
We will focus on methods estimating gaze from the perspective of the beholder. For approaches that utilise remote eye trackers to process frames containing the wearer's face, we refer to the work by \cite{cazzato2020look}.

\paragraph{Seminal works}
Predicting gaze is inherently challenging, leading initial works to explore different cues to understand the wearer's focus of attention. For instance, \cite{yamada2011can} highlighted the challenges associated with using visual saliency maps based on colour, intensity, and orientation in egocentric vision, especially when there is significant egomotion. To address this issue, \cite{yamada2012attention} combined visual saliency maps with rotation- and translation-based attention maps obtained through egomotion estimation. By doing so, they aimed for more robust and accurate egocentric visual attention predictions.

The first approach for predicting gaze from egocentric videos was presented by \cite{fathi2012learning} who simultaneously tackled daily activities recognition and gaze location prediction using a common probabilistic generative model. \cite{li2013learning} leveraged implicit cues in egocentric videos, such as hand location, pose, and motion, to predict gaze. Additionally, they modelled gaze behaviour to enhance prediction performance. Building on this work, \cite{huang2018predicting} modelled patterns in the temporal shift of gaze fixation. Their work is based on the assumption that during fixation, the gaze tends to be located on the same object, and patterns of gaze shift depend on the high-level task, which can be learned.

A more recent study by \cite{al2019ogaze} predicted gaze based on the objects framed in the video. They used features and bounding boxes from an object detection model and combined both classic gaze point regression formulation and classification for prediction. In parallel, \cite{tavakoli2019digging} analysed both top-down and bottom-up factors influencing egocentric gaze prediction. Their work confirmed the relevance of the manipulation point over hand regions and the importance of hand-object interaction for gaze prediction. 

An innovative approach was introduced by \cite{thakur2021predicting}, where the information from the video stream was combined with head movement obtained from IMU (Inertial Measurement Unit) data to improve gaze estimation. On a slightly different note, \cite{su2016detecting} focused on understanding \textit{when} engagement with the environment happens instead of \textit{what} the wearer is looking at. This perspective aimed to detect moments of interaction from egocentric videos.

\paragraph{State-of-the-art papers}
The current state-of-the-art method for gaze estimation on both the EGTEA Gaze+ dataset~\citep{li2021eye} and Ego4D dataset~\citep{grauman2022ego4d} is the approach proposed by \cite{lai2022eye}. The authors tackled the challenge of integrating different gaze cues, such as the likelihood of scene objects to be targets, their location, and the head motion pattern related to gaze shifts, into a comprehensive analysis of visual attention.
To achieve this, they developed a transformer-based model that captures the connection between the global scene context and local visual cues using a Global-Local Correlation module. By combining these various gaze cues and context information, their method achieves state-of-the-art performance in predicting gaze in egocentric videos on both datasets.

Recent devices such as the Meta Aria glasses have an onboard estimation of gaze through sensors embedded with eye-tracking. Their modern eye trackers use corneal reflection, a method involving near-infrared light to illuminate the eyes, causing a reflection that is detected by a high resolution camera. 

\paragraph{Datasets}
Most of the datasets used in this section are general datasets described in Section \ref{sec:datasets} with gaze tracking data to serve as ground truth -- GTEA Gaze dataset by~\cite{fathi2012learning}, EGTEA Gaze+ by~\cite{li2021eye}, GTEA-sub by~\cite{huang2018predicting} and Ego4D by~\cite{grauman2022ego4d}. 
In addition, other datasets have been used in research for specific purposes, but are not publicly available. 
Among the limited public datasets, the Object Search Tasks (OST) Dataset by~\cite{zhang2017deep} includes 57 sequences of search and retrieval tasks performed by 55 subjects provided with eye-tracking data. 

\paragraph{For the future}
The future of gaze prediction in wearable devices holds immense potential to revolutionise human-computer interaction and user experience. As wearable technology becomes more advanced and pervasive, integrating accurate and real-time gaze tracking capabilities will enable seamless interactions with digital content. Wearable devices with built-in gaze prediction algorithms could offer intuitive and hands-free control, improving accessibility and usability across applications. With precise gaze tracking, wearable devices can adapt their interfaces dynamically, presenting relevant information based on the user's visual focus. Despite the advances in this area, gaze analysis still poses various challenges, including the need for large annotated datasets, subjective bias due to individual differences, handling eye blinks and data attributes like occlusion and illumination \citep{ghosh2021automatic,pathirana2022eye}. Nonetheless, even with the application of state-of-the-art techniques \citep{lai2022eye}, results are not ideal, as evidenced by the F1 scores of 44.8 and 43.1 on EGTEA Gaze+ \citep{li2021eye} and Ego4D \citep{grauman2022ego4d}, respectively.
\subsection{Social Behaviour Understanding}
\label{sec:social}
The wearable devices of the future will be able to support users in a variety of scenarios related to their daily lives. As humans are by nature social animals, we expect wearable systems to be able to understand the social behaviours of the camera wearers and of others they engage with. 
The research community has investigated this area with different topics. 
We organise our literature overview by considering the works related to 
\textit{understanding the relationship with the speaker} (Section~\ref{sec:social1}), \textit{detecting/modelling social interactions} (Section~\ref{sec:social2}), \textit{estimating attention towards the camera wearer} (Section~\ref{sec:social3}) and \textit{joint attention} (Section~\ref{sec:social4}). We discuss the relevant publications in each sub-area and provide insights into what future directions may benefit the community with respect to this topic.
It should be noted that the approaches discussed in this section aim to analyse social behaviour from the point of view of the camera wearer, which involves specific challenges and opportunities as compared to approaches based on fixed cameras. Indeed, analysing social interactions from egocentric vision gives a privileged views into behaviours directed towards the camera wearer such as facial expressions, speaking acts, and eye contact. Estimating visual attention through two synchronised wearable cameras further enables the study of joint attention which can have useful applications, including in diagnosing and monitoring social-related health conditions.

\subsubsection{Modelling the relationships with the speakers}
\label{sec:social1}
This area focuses on modelling the relationship between the camera wearer and speaking subjects appearing in the egocentric field of view. Previous works have addressed different objectives that improve the camera wearer's audio-visual interactions, including improving speech quality in a noisy environment, determining auditory attention towards one of a set of speakers, determining which subjects are talking to the camera wearer, and detecting speakers and transcribing their speech. 
Existing approaches have considered an array of similar, yet distinct, tasks related to this area, generally proposing approaches based on the processing of both audio and video. Segmentation maps or bounding boxes are usually produced to spatially detect the speaker. The evaluation is carried out by comparing the predicted areas with ground truth annotations.

\paragraph{Seminal works} \cite{kumano2015automatic} first considered the use of egocentric vision to perform automatic conversation analysis. The work targeted a multi-party conversational scenario, where participants were equipped with in- and out-cameras with microphones and the gaze behaviour of each interlocutor was hence estimated via self-calibration. \cite{donley2021easycom} proposed the task of enhancing a target speech source and speech intelligibility in conversations held in noisy environments and recorded through egocentric devices.

\paragraph{State-of-the-art papers}
\cite{lu2022sound} presented a study to assess whether head angle estimated via egocentric devices is predictive for sound source selection. \cite{jiang2022egocentric} tackled the problem of active speaker detection by using both video and multi-channel microphone array audio. \cite{grauman2022ego4d} presented the ``Social Interactions'' benchmark, which includes tasks aimed at identifying communicative acts directed towards the camera-wearer. 
The most relevant for this section is the ``Talking To Me" task, which focuses on classifying whether each visible face, based on a video and audio segment with tracked faces, is talking to the camera-wearer. Additionally, the researchers introduced the AV diarisation benchmark, to understand the camera-wearer's ongoing interactions with people starting from speech. Those are: localisation and tracking of the participants, active speaker detection, diarisation of each speaker’s speech activity, and transcription of each speaker’s speech content. For the latter, \cite{gabeur2022avatar} recently proposed a new model for audio-visual automatic speech recognition based on a multi-modal audio-visual transformer trained end-to-end from spectrograms and RGB frames. 
\cite{ryan2023egocentric} proposed a novel task of egocentric auditory attention localisation, which identifies the person the camera wearer is talking to in a multi-people multi-conversation scenario. The task is carried out by considering audio-visual signals. The auditory signals are given by a directional array of microphones, while the visual signals are given by egocentric video.

\subsubsection{Detecting and modelling social interactions}
\label{sec:social2}
Works in this area detect the presence of social relationships from egocentric images or video and potentially characterise such relationships, by highlighting the engaged subjects and classifying the behaviour (dialogue, monologue, discussion, etc.). 
Being able to detect and characterise social relationships can allow wearable systems to gain an understanding of the social context of the camera wearer, consider video segments relevant for later recollection (e.g., record important conversations), and track the camera wearer's social relationship for monitoring and diagnosis of potential disorders.
The approaches discussed below did not follow a common task definition, instead they analysed related problems which were tackled with disparate techniques. 

\paragraph{Seminal works} \cite{fathi2012social} proposed the first work to detect and categorise social interactions in egocentric video among a group of individuals. The location and orientation of each subject's face were used to compute a line of sight and obtain a location in space indicating the focus of attention. Head movements of egocentric cameras were also used for a better understanding of attentional focus. 
\cite{narayan2014action} evaluated the performances of dense trajectories to recognise social interactions performed by the camera wearer and other subjects acquired from the egocentric point of view.
\cite{alletto2015understanding} addressed the problem of partitioning people in an egocentric video sequence into socially related groups. Interactions are then detected with clustering and structural learning. \cite{bambach2015lending} focused on hands to detect interactions, and investigated the tasks of hand detection, disambiguation, and segmentation from videos of interacting people using appearance models based on CNNs. 
\cite{yonetani2016recognizing} considered the problem of modelling dyadic interactions (interactions between two people) from paired videos, where micro-level actions and reactions such as slight shifts in attention, subtle nodding or small hand actions are detected. 
\cite{yang2016wearable} proposed the concept of ``wearable social camera'', a camera that summarises the video of the user’s social activities. To achieve this goal, common features among different social interactions, called interaction features, are extracted and processed. 

\cite{su2016social} presented a method to predict future movements of basketball players based on the analysis of social behaviours. 3D reconstruction of multiple first-person cameras and gaze information were used to automatically annotate each player's video the visual semantics. A Siamese neural network was later trained to retrieve future trajectories based on group movements. 
\cite{aghaei2017social} considered the problem of social style characterisation from egocentric photostreams. This is done by detecting temporal segments characterised by social interactions, detecting faces, extracting social signals and classifying the social interaction into formal or informal. 
\cite{duarte2018action} investigated the non-verbal visual cues to ``read the intention'' of other humans in social interactions from egocentric videos. 
Other works focused on robot-centric activity recognition \citep{ryoo2013first,xia2015robot,gori2016multitype}, where the goal is to enable an observer (e.g., a robot or a wearable camera) to understand what activity others are performing towards it. \cite{xia2015robot} proposed to extract features from an ego-motion region and an independent motion region separately and combine the descriptors using multiple kernels. 
\cite{gori2016multitype} proposed a unified mid-level descriptor capable of discriminating between different types of activities. They called it Relation History Image (RHI), and it is built as the variation over time of relational information between every pair of local regions (joints or image patches) belonging to one or a pair of subjects.

More recently, \cite{bertasius2017using} proposed to predict cooperation patterns in the near future without requiring manually labelled intention labels. To do that, they modified the output of a pretrained pose estimation network to represent the camera wearer's internal state, including visual attention and intentions. They then employed this transformed output as a supervisory signal to train another network for the cooperative basketball intention task. 

\paragraph{State-of-the-art papers} 
Due to limited datasets on the topic, there are only a handful of recent works proposing methods that could be deemed state-of-the-art on this task.
\cite{li2019deepdual} addressed the problem of modelling dyadic interactions by explicitly modelling the relations between the interacting subject and the camera wearer using a dual recurrent network. That incorporates two interconnected sub-tasks, namely individual action representation learning and dual relation modelling.  
\cite{lai2023werewolf} recently proposed the task of modelling persuasive behaviours during multi-player social deduction games leveraging language models. Given an utterance and its corresponding video segment, they seek to predict the persuasion strategies
adopted in the utterance. They first leverage a pretrained language model as the text encoder to obtain the utterance embedding, and a vision transformer to obtain the visual embedding. They then concatenate the textual and visual features to
predict the persuasion strategy. 

\subsubsection{Estimating attention towards the camera wearer}
\label{sec:social3}
This line of work focuses on understanding when a subject appearing in the egocentric field of view is attending to (e.g., by looking at or talking to) the camera wearer. This ability can allow egocentric vision systems to facilitate social interactions (e.g., by notifying the camera wearer when a person is trying to make contact), improving diagnosis of potential disorders of the observed subjects by studying their attentional pattern towards the camera wearer (e.g., Autism Spectrum Disorder when the camera wearer is a doctor), and logging conversations with the camera wearer for later recollection. Current approaches have analysed different tasks, ranging from detecting eye contact to detecting people looking at or talking to the camera wearer. 

Methods are evaluated by validating predictions against manually annotated ground truth or comparing algorithms' performance to human performance. Similar to other tasks, the evaluation has not yet been investigated in a systematic way. 

\paragraph{Seminal works}
\cite{ye2012detecting} developed the first system capable of recognising eye contact between the wearer and a person in front of the camera. The system aimed to detect atypical patterns of gaze and eye contact in children to understand early signs of autism. The approach used face detection to find the child's face and then estimated its gaze in 3D space to determine whether it pointed towards the wearer. Another autism diagnosis system was developed by \cite{petric2014four}, leveraging the idea that autistic children tend to interact more with technological devices than with humans. They utilised robots equipped with frontal cameras to understand joint attention patterns. Subsequently, \cite{smith2013gaze} focused simply on capturing whether people in images were looking towards the wearer or not, tackling what they called ``gaze locking''. By discretising the problem and avoiding continuous tracking of the observer, they simplified the task. To address the challenges of strong appearance diversity in human eyes, \cite{ye2015detecting} proposed a model that couples eye appearance with head pose for improving eye contact detection.

\paragraph{State-of-the-art papers}
The current state-of-the-art approach in eye contact detection, as proposed by \cite{chong2020detection}, scales the training dataset to 4.7 million human-annotated eye contact images and leverages large-scale datasets from other tasks like face recognition to build initial representations that understand the relationship between head pose and gaze direction.

Recently, \cite{grauman2022ego4d} introduced the ``Looking At Me'' task, which involves classifying whether each visible face in a video, with localised and identified social partners, is looking at the camera-wearer. \cite{xue2023egocentric} takes a unified approach by building on the idea that various video understanding tasks are related. They propose EgoT2, a framework that combines different task-specific models to improve performance. They also integrate tasks like ``Talking To Me'' and ``Active Speaker Detection'' to understand if a specific person is talking to the wearer and who is speaking, respectively.

\subsubsection{Estimating joint attention}
\label{sec:social4}
Works in this area focus on modelling the joint attention of multiple subjects towards scene regions, objects, or people. The ability to estimate joint attention can allow egocentric systems to monitor and improve social interactions (e.g., by monitoring the joint focus of attention and notifying the camera wearer when it changes), improve diagnosis of potential disorders, and enhancing video curation and summarisation by detecting the most popular scenes from a set of synchronised egocentric video streams. Modelling is usually performed by predicting social saliency maps, detecting jointly attended objects, or determining a subset of subjects with coherent attention patterns. Approaches are evaluated by comparing predictions against manually annotated ground truth labels or assessing how the estimated joint attention is predictive of other subjects' attention. Previous works have considered different but related task formulations. 

\paragraph{Seminal works}
Perceiving joint attention of wearers towards a common scene was introduced by \cite{park20123d}. They constructed a 3D social saliency field, and located gaze concurrences by localising wearable cameras via structure-from-motion in a common coordinate system and triangulating the attention of each wearer. Subsequently, \cite{park2013predicting} introduced the concept of ``social charges'' as latent quantities driving the attention of people in a social group, defining the relationships between these charges and the primary gaze of each member. They estimated time-variant social saliency fields from observed primary gaze, enabling the prediction of gaze direction at any proximal location or time. Building on previous works, \cite{soo2015social} proposed a method to estimate the likelihood of joint attention as a function of a social formation, without relying on the gaze of group members. 
Using the dataset from \cite{park20123d}, where the locations of group members and their joint attention were measured, their learned representation demonstrated the ability to predict the social saliency of real-world scenes.

Another line of work explores the multiple wearable camera setting to automatically edit footage in a smart manner. For example, \cite{arev2014automatic} used the centre of attention of different cameras as an indicator of what is important in the videos and combined it with cinematographic guidelines to produce effective summaries of the original footage. \cite{hoshen2014wisdom} estimated the cameras looking at the same region, without reconstructing the scene in 3D.

A slightly different approach is presented by \cite{lin2015co}, where the goal is to locate the person who draws attention from most wearers in a multi-camera setting. They used motion patterns to correlate people across videos to avoid appearance-related challenges, such as groups of people dressed similarly. 

\cite{kera2016discovering} authored one of the pioneering works on discovering joint attention based on visual appearance. They proposed to locate objects that are attended by multiple camera wearers. They used multiscale spatiotemporal tubes around points of gaze as potential objects of interest and performed unsupervised clustering on them. 

\paragraph{State-of-the-art papers}
The estimation of joint attention is an emerging topic, and it has mostly seen pioneering works proposing various variations on the task. 
More recently, \cite{huang2020ego} improved on the approach of \cite{kera2016discovering} on locating objects that are attended by multiple camera wearers. They tackled the challenges of cluttered scenes and noisy gaze by first temporally locating joint attention periods and then spatially segmenting the object of joint interest. Those contribute to a more reliable spatial segmentation than simply using regions in proximity to the points of gaze, which might be noisy. They achieve that by means of a hierarchical graphical model composed of multiple linear chain conditional random fields.

\paragraph{Datasets}

Numerous public datasets are now available to evaluate model performance in social behaviour understanding tasks. Among the most widely used is the First Person Social Interaction Dataset (FPSI) \citep{fathi2012social} dataset described in Section \ref{sec:datasets}.
The JPL First-Person Interaction dataset \citep{ryoo2013first} stands out as the first one to annotate actions performed by a robot. It includes 7 actions, comprising 4 friendly interactions, 1 neutral interaction, and 2 hostile interactions, aiming to study robot-centric activity recognition. Similarly, the datasets from \cite{xia2015robot} also address robot-centric activity recognition, leveraging the depth modality in addition to other features.
For dyadic interactions, the Paired Egocentric Video dataset by \cite{yonetani2016recognizing} contains over 1000 pairs of egocentric videos capturing micro-action and reaction patterns from the perspectives of both interacting individuals.
The EGO-GROUP and EGO-HPE datasets \citep{alletto2015understanding} offer videos featuring groups of people, enabling testing of group detection in egocentric vision and head pose estimation of the participants.
The Focused Interaction dataset \citep{bano2018multimodal} provides multimodal representations of individuals interacting, supporting the development of automatic interaction detection. \cite{park20123d} proposed a dataset for evaluating joint attention. Three video sequences were recorded: a meeting with two groups, a musical with alternating performances, and a party with multiple activities. The UTJA-M dataset \citep{huang2020ego} captures moments of joint attention among individuals and releases the tracked gaze for all participants.
Targeting the development of conversational AI, the EgoCom \citep{northcutt2020egocom} and EasyCom \citep{donley2021easycom} datasets offer multimodal recordings, with a primary focus on audio. While EgoCom contains nearly 40 hours of recordings, EasyCom provides synchronised audio from different participants, incorporating realistic acoustic noise into the setting.
\cite{lai2023werewolf} proposed a multi-modal dataset for studying persuasive behaviours during social games. Videos are sourced from both YouTube and the Ego4D social dataset and include text, video, and audio signals. In total, it contains 5,815 utterances from Ego4D and 20,832 utterances from YouTube.

\paragraph{For the future}
As discussed in the previous sections, the literature on social behaviour understanding is less stratified as compared to the other tasks considered in this paper. The current state-of-the-art performance varies significantly depending on the task. For the ``Talking To Me'' task in the Ego4D benchmark \citep{grauman2022ego4d}, the results show 53.9\% mAP and 54.3\% accuracy on the test set, which is still far from human-level performance. Conversely, in the task of eye contact detection has reached acceptable performance -- the state-of-the-art achieves an overall precision of 0.936 and a recall of 0.943 across 18 validation subjects. This performance is comparable to that of 10 trained human coders.
While several datasets have been collected, the focus has been on diagnostic and summarisation tasks. 
The potential for understanding social behaviour can be a game changer in strategic interactions -- whether gaming or even offering advice in live negotiations and group meetings.
Such potential requires interdisciplinary research, beyond computer vision expertise and is currently at its infancy.

\subsection{Full-body Pose Estimation}
\label{sec:fullbody_egopose}
The reconstruction of the wearer's body pose is crucial to enable applications such as daily life monitoring and AR. Consequently, the research community has devoted increasing attention to this field in recent years.
Human pose estimation aims to create representations of the human body either in the local egocentric camera space or in a world coordinate system. Two main approaches are employed for constructing body representations: \textit{kinematic models}, which utilise joint positions and limb orientations without capturing detailed texture and shapes, and \textit{volumetric models}, which provide more realistic representations and capture deformations.
Methods are assessed by their Mean Per Joint Position Error (MPJPE) measuring the average distance between the predicted joints and the ground truth joints.
Works addressing this problem in egocentric vision tend to differ from works focusing on fixed cameras due to the limited field of view of wearable cameras, which rarely captures the full view of the human body.

\paragraph{Seminal works}
Wearable cameras, with their limited field of view primarily focused on the wearer's attention, only capture a partial view of the wearer's body. As a result, significant efforts in egocentric 3D pose estimation have been dedicated to designing systems that can overcome this limitation. One pioneering study conducted by \cite{shiratori2011motion} tackled this challenge by employing a Structure-from-Motion technique. They utilised 16 outward-looking body-mounted cameras to reconstruct both the relative and global joint motion of a person in outdoor environments. The objective of their work was to develop a motion capture system that could operate effectively ``in the wild''.

Drawing inspiration from the concept of overcoming the wearer's body invisibility, subsequent works in the field have explored the use of cameras positioned to face downwards towards the body. One notable approach in this regard is the development of EgoCap by \cite{rhodin2016egocap}. It involves a specially designed head-mounted stereo rig setup with downward-facing cameras. Expanding on the downward-facing camera setup, \cite{xu2019mo} proposed the first real-time motion capture system utilising a single monocular fisheye camera mounted on a cap. 
Finally, addressing the specific setting of head-mounted displays in AR/VR, \cite{tome2019xr} positioned the camera on the rim of a VR headset and generated a photorealistic dataset which served as a valuable resource for research.

In the study conducted by \cite{wang2021estimating}, a significant focus was placed on addressing the limitations associated with adopting a local egocentric camera reference system, particularly in applications such as animating body locomotion in a virtual environment. Recognising this restriction, the researchers proposed a novel framework that combines the local pose estimation with the world coordinate system obtained through SLAM. The aim was to achieve a temporally stable integration of both perspectives, enabling more robust and accurate results.

In parallel, chest- and head-mounted outward-looking cameras have also been employed to infer the pose of the wearer in more challenging scenarios where most of the body is out of the camera's field of view. In particular, \cite{rogez2015first} extended the estimation of human body part joints from hands to the entire upper limb using synthetic depth data training. \cite{jiang2017seeing} went further by attempting to estimate the full-body of the camera wearer from a single outward-looking camera, leveraging dynamic motion signatures and static scene structure to infer the ``invisible'' human body pose.

Unlike previous approaches focused solely on smooth and accurate poses, \cite{yuan20183d} introduced a method that formulates body pose estimation as a Markov decision process adopting a dynamics-based perspective. They leveraged a physics simulator to train a policy that generates physically plausible poses. In a subsequent study, \cite{yuan2019ego} improved upon this approach by adopting a control-based methodology that not only estimates poses but also forecasts valid future poses, going beyond pure estimation.
In a similar vein, \cite{luo2021dynamics} employ a combination of kinematics- and dynamics-based modelling to achieve the first-ever estimation of physically plausible 3D human-object interactions. The authors collected their own dataset, which include 6 degrees of freedom (DoF) object poses. These object poses are factorised within the scene and subsequently utilised by their method to estimate realistic 3D human-object interactions, accounting for the physical constraints and dynamics involved. 

In embodied AI, understanding social interactions holds great significance, leading to a focus on pose estimation tasks during social interactions. \cite{ng2020you2me} introduced a method called ``You2Me'' that utilises the action-reaction social interaction dynamics between the wearer and a second person as prior to estimate the wearer's pose. This work emphasises the influence of inherent synchronisation during interactions. On the other hand, \cite{liu20214d} made a significant contribution by being the first to attempt the estimation of a second person's pose from an egocentric perspective while simultaneously grounding it in the given 3D environment.

\paragraph{State-of-the-art papers}
The current state-of-the-art performance for downward-looking fisheye datasets, such as Mo2Cap2 \citep{xu2019mo} and the dataset introduced by \cite{wang2022estimating}, is achieved by the method proposed in \cite{wang2023scene}. It integrates scene constraints into pose prediction to avoid obtaining physically unrealistic poses like body floating or penetration with the environment. The approach consists of two primary steps. Firstly, the depth modality is inferred to capture the spatial information of the scene. Secondly, the inferred depth is inpainted in areas of the image where the body occludes the scene. The inpainted depth is then combined with 2D pose features in a shared 3D voxel space. Integrating scene constraints in this common 3D voxel space allows for pose estimation while enforcing adherence to the physical constraints imposed by the scene.

At the same time, the work presented in \cite{li2023ego} achieves the best performance on a set of egocentric datasets \citep{luo2021dynamics, zheng2022gimo} captured from outward-looking camera perspective, including their proposed synthetic egocentric dataset. Given that directly matching egocentric video with full-body pose is challenging due to the frequent absence of visible body parts, the authors address the task by introducing an intermediate step of head motion estimation. This approach eliminates the requirement for a training dataset of paired egocentric video and 3D human motion, while accurately predicting head motion using SLAM and a transformer-based model. Subsequently, a diffusion model conditioned on the estimated head pose is employed to derive the full-body pose. However, in this work, evaluation sequences only contain people navigating a virtual scene, and are not undergoing any activities. 

Recent research has also placed significant emphasis on the task of estimating the complete body pose of individuals within the recorder's field of view. In the study by \cite{ye2023decoupling}, the focus is on simultaneous localisation and human mesh recovery to reconstruct the global poses of individuals featured in egocentric videos, all without relying on dense 3D reconstructions of the surroundings. The proposed method, SLAHMR, firstly predicts relative camera motions, identifies individuals, and determines their local 3D poses. Leveraging this information, the model initialises trajectories for both humans and cameras within a common world reference system, optimising them for consistency across 2D observations in the video and learned human motion priors. \cite{zhang2023probabilistic} takes a different approach by incorporating the 3D scene and conditioning a diffusion model for human pose generation on it. The authors combine human-centric scene regions with a physics-based collision score to guide the generation of plausible human poses that avoid environment penetration. To further enhance the accuracy and diversity of poses, they employ a visibility-aware graph convolution model, enabling the learning of precise body poses for visible joints while encouraging diversity in truncated parts.

\paragraph{Datasets}
Current datasets for pose estimation from downward-facing camera systems range from simulated \citep{tome2019xr} to real-world datasets \citep{rhodin2016egocap, xu2019mo, wang2021estimating}. These datasets provide ground truth in the form of 3D poses of the camera wearer. Recently, there has been a growing interest in generating large-scale real-world datasets, such as EgoPW \citep{wang2022estimating}, as well as simulated datasets with a diverse range of motions, such as UnrealEgo \citep{akada2022unrealego}.

In the context of outward-looking camera setups, interest has also been increasing. In addition to the previously mentioned works \citep{luo2021dynamics, li2023ego, ng2020you2me}, datasets with orthogonal characteristics continue to be released. For example, the EgoBody dataset by \citep{zhang2022egobody} plays a crucial role in modelling interactions, as it encompasses multi-modal egocentric data streams and provides 3D ground truth for multiple individuals in complex 3D scenes. Furthermore, \cite{zheng2022gimo} recently introduced a large-scale dataset for human motion prediction that includes gaze information. They argue that accurate motion prediction depends on understanding human intentions, which can be studied using gaze in the egocentric setting. Additionally, the EgoHumans benchmark by \cite{khirodkar2023egohumans} captures multiple subjects in realistic outdoor environments from multiple egocentric viewpoints, serving as a valuable resource for multi-view multi-human analysis.

\paragraph{For the future}
Although some works \citep{yuan2019ego,xu2019mo} have prototyped real-time egocentric body pose estimation, performance is significantly below that of third-person (or remote) cameras. Body-oriented camera methods \citep{wang2023scene} perform slightly better than those with outward-looking camera \citep{li2023ego} obtaining a MPJPE of 118.5 millimeters (mm) against MPJPEs ranging from 121.1 to 152.1 mm despite being tested on different datasets. Both still exhibit MPJPE values which are far from recent ones \citep{tang20233d} obtained on the Humans3.6M third-person benchmark \citep{ionescu2013human3} ranging around 20mm. Even in the context of estimating the poses of other individuals within the camera wearer's field of view, the current state-of-the-art \citep{ye2023decoupling} achieves a World PA First - MPJPE, i.e. an MPJPE obtained by aligning the first frame of the prediction with the ground truth -- of 141.1 mm on EgoBody \citep{zhang2022egobody}. This performance remains notably distant from the results achieved with third-person perspectives. The recent release of more realistic datasets \citep{zheng2022gimo, khirodkar2023egohumans} can assist in bridging the gap between research and practical solutions.

Importantly, full-body estimation during natural activities, beyond navigation and full-body motion like jumping and squatting, is yet to be explored.
For example, consider a person knealing to retrieve an object from a cupboard, where their hand is occluded by the cupboard itself. 
Such poses are not available in any full-body datasets currently available.
To date, the task of full-body estimation is not integrated with other tasks such as action understanding, trajectory forecasting, and hand-object estimation. 
It is thus difficult to assess the usefulness of current techniques in isolation.
\subsection{Hand and Hand-Object Interactions}
\label{sec:hand-object}
The significant presence of hands in egocentric videos and their primary importance in understanding humans' behaviour in an environment have lead to a proliferation of research on hands and their interaction with objects.
While other research lines focus on human-object understanding from fixed cameras, egocentric vision provides a more fine-grained view into object interactions in which hands are central. As a result, methods for hand-object interaction understanding from egocentric vision greatly differ from human-object interaction detection from fixed cameras.
In the next sections, we review works that focus on estimating \textit{hand pose} (Section~\ref{sec:hand1}) and classifying \textit{hand gestures} (Section~\ref{sec:hand-gestures}).
Additionally, we analyse works that deal with hand-object interaction aiming at understanding how the camera wearer engages with the surrounding environment and the objects present therein.
We divide hand-object interaction methods into those that estimate \textit{2D information} (Section~\ref{sec:hand3}) vs others which exploit \textit{3D meshes of hands and objects} (Section~\ref{sec:hand4}).

\subsubsection{Hand Pose Estimation}
\label{sec:hand1}
Predicting the pose of hands from an egocentric viewpoint, especially during human activities, is a challenging task due to severe occlusions caused by object manipulations, limited field of view and head motion. 
The goal of hand pose estimation approaches is to efficiently regress 3D hand keypoints from various input signals such as RGB images, videos, depth maps, or 3D meshes.
To evaluate the quality of the predicted hand pose, evaluation measures focus on the mean error for each hand joint or re-projection errors in meshes.

\paragraph{Seminal works}
First works on hand pose estimation exploited both RGB and depth signals, thanks to the availability of RGB-D sensors like Kinect.
\cite{Oikonomidis2011EfficientM3} were the first to study the problem without requiring special markers and a complex hardware setup.  
\cite{hand_pose_rogez15} analysed hands performing daily activities from the egocentric point of view, predicting their poses through a tracking-by-detection framework.
\cite{Qian2014RealtimeAR}  introduced the first real-time system capable of accurately tracking a fully articulated hand. 
\cite{Cem_handpose_hand_class} used the depth sensors to address two tasks simultaneously: hand pose estimation and hand shape classification.
\cite{Sabater2021DomainAV} proposed a novel skeleton-based approach which is robust for predicting hand actions in different domains.
Pose features are estimated using a temporal convolutional network, and aggregated to predict hand actions.
\cite{Tang2013RealTimeAH} were the first to explore the use of synthetic data to address the articulated hand pose estimation problem in a semi-supervised manner.
They aimed at minimising the synthetic-to-real domain shift by leveraging a large synthetic dataset and a small amount of labeled real data. Synthetic data have been also used to predict the 3D pose of hands by \cite{Liu2021SemiSupervised3H}. They proposed  a unified approach which uses labelled synthetic and unlabelled real videos for joint 3D hand and object pose estimation. 
Similarly, \cite{Mueller2017GANeratedHF} used synthetic data for real-time 3D hand tracking for estimating hand poses. 

\paragraph{State-of-the-art papers}
Recently, different works focused on the optimisation and refinement of 3D hand pose estimation methods. 
\cite{Cheng_2021_ICCV} introduced HandFoldingNet, a network designed for estimating 3D hand joint coordinates from an input hand point cloud. The optimisation process is achieved through a guided folding step, which computes the 3D pose by leveraging a 2D hand skeleton. The folding step is further guided by multiscale features, representing both global and local information. \cite{Yang_2022_WACV} presented a shallow deep neural network that incorporates specific layers capable of iteratively refining the predicted hand pose.
Hand pose estimation has expanded beyond the use of depth maps and RGB signals. \cite{rudnev2021eventhands} were the first to address this task using an event-based camera. They proposed EventHands, an approach which regresses 3D hand poses exploiting locally-normalised event surfaces, which is a new way of accumulating events over temporal windows.
Of these works, only \cite{Yang_2022_WACV} evaluated their method on egocentric hand pose, though the method was tested for general views.

Several works have leveraged hand pose estimation to perform action classification \citep{wen2023hierarchical} and hand reconstruction through neural representation~\citep{Karunratanakul_2023_CVPR, Lee_2023_CVPR}.
\cite{wen2023hierarchical} built a framework which exploits the relationship between frames and the hand poses in an end-to-end manner. Given an egocentric video, a feature extractor encodes spatial information for each frame. Sequences of per-frame features are then fed to a hierarchical temporal transformer to capture temporal information. This transformer is composed of two parts, one for predicting the 3D hand pose and the other one for estimating the action. 

\cite{Karunratanakul_2023_CVPR} presented an approach named HARP (HAnd Reconstruction and Personalisation) designed to create personalised hand avatars from short monocular RGB videos.  They proposed a method to estimate a coarse hand pose and shape and then optimise the hand mesh, the albedo and the normal map using an analysis-by-synthesis strategy that compares the input image to the reconstructed ones. The HARP representation not only enhances the quality of 3D hand pose estimation but also allows for synthesising hand poses from new viewpoints.
\cite{Lee_2023_CVPR} proposed the first neural implicit representation of two interacting hands, called Im2Hands. This approach enables the reconstruction of two interacting hands regardless of their resolution and geometry. It achieves this through two novel attention-based modules: one for initial occupancy estimation and the other for context-aware occupancy refinement.

Recently, \cite{Tse_2023_ICCV} presented a novel transformer-based approach that exploits multi-view RGB images to reconstruct two hands meshes directly. In particular, the proposed approach is able to reconstruct hands avoiding the use of deep network to regress hand model parameters. A larger dataset was used in~\cite{pavlakos2023reconstructing} allowing an improved transformer learning for 3D hand mesh estimation. Their annotations include 5.3K egocentric images from EPIC-KITCHENS VISOR~\citep{VISOR2022} and 23.2K images from Ego4D~\citep{grauman2022ego4d}.

\subsubsection{Hand Gestures}
\label{sec:hand-gestures}
Hand gestures provide key information to enable human-computer interaction for AR/VR helmets, glasses and robots. Hands can be conveniently captured by wearable devices which are equipped with cameras able to observe the scene from the first-person view.
While hand pose estimation and gesture recognition have been traditionally treated as separate tasks, they are inherently related. Recognising hand gestures can be seen as a discrete version of hand pose estimation focusing on understanding the semantics of gestures. Methods are usually evaluated with standard classification measures.

\paragraph{Seminal works}
The interpretation of hand gestures for human-computer interaction has been a topic of research for a while \citep{pavlovic1997visual}.
A pioneer work on hand gesture recognition in the context of egocentric vision has been presented by \cite{gestures_baraldi14}. Inspired by dense trajectories approaches introduced for action recognition, they proposed to extract dense features around regions selected by a designed hand segmentation method, enhancing temporal and spatial coherence.
In addition to RGB, other signals have been also used such as depth, skeleton information or stereo-IR.
\cite{skeleton_gesture_Smedt} exploited time series of 3D hand skeleton to extract an informative descriptor for gesture classification, which is commonly employed for interacting with devices, such as \textit{pinch, swipe right, swipe left and tap}. 
\cite{MolchanovYGKTK16} classified hand gestures considering depth, RGB and stereo-IR data streams through a recurrent 3D-CNN. 3D features have been also exploited by \cite{cao2017egocentric}. They proposed a novel spatiotemporal transformer module to classify gestures from RGB videos without explicitly detecting hands (e.g., hand detection or segmentation) and estimating head motion to rectify deformations.
In the context of human-computer interaction with wearable glasses, 
\cite{Huang_2016_CVPR_Workshops} proposed to study pointing gestures focusing on fingers, based on the observation that pointing gesture and its fingertip trajectory are crucial to recognise hand gestures like \textit{pointing, selecting and writing}.

\paragraph{State-of-the-art papers}
Several works addressed the hand gesture recognition task from the egocentric point of view to enable human-device interaction, especially for AR/VR devices (e.g., smart glasses). 
\cite{Chalasani_gestures_AR_2018} proposed a deep network comprising an encoder responsible for extracting hand features, which are then fed into an LSTM to capture temporal patterns. RGB input sequences can be of an arbitrary length and repetitive gestures.
\cite{Bai_gestures_depth_2018} proposed a method to recognise hand gestures from a single depth camera, which can be integrated into VR/AR applications. They presented a two-stage method. First, they used a CNN to estimate the hand pose from bone lengths and joint locations. Then, they classified the gesture by leveraging hand language. The latter is composed of four basic predicates (\textit{pointing direction, relative location, fingertip touching and finger flexion}) which are applied to the six most important areas of the hand, the 5 fingertips and the palm.

In addition to human-computer interaction, the concept has also been extended to human-robot interaction.  \cite{Papanagiotou_Frontiers_gestures} proposed a multi-task approach including gesture recognition to enable human-robot collaboration on an industrial assembly line. The main component is represented by a gesture recognition module which is based on 3D CNN trained on egocentric data acquired with a GoPro camera.

Some works proposed to use multiple signals to extract richer information.
\cite{Chan2016RecognitionFH} used HandCams (i.e., a wrist-mounted camera) together with a HeadCam. 
By using HandCams, it is not necessary to detect hands and infer manipulation regions as in classic egocentric approaches due to the fact that hands are always in the foreground.
Considering this camera setting, the authors proposed a two-streams deep CNN, with one stream dedicated to the head and the other to the hands, respectively. Extracted features from both streams are then fused through concatenation and used to predict hand states (free vs. active), object categories and hand gestures.
\cite{uni_multimodal_19} proposed a \textit{multimodal-training/unimodal-testing} scheme, which involves sharing the knowledge between individual modality networks (e.g., RGB, Depth and Optical Flow) in the training phase in order to derive a common representation of hand gestures.
To do this, a new spatio-temporal semantic alignment loss has been proposed which is similar to the covariance matrix alignment of the source and target features maps in domain adaptation methods.  At inference time, each network has learned to recognise hand gestures from its specific modality but it also gained the common knowledge from the other networks.

Based on the idea that the background is not relevant for recognising hand gestures in AR/VR applications, \cite{simsegrec_19} focused on hand segmentation to improve gesture recognition accuracy. They proposed a new encoder-decoder architecture capable of generating embeddings from RGB images. These embeddings were then used for hand segmentation and gesture recognition simultaneously through multiple LSTMs. 

\subsubsection{2D Hand-Object Interaction}
\label{sec:hand3}
2D hand-object interaction methods aim to associate each hand with one or more objects present in the scene, thereby determining their relationship (e.g., the hand is holding a plate).
Formally, this task involves detecting and recognising the hands of the user, along with the objects involved in the interaction.
To do so, methods have been developed to predict hand-object interactions by estimating information such as 2D bounding boxes or hand-states (i.e., contact or no-contact). The performance of hand-object approaches is assessed by evaluating their classification and regression abilities. 2D object interaction methods also predict object state changes and object transformations. 

\paragraph{Seminal works}
Relations between objects and tasks are important for critical modelling activities and behaviour. 
\cite{you-do_damen} defined and discovered \textit{Task Relevant Object (TRO)} which refers to an object or a part of it that a human interacts with while performing a specific task, in an unsupervised approach. Crucially, they also aimed to distinguish and classify the different \textit{Modes of Interaction (MOI)} with these TROs.  \cite{Cai2016UnderstandingHM} explored the relations between hands and objects to detect the grasped part of an object during human manipulation. They extracted object attributes such as the \textit{thick or long} shape of a bottle, and observed how these attributes influenced the type of grasp used. 
Their unified model combines the prediction of the object, its attributes, the grasp type, and the action performed.
\cite{Rogez_2015_ICCV} formalised the problem of classifying handled objects using both RGB and depth signals. Depth provides additional information on the exact touch/contact between the hand and the objects present in the environment.
\cite{Object_fluents_Yang} focused on the effect of interactions on objects (e.g., a mug can be empty or full). 

\paragraph{State-of-the-art papers}
\cite{Shan_2020_CVPR} proposed a method to detect and localise hands in the scene, distinguishing between left and right hands. Additionally, they aimed to classify objects into two classes: \textit{active} or \textit{passive}. In particular, if an object present in the scene is in contact with at least one hand, it is considered as \textit{active} object, otherwise, it is considered \textit{passive} object. They also considered 5 different contact states: \textit{no contact, self contact, other person contact, portable object contact} and \textit{stationary object contact (e.g., furniture)}.
While originally designed for YouTube videos, a modified model with additional annotations was successfully used to automatically annotate EPIC-KITCHENS-100~\citep{Damen2021rescaling} with hands and active objects.
\cite{grauman2022ego4d} introduced the task of object state change detection and classification. The task involves distinguishing transformative interactions from those that are purely translational.
Detecting the temporal moment at which an object changes state during transformation is introduced with manual annotations.

A related task to object transformations is tracking objects in egocentric views. \cite{dunnhofer2023visual} analysed the performance of state-of-the-art visual trackers in the egocentric domain highlighting challenges in the ego domain.

Although the analysis of hand-object interactions mostly involves bounding box annotations, a few works have focused on studying hand-object relations using semantic segmentation mask annotations \citep{GonzlezSosa2021RealTE, Fine_grained_HO_Zhang, VISOR2022, tokmakov2023breaking}. These works focus on hands and active objects semantic segmentation considering egocentric images \citep{GonzlezSosa2021RealTE, Fine_grained_HO_Zhang} or videos~\citep{VISOR2022,tokmakov2023breaking}.
\cite{VISOR2022} defined and predicted hand-object relations, including cases where the on-hand glove is in contact with an object in the environment. After segmenting active objects, a binary classifier is used to predict the state of each hand as well as the object-in-contact in each case.

Linking 2D to potential 3D information, by predicting 2D hand-object relations, \cite{qian2023understanding} addressed the task of understanding what a user is able to do (i.e., how can I manipulate the objects in an image?) considering the environment where the user is.
They introduced a transformer-based encoder-decoder which takes in input an image and a set of 2D query points to predict the potential interaction. In particular, for each query point, the transformer head predicts an interaction represented by depth, surface normal of the objects, physical properties and affordance.

\subsubsection{3D Hand-Object Interaction}
\label{sec:hand4}
The task of 3D hand-object interaction predicts 3D information about the hands and objects involved in observed interactions through 3D bounding boxes, 3D meshes as well as 6 DoF hands and objects poses.
Performance is measured by metrics such as 3D mean joint position error for hands, symmetric Chamfer distance for objects, differences in 6 DoF pose including translation and rotation errors as well as re-projection errors including IoU metrics of the 2D re-projections.

\paragraph{Seminal works}
\cite{Tekin2019HOUE} proposed an end-to-end framework to understand 3D human-object interactions from still RGB images. The model takes as input a single RGB image and estimates hand and object poses, recognises objects and predicts the class of the activity.
\cite{GarciaHernando2017FirstPersonHA} used 3D hand poses, 6D object poses and RGB-D images to classify hand actions.
In \cite{Chen2019} hotspots from hand touch are automatically detected and associated with actions from egocentric videos capturing the camera wearer using a sewing machine.
\cite{Hasson2019LearningJR} studied the problem of reconstructing hands and objects during manipulation, in the case the latter is affected by occlusions. They proposed a new architecture composed of two branches, one for the object shape and the second one for the hand mesh.
Differently from previous works which focused on instance-level human-object interactions where 3D models and sizes of objects are known beforehand, \cite{HOI4D_Liu} studied human-object interactions considering the vast diversity of objects in our daily life. They addressed this task by exploiting 4 dimensions of input data: the scene point clouds and object meshes (3D) along the time interval (1D).

\paragraph{State-of-the-art papers}
Few works have recently addressed the 3D hand-object interaction task. 
\cite{chen2023gsdf} introduced a geometry-driven signed distance function (gSDF) method that incorporates robust pose priors, leading to improved hand-object reconstruction by disentangling pose and shape estimation.
\cite{fan2023arctic} proposed two novel tasks based on hand-object interactions: consistent motion reconstruction and interaction field estimation. They also presented two novel approaches to address these tasks. ArcticNet is an encoder-decoder architecture able to reconstruct the motions of both hands and the articulated object, while InterField estimates, for each hand vertex, the distance to the closest object mesh.

Temporal information has also been considered to estimate 3D hand poses and actions \citep{wen2023hierarchical}, and it has a relevant role even in the work of \cite{Hampali2022InHand3O} which proposed a novel method based on UNISURF \citep{Oechsle2021ICCV} to reconstruct 3D objects during hand-object manipulation.
Given a sequence of RGB frames in which a hand is manipulating an unknown object, the method captures both geometrical and appearance features of the object by constructing a neural implicit representation. The latter is then used to reconstruct the object. Differently from other NERF-based methods, the proposed approach assumes that the camera pose is not available.

\paragraph{Datasets}
For \textit{hand pose estimation}, \cite{Yuan2017BigHand22MBH} acquired a large scale dataset named BigHand2.2M, which covers a wide and dense range of hand poses. The dataset contains 2.2 million depth maps annotated with hand joints, utilising six 6D magnetic sensors and inverse kinematics.
Since hand poses become more complicated when involving object interactions, \cite{Ohkawa_2023_CVPR} published the AssemblyHands dataset which includes synchronised egocentric and exocentric images sampled from the Assembly101 dataset \citep{sener2022assembly101} in which users assemble toy vehicles. The dataset is composed of 3.0 million images and has been labelled with high-quality 3D hand poses, using a proposed automatic annotation model that exploits the exocentric view.

Tackling both \textit{hand poses} and \textit{gesture recognition}, \cite{mediapipe_hands_20} acquired a dataset composed of real and synthetic images. 
The data collection has three sets created to address different aspects of the problem: 
1) capturing ``hands in the wild'' with geographical diversity, varying lighting conditions, and diverse hand appearances, 2) covering a wide range of angles representing all physically possible hand gestures, and 3)~incorporating synthetic data to enhance the study of hand poses and gestures. 

Among the large datasets specifically focusing on \textit{hand gestures} \cite{Huang_2016_CVPR_Workshops} proposed EgoFinger which is composed of egocentric videos of different pointing gestures acquired in multiple scenarios. The dataset contains 93.729 RGB frames and it has been collected by 24 subjects in 24 indoor/outdoor scenes.  
To scale up research on hand gestures, \cite{egoGesture_18} introduced the EgoGesture dataset, which comprises 24000 gesture samples (RGB and depth) acquired by 50 different subjects. The dataset contains 83 classes of static and dynamic gestures, designed specifically for interaction with wearable devices. 

To study \textit{hand-object interactions} many datasets of \textit{real} images and videos have been proposed. \cite{Shan_2020_CVPR} collected the 100 Days of Hands (100DOH) Internet scale dataset to enhance size and diversity in hand-object research. It consists of 100K frames acquired over 131 days in which humans were involved in 11 categories of interactions labeled with bounding boxes around the hands and the active object, hand side and hand contact state (indicating if there is a contact between the hand and an object or not). 
\cite{Lu2021UnderstandingEH} introduced a dataset to study hand poses during manipulation with objects. The dataset has been captured in a multi-cam setting using two HD cameras and an iPhone 12. The authors collected 2000 pairs of hand-object interactions performed with a single right hand. GUN-71 is composed of 12K frames annotated with 71 action classes and 28 object classes. 
THU-READ (\cite{GonzlezSosa2021RealTE}) is an egocentric dataset composed of 960 RGB-D videos captured from people performing 40 different daily-life interactions. The data is labeled with pixel-wise annotations of egocentric objects and hands. 

\cite{HOI4D_Liu} presented a large-scale dataset named HOI4D. It is composed of 2.4 million RGBD egocentric video frames acquired in indoor environments where people interact with 800 object instances. It has been labeled with a rich set of 2D and 3D annotations. In particular, hands are annotated with their pose, while objects have labeled segmentation masks, 3D poses and also their CAD models have been released.
The MECCANO dataset by \cite{ragusa2021meccano, ragusa2023meccano} focuses on human-object interactions in an industrial environment where 20 subjects assemble a toy model of a motorbike. It is composed of 20 videos with average duration of 20.79 min and it is multi-modal, comprising synchronised gaze signals, depth maps and RGB videos. 

Unlike other datasets that primarily focused on rigid object manipulation, \cite{fan2023arctic} introduced the ARCTIC dataset, which is specifically designed for interactions involving hands manipulating articulated objects, such as scissors or laptops. This dataset is unique as it includes paired 3D hand and object meshes along with detailed dynamic contact information. 

Only a few works have specifically focused on hand-object interactions with fine-grained information.
\cite{Fine_grained_HO_Zhang} annotated 11K egocentric hand-object interactions with semantic segmentation masks and contact boundaries. These images have been collected from three existing datasets: Ego4D \citep{grauman2022ego4d}, EPIC-KITCHENS \citep{damen2018scaling} and THU-READ \citep{GonzlezSosa2021RealTE}. 
\cite{VISOR2022} extended EPIC-KITCHENS-100 dataset with pixel-level annotations, obtaining 272K semantic masks interpolated to 9.9M dense masks. As a result, they captured long-term object segmentations of the same instance which is subject to a series of transformations during hand-object interactions.
Recently, \cite{tokmakov2023breaking} presented VOST, a dataset composed of 713 videos in which 51 different object transformations (i.e., objects which dramatically change their appearance), have been annotated with segmentation masks. Videos have been annotated at 5 fps obtaining 76K annotated frames.

Recent works investigated the use of egocentric \textit{synthetic} data to mitigate the need of annotating real domain-specific data for model training in \textit{hand-object interaction}.
Real data with ground truth labels are difficult to obtain. Acquiring egocentric images/videos of hand-object interactions as well as manually annotating hands and objects with keypoints, 2D and 3D bounding boxes, semantic masks, relations and action descriptions are a time consuming and expensive task.
\cite{Rogez20143DHP} is a pioneer work, in which synthetic images have been generated to demonstrate their potential for the 3D hand pose detection task.
They focused on hand pose estimation while humans perform object manipulation, proposing a photorealistic synthetic model of egocentric scenes to generate training data for learning depth-based pose classifiers. 
\cite{Hasson2019LearningJR} proposed ObMan, a large scale synthetic dataset made of images of hands grasping objects. By randomising the background and selecting images from the LSUN \citep{yu2015lsun} and ImageNet \citep{russakovsky2015imagenet} datasets, they successfully generated 20K diverse hand-object interactions.

In the pursuit of synthetic data generation and annotation, various studies have directed their attention towards creating photorealistic datasets and tackling the domain-shift problem that emerges when transitioning between real and synthetic domains. 
\cite{leonardi2023exploiting} presented a comprehensive pipeline and framework for automatically generating egocentric hand-object interactions, including  annotations such as depth maps, semantic segmentation masks, bounding boxes for objects and hands, as well as attributes and their respective 3D distances. 
In the work of \cite{ye2023affordance}, diffusion models have been used to generate complex hand-object interactions, allowing reasoning about \textit{where} to interact and \textit{how} to interact. 
\cite{Tendulkar2022FLEXFG} proposed FLEX, a framework capable of generating full-body and hands grasping poses for everyday objects. FLEX is able to synthesise a wide range of natural grasping poses, ensuring diversity and generalisation, while considering 3D geometrical constraints in complex scenes.  Recently, \cite{xu2023visualtactile} utilised tactile sensing for in-hand object reconstruction. Given the difficulty of obtaining ground truth data for object deformation, they addressed this challenge by synthesising images using the proposed simulator. 
Despite the significant progress towards narrowing the gap between synthetic and real domains, the problem has not been solved and there is space for future investigations.

\paragraph{For the future}
Even though there has been advancement across various research areas related to hand analysis, current approaches still come with their own set of limitations. 
State-of-the-art works in \textit{hand pose} focus on the analysis of the posture considering different signals \citep{rudnev2021eventhands, Yang_2022_WACV} or challenging scenarios in which both hands interact simultaneously \citep{Lee_2023_CVPR}. At this stage, methods are able to predict hand pose in a large variety of domains thanks to the availability of large datasets acquired and labelled explicitly for these domains. However, they typically fail in predicting the hands pose when hands are involved in the interaction with objects and in complex scenarios. The state-of-the-art approach for 3D hand pose estimation \citep{Ohkawa_2023_CVPR} achieves a MPJPE of 23.46 mm on the test set of AssemblyHands dataset. Although these results are promising, there is still room for further research and advancements in this field.

\textit{Gestures} have been studied to allow humans to interact with devices such as AR/VR glasses \citep{Bai_gestures_depth_2018} or robots \citep{Papanagiotou_Frontiers_gestures}. The current state-of-the-art performance on the gesture recognition task \citep{Chalasani_gestures_AR_2018} achieves a high accuracy of 96.9\% on the test set of the EgoGesture dataset. These results confirm that the performance of gesture recognition models is comparable to that of humans when considering a discrete number of gestures. However, there is still room for exploration when larger numbers of gestures are considered.
Nowadays, AR/VR devices have a gesture recognition system able to recognise simple gestures like \textit{push, pinch, point, air tap} useful to interact with the system and with virtual objects placed in the environment (i.e., holograms). Furthermore, devices such as Xreal, implemented custom gestures to interact with the device such as \textit{victory} or \textit{open hand} as well as HoloLens2 and Apple Vision Pro allow to use gestures using both hands and gaze.
However, the total number of gestures that can be recognised is small, even though users can usually implement custom gestures specifically for their devices.

For \textit{hand-object interaction}, specific sets of interactions in constrained environments, such as kitchens \citep{VISOR2022} and industrial workplaces \citep{ragusa2023meccano, Synthetic_EHOI_Leonardi}, are starting to be analysed. Despite a few initial efforts to develop approaches capable of understanding generic interactions (e.g. \cite{Shan_2020_CVPR}), we are still far from having robust methods that can generalise across objects and environments, for example industrial environments where hands are in contact with both large machines as well as small tools and objects (e.g. screws). 
One of the main problems lies in the availability of datasets explicitly labelled with human-object interactions, as it requires a significant amount of effort to acquire and manually label such data. A promising direction is to leverage automatically labelled synthetic data, as it enables the acquisition of large-scale hand-object interactions across multiple domains and diverse sets of objects \citep{Tendulkar2022FLEXFG, Synthetic_EHOI_Leonardi}. However, synthetic images often lack the level of photo realism of real-world images. Additionally, dealing with object transformations under manipulations is a very challenging task that has stated to be studied only recently \citep{VISOR2022,fan2023arctic}. 

Actually, the state-of-the-art performance on the Hand-Object Interaction Segmentation task \citep{VISOR2022} achieves a Hand Mask Average Precision of 95.6\% and an Active Object Average Precision of 25.7\% on the test set of the EPIC-KITCHENS VISOR dataset. On the other hand, performance on the object state change task on the test set of Ego4D \citep{grauman2022ego4d} reached an accuracy of 67.6\% and an Average Precision of 15.5\%. These results highlight the difficulty in bridging the gap between action perception and the relations between actions, objects, and the environment.
\subsection{Person Identification}
\label{sec:personid}
Person identification is very relevant for surveillance and security applications and has been extensively studied in third-person literature while it has been less investigated from the egocentric point of view.
While person identification from first person cameras can leverage some algorithms already investigated from remote cameras.
Particularly, in egocentric vision, person identification includes two distinct sub-tasks: recognising people in the field of view of the camera, and identifying the camera wearer. Both rely on the definition of a robust representation of faces as well as other body parts and their movement.
The goal of the former is recognising whether two images depict the same person, or searching for a person within a gallery given a reference image query. More precisely, in face recognition the focus is only on faces and 
the gallery contains a pre-defined set of identities. Person re-identification considers instead the whole body, and the gallery contains many distractors without specific identities. 
Different from remote cameras, the observed individual is often truncated or highly occluded due to the camera being at human head height, compared to remote cameras which are elevated, decreasing the amount of occlusions.
The performance is assessed either in terms of accuracy by considering image pairs and their prediction (each pair with positive/negative label), or by counting how many of the true match appears at the top of the ranked gallery. 

The identification of the person who wears the camera can be formalised as classification in a closed-set scenario, or matching in an open-set one. It involves the wearer's hands and gait, with applications in theft prevention and personalisation.

\paragraph{Seminal works}
Initial efforts to address person identification on wearable devices primarily focused on facial recognition. For instance, \cite{farringdon2000visual} were pioneers in developing a wearable application that automatically identified and stored faces to enhance the camera wearer's memory. This approach was further expanded to various ending goals. 
\cite{krishna2005wearable} developed iCare, an interaction assistant for the visually impaired. It recognises individuals in the scene and notified the wearer through audio signals. \cite{wang2013computerized} targeted prosopagnostics patients, i.e. people who cannot distinguish faces, proposing a system that displayed the identity of people in the scene directly on a screen mounted on the wearable device. 
\cite{10.1145/2493432.2493509} explored the use of face detection, image cropping, location- and motion-based filtering to remove privacy-sensitive information from collections of egocentric images while still allowing to carry out the downstream task of eating behaviour recognition. 
Additionally, \cite{sajjad2020raspberry} proposed a system for enhanced law enforcement that collects data from wearable devices to identify suspects or missing individuals.

Some works proposed to explore external sources of knowledge for specific objectives. 
\cite{kurze2011smart} coupled the face recognition system with data from social networks in order to automatically link the person retrieved with the corresponding online information. 
\cite{reid-ego_chakraborty_WACV2016} considered a face recognition model running on a network of wearable cameras, extending the system developed on Google Glasses  \citep{mandal2015wearable} for face re-identification.
\cite{reid-ego_fergnani_CVPRW2016} proposed a method for full-body person re-identification: a metric learning approach that evaluates instance similarity by dividing images into meaningful body parts and considering a part-related weight defined from human-gaze information.
\cite{basaran2018egoreid} exploited additional metadata collected by mobile phones to reduce the search space. Their approach predicts the next moving camera where the target may appear and aggregates temporal information within sequences of body parts to re-identify individuals.

Three cross-view works are at the interface between recognising the camera wearer and recognising bystanders. 
\cite{7299183} considered a scenario where multiple people are wearing a camera and recording each other. In such cases, the work proposes to use motion correlation between the target person’s video and the observer’s video to uniquely identify instances of oneself, which can be useful for privacy filtering. 
\cite{10.1007/978-3-319-16811-1_21} leveraged head motion patterns to identify the camera wearer in other videos recorded simultaneously, both from third-person and egocentric perspectives. 
The head-motion signature allows the wearers to recognise themselves in videos and decide whether to keep or delete them. 
\cite{fan2017identifying} proposed to use both a third-person camera capturing the scene and multiple subjects recording from an egocentric view. The goal is to match people across different views and identify the source of the egocentric video in the third-person one. 

The identity of the camera wearer from the single egocentric perspective is challenging to discover due to the limited field of view often obscuring the wearer's body. \cite{shiraga2012gait} attempted to recover this information using a complex system of stereo cameras on the user's backpack to analyse motion during walking. On the other hand, \cite{7780833} proposed an approach based solely on a front camera and the head motion signature present in the egocentric video. This approach tries to match the identity of the camera wearer across different videos, with potential applications in theft prevention.
In a different setting, \cite{ardeshir2016ego2top} utilised both egocentric and top-view camera recordings to match each camera wearer with their corresponding top-view identity. Furthermore, \cite{10.1145/3394171.3413654} demonstrated that even using hand motion, captured in the form of dense optical flow, can reveal the identity of the camera wearer across various activities and subjects.

\paragraph{State-of-the-art papers}
For person re-identification, \cite{reid-ego_choudhary_CVIP2021} recently proposed to leverage third-person large-scale datasets in the egocentric domain.  
They employed Neural Style Transfer (NST) to generate high-quality images with a fixed camera style from the egocentric ones. A  content loss ensures that the architecture maintains coherent predictions despite style differences, and a style loss is adopted to improve the transfer capabilities of the model. 
The most recent work combining top-view and egocentric videos is the one presented by \cite{reid-ego_ardeshir_ECCV2018}. It proposes a learning method that matches corresponding pairs of bounding boxes from egocentric and top-view surveillance videos, combining also some geometrical and spatiotemporal reasoning. The former evaluates  the probability of each identity being present in the field of view of the camera holder on the basis of the output of a multiple object tracking algorithm. The latter defines a cost for assigning the same identity label to a pair of bounding boxes depending on whether they are present in the same frame and if they overlap in temporally nearby frames.  This approach achieves state-of-the-art performance in both self-identification and re-identification.

In the context of identifying the wearer from the egocentric perspective, \cite{10.1007/978-3-030-58520-4_24} introduced EgoGaitNet, a model capable of extracting the wearer's gait from the optical flow of an egocentric video, enabling the matching of videos from the same wearer. Additionally, they propose a Hybrid Symmetrical Siamese Network that can match third-person views of a subject with their egocentric videos recorded at different times, raising important privacy concerns.
\cite{Tsutsui2020WhoseHI} explore various sources of information related to the wearer's hands to investigate the feasibility of wearer identification across different videos. They experiment with both RGB modality containing the texture of the hands and the depth modality providing information about their shape. Furthermore, they extract the silhouette of the hands from the depth information and get a multi-modal robust representation for their experiments.

\paragraph{Datasets}
EgoSurf, is a small-scale dataset introduced by \cite{7299183}. It consists of egocentric videos recorded by individuals engaged in face-to-face communications,   captured in eight different scenes (four indoors and four outdoors) by two or three people. Its purpose is to match the camera wearer of a video with their third-person view from another person's egocentric recording.
The Ego2Top dataset was introduced by \cite{ardeshir2016ego2top} and allows both self-identification and re-identification. It comprises 50 top-view and 188 egocentric videos, amounting to approximately 225 thousand frames. However, the number of distinct identities in this dataset is relatively limited. 

The Egocentric Video Photographer Recognition (EVPR) dataset by \cite{7780833} includes videos featuring $32$ subjects: it was created with two distinct types of cameras and primarily used for egocentric camera wearer identification. 
\cite{10.1007/978-3-030-58520-4_24} additionally created the IITMD-WFP and IITMD-WTP datasets to investigate potential biometric signature leakage in egocentric videos. The IITMD-WFP dataset encompasses $3$ hours of videos recorded by $31$ different subjects, while the IITMD-WTP dataset serves as the third-person counterpart.

\paragraph{For the future}
Third-person face recognition and person re-identification research has recently took great advantage of the development of transformer modules able to produce robust feature representations \citep{reid_Liao_NeurIPS2021,reid_zhang_CVPR2023}, and of self-supervised pretraining \citep{reid_fu_CVPR2021,reid_zhu_ECCV2022,reid_fu_CVPR2022}. 
However egocentric person identification is still lagging behind. 
The mean Average Precision (mAP) of state of the art approaches \citep{wieczorek2021unreasonable} over fixed camera benchmarks \citep{zheng2015scalable} is close to 98\% while egocentric models \citep{reid-ego_choudhary_CVIP2021} barely reach 65\% on small scaled egocentric datasets. In the identification of the camera wearer, the low mAP results by \cite{fan2017identifying} indicate that current research is still in the proof of concept phase. In particular, future developments should focus on producing larger and more diverse egocentric person Re-ID benchmarks in order to train reliable models directly from egocentric data.
\subsection{Summarisation}
\label{sec:video_summarisation}
The increasing prevalence of wearable cameras has led to a proliferation of long and unstructured video recordings documenting people's lives. However, users may not revisit much of this recorded content, and important events can be hidden among repetitive or uninteresting segments. 
Video summarisation is a valuable task that aims to produce a concise summary of the input recording. Current methods produced summaries in different forms. Keyframe-based summaries involve the selection of a sequence of relevant frames to represent the most critical events or information in the video. Video skimming approaches segment the video into relevant portions and then collect them to produce a reduced version of the initial recording. Finally, fast forwarding techniques prioritise significant sections while reducing the reproduction speed of less important segments without necessarily trimming any section. The most commonly used metric for evaluating video summarisation is the F1-score.

\paragraph{Seminal works}
Traditionally, determining the most pertinent segments of a video for summarisation was achieved through heuristic criteria such as visual saliency \citep{itti1998model}, motion \citep{wolf1996key}, or high-dimensional curve simplification \citep{dementhon1998video}.   

Egocentric video summarisation was first introduced in \cite{aizawa2001summarizing}, which starts by segmenting the long egocentric recording based on visual motion features and then adopts brainwave signals to determine the subjective interest of the camera wearer throughout the subshots. This innovative approach, which optimises the summarisation based on the viewer's brain response has not been explored further.

In~\cite{lee2012discovering}, the approach relied on RGB data along with higher-level features to produce object-driven storyboards across multiple environments. In contrast, \cite{lu2013story} generated the summary as a coherent set of subshots in a story-like manner. The use of web images as a prior to skip uninformative views of objects caused by the motion of a hand-held camera was proposed in \cite{khosla2013large}. \cite{zhao2014quasi} were the first to consider the online aspect of summarisation, allowing the processing of arbitrarily long videos in real-time. 

Concurrently, there has been a focus on developing fast-forward video summarization techniques, primarily aimed at generating hyperlapse videos.
The intent here is twofold - to stabilise the captured footage and simultaneously highlight the most salient sections. Specifically, the method presented by \cite{kopf2014first} achieves this by rebuilding the three-dimensional spatial geometries within the environment and then sampling a virtual camera path from which the output video is reconstructed. 
\cite{okamoto2014summarization} advance this approach by incorporating semantic considerations into the video content. This method prioritises certain segments of the footage, such as crosswalks, which bear a greater significance in navigational guidance videos.

Additional input modalities were used for personalising summarisation, such as feeding textual concepts of interest in \cite{sharghi2016query}, adopting the gaze modality to understand the wearer’s attention in \cite{xu2015gaze}, or using sound in the form of psychoacoustic metrics to discard moments with unpleasant noises in \cite{bajcsy2018fast}. Aesthetic aspects of keyframes were explored in \cite{xiong2014detecting}, which used web photos as a prior to identify video frames that resemble intentional snapshots, defined as snap points, and demonstrated improved performance in downstream summarisation tasks. \cite{bettadapura2016discovering} used quality measures together with GPS data to extract picturesque highlights from large amounts of egocentric video. Social networks were used by  \cite{ramos2020personalizing} to mine topics of interest and fast-forward the video in a semantically-coherent manner.

Egocentric touristic recordings are the main target for  video summarisation. \cite{xiong2015storyline} experimented on egocentric sequences collected at Disneyland and proposed a storyline representation with actors, events, locations, and objects, allowing story-based queries across different tracks. Similarly, \cite{varini2017personalized} focused on providing a touristic summary more dependent on user preferences. This was achieved by adopting metrics based on the wearer's attention, semantic coherence with preferences, and a narrativity grade to effectively extract sub-shots of interest.

Finally, with the advent of deep learning, video summarisation began to adopt two-stream CNN models that consider motion and appearance as two complementary aspects in understanding the highlight score of a video segment~\citep{yao2016highlight}. 
LSTMs were also used by \cite{zhang2016video} in a supervised video summarisation setting. By modelling variable-range dependencies among frames, the approach captures some high-level temporal understanding which is necessary to avoid relying solely on visual cues.

\paragraph{State-of-the-art papers}
The general video summarisation literature mainly trains and/or evaluates using two benchmarks: SumMe \citep{gygli2014creating} or TVSum \citep{song2015tvsum}, where respectively just 4 out of 25 and 5 out of 50 videos are taken from an egocentric perspective. The current state-of-the-art in this setting is achieved by \cite{he2023align} using a multimodal summarisation method that adopts a transformer-based architecture and an alignment-guided self-attention module to exploit the time correspondence between video and text modalities, and inter- and intra-sample contrastive losses.

However, egocentric videos pose unique challenges due to the significant head motion and long ordinary portions, with the camera wearer moving through a variety of scenes in order to perform daily activities. Consequently, the noted state-of-the-art above is not particularly designed to handle these challenges, though direct evaluation of this model on egocentric benchmarks has not been carried out. 

The latest approach for personalised egocentric video summarisation is proposed in \cite{nagar2021generating}. The approach is customisable to adjust both the length and summary content and uses a reinforcement learning~(RL) approach on top of C3D features. The RL action involves either selecting or discarding the sub-shot, and the approach uses basic rewards (distinctiveness, indicativeness, and overall length) as well as customisable ones (social interaction, face identity, and customised target length). While developed for egocentric videos, the approach uses outdated features and using more recent architectures and/or features is yet to be assessed.

Recently, a novel query-focused approach was introduced by \cite{wu2022intentvizor} to provide an interactive method for video summarisation. The authors developed a framework called IntentVizor, which enables the formulation of generic multi-modal queries and facilitates interactive editing of video summaries. Although the authors experimented just with textual and image queries, its underlying principle is centered around the notion of intent, referring to the high-level requirements of the user, irrespective of the modality of the input query. The intent is determined based on a learned distribution of users' needs, which takes into account various query inputs. Furthermore, the proposed Granularity Scalable Ego-Graph Convolutional Network (GSE-GCN) establishes correlations between the video features and the generic intent, thereby facilitating the extraction of noteworthy sections within the video.

Finally, \cite{elfeki2022multi} is the first work on egocentric multi-stream summarisation, which summarises the videos of multiple wearable cameras intermittently sharing the field of view.
The authors proposed a multi-view extension of the Determinantal Point Process, processing all the camera recordings in parallel and selecting inter-stream diverse events and the best ego-camera viewpoint for each event.

\paragraph{Datasets}
The progress in egocentric video summarisation is significantly hindered by the fact that many datasets used in previous studies have not been publicly released and remain confined to specific research projects. As a result, despite having just roughly 15\% of egocentric videos, SumMe \citep{gygli2014creating} and TVSum \citep{song2015tvsum} are the most commonly used benchmarks. These benchmarks provide frame-level interestingness scores, enabling automatic evaluation of summarisation results without the need for user studies. 

The FPVSum dataset by \cite{ho2018summarizing} is a more recent contribution which tries to mitigate this issue. Indeed, it is composed just of egocentric videos, but only partially annotated. The authors also incorporated unlabelled egocentric data to develop a summarisation model capable of better generalisation to the egocentric domain in a semi-supervised manner.

In fast-forward summarisation methods, the Dataset of Multimodal Semantic Egocentric Videos \citep{silva2018weighted} stands out as an extensive dataset. Spanning 80 hours, this dataset provides valuable information about the activities being performed, the attention of the recorder, and the presence of interactions. By incorporating the recorder's interest scores across a wide range of object categories, it gives the possibility to evaluate both the smoothness and the semantic highlights of the summary.

\paragraph{For the future}
Despite \cite{nagar2021generating} having developed an approach capable of handling day-long recordings using sliding windows, thus avoiding the need to feed the entire video as input to the model, this task still remains far from being considered solved. In fact, on the TVSum and SumMe datasets, the current state-of-the-art \citep{he2023align} achieves F1-scores of 63.4 and 55.0 respectively, while even the more intriguing query-focused egocentric summarisation presented in \cite{wu2022intentvizor} only manages to achieve a F1-score of 50.9 on the egocentric query-based dataset presented in \cite{sharghi2017query}. This limited performance is due to the restricted number of test query-concepts, which still falls short of reflecting the real-life scenarios depicted in the envisioned \egox\ scenarios. 

\subsection{Dialogue}
\label{sec:dialogue}
Fostering the integration of vision and language has great potential in advancing human-machine interaction, allowing artificial agents to dialogue as they can \emph{see} and \emph{communicate} 
in a natural way. To this end, the research community has been working mainly in two directions: visual question answering (Section \ref{subsec:vqa}), and more general ego-language models (Section \ref{subsec:dialogue}).

\subsubsection{Visual Question Answering (VQA)}
\label{subsec:vqa}
Visual Question Answering (VQA) consists of developing systems able to answer questions related to the semantic content of images and videos. Thus, the system takes a visual and a language input and produces a language output, or a referral to a part of the image or video. This is a crucial task for the development of \egox able to support users and skills training.
VQA is also considered an effective way to investigate the reasoning capabilities of deep models as the questions can be designed to obtain not only descriptive answers but also complex predictive 
and explanatory outputs.

The adopted settings for VQA are mainly two: multiple-choice VQA and open-ended VQA. The first one is formalised as a classification problem and evaluated on the basis of prediction accuracy. 
The latter is more challenging and realistic as it requires either identifying the correct answers in a large pool of candidates or generating a free-form response. 
In these cases, model assessment is performed by measuring how often the ground-truth answer is selected in the top predicted choices (recall@k) or via language metrics (e.g. ROUGE) and user studies. 

\paragraph{Seminal works} 
\cite{VizWiz_Gurari0SGLGLB18} were the first to highlight the need for egocentric views in real VQA applications to support blind people. The task poses interesting peculiar challenges as some of the images do not contain enough information, so the associated question is not answerable. 
When dealing with videos, the perspective of VQA has initially shifted from third- to first-person to support the training of navigational agents in indoor environments, solving a task originally referred to as embodied question answering  \citep{embodiedqa,yu2019mteqa}. 
After having observed a sequence of egocentric visual frames of a synthetic scene, the agent is interrogated on the position of single or multiple target objects, or asked to plan a series of actions conditioned on the questions \citep{gordon2018iqa}.
\cite{parick-VQAphotorealistic3D} and \cite{VQArealenvironments} studied the same task in photo-realistic scenes obtained by involving point cloud perception and 3D simulators. 

In all the referred works, each VQA episode is considered in isolation, with no memory or information persistence among the episodes. To address this limitation, \cite{Gao_2021_ICCV} proposed a method that decomposes long videos into separate events and exploits multi-step temporal attention. 
\cite{Fan_2019_ICCV} moved the focus from synthetic to real-world scenarios with a human egocentric view, aiming for a better understanding of the footage of wearable cameras. 
This work discussed the shortages of third-person VQA methods when applied in the egocentric setting and pointed out the need for simultaneous estimation of ego-motion and third-person motion, while disentangling attention for first and third-person activities to find out relevant visual content.

\paragraph{State-of-the-art papers}
In the VQA literature for visually impaired people, \cite{Chen_2022_CVPR} have investigated how to ground the answers by segmenting the relevant image region. The work by \cite{Dancette_2023_CVPR} focused instead on how to avoid answering when the visual information is not sufficient.  They proposed to train multimodal selection functions that indicate for which samples the model can be generalised, and which samples are too hard and should be abstained on. 

Among the embodied question answering works with egocentric videos recorded by robotic agents, \cite{zhu23cvpr} recently proposed a new reinforcement learning framework involving multiple phases of environment exploration and reasoning.
\cite{ma2022sqa3d} discussed the challenges of situated VQA in 3D scenes. 
Several publications have also presented novel solutions to answer questions related to long real-world videos. In particular, the approach proposed by \cite{Gao_2023_CVPR} decomposes traditional dense spatial-temporal self-attention into cascaded segment and region selection modules that adaptively select frames and image regions closely relevant to the question. 

Another sub-part of the state-of-the-art egocentric VQA literature targets episodic memory to search for the temporal window that shows a frame relevant to the question, complemented by informative language answers. 
Starting from the Ego4D challenge on \emph{Episodic Memory - Natural Language Query}, \cite{Barmann_2022_CVPR} defined the QAEGO4D dataset with textual answers from human annotators. The authors have also benchmarked several baseline methods on the newly introduced testbed, using a temporal sequence of feature vectors rather than on raw video data to limit the memory and computational burden. \cite{Datta_2022_CVPR} studied the same  problem from a stream of RGB-D images and proposed a method that combines the semantic features extracted from egocentric observations into a single top-down feature map of the scene. This helps to create a consolidated spatiotemporal memory which is provided as input to an encoder-decoder architecture that grounds answers to questions.

Ego4D \citep{grauman2022ego4d} proposes the episodic memory challenges towards querying long-term videos with natural language questions. As an example, \cite{ramakrishnan2023naq} adopted also Ego4D narrations to overcome the scarcity of query-response pairs. 

The most recent research trend is on goal-oriented VQA to improve reasoning models and get a deeper task understanding from egocentric videos. \cite{jia2022egotaskqa} introduced the EgoTaskQA benchmark with questions designed to investigate actions that imply world state transitions, agents' intents in task execution, and their belief about others in collaboration. The same work presented an extensive analysis of several VQA methods highlighting the effective support provided by large language models. 
\cite{assistQ} defined the affordance-centric VQA problem where the AI assistant should learn from instructional videos to provide step-by-step help in the user's view. The authors introduced a new dataset and developed a novel question-to-action model based on an encoder-decoder architecture. More precisely, the encoder is composed of multiple modules that extract features from video, script, question, and answers. The decoder performs cross-attention among the information obtained from the different modalities and produces operational localised answers (text and bounding boxes) in multiple ordered steps. 

\paragraph{Datasets} The research on Egocentric VQA is still in its infancy and rapidly evolving with many publications proposing specific subtasks and dedicated datasets. Indeed most of the references mentioned above came with a novel data collection. 

The VizWiz dataset introduced by \cite{VizWiz_Gurari0SGLGLB18} consists of over 31K visual questions originating from blind people who took pictures using a mobile phone and recorded a spoken question about it, together with 10 crowdsourced answers per visual question. This collection was also recently exploited for an international VQA challenge both on answerability evaluation and answer prediction \citep{vizwiz_challenge}.

The Env-QA dataset by \cite{Gao_2021_ICCV} was the first collection of egocentric videos covering several events, designed for the  analysis of the whole trajectory of state changes. The events include interactions with the environment (e.g. move the pot, turn on the faucet), thus more skills beyond an understanding of the scene composition are needed to solve the task. 
It contains 23K egocentric videos with an average length of 20 seconds, along with 85K questions querying object number, attributes and states as well as events number and their temporal order. The annotations make the dataset suitable for free-form open-ended questions.
\cite{Fan_2019_ICCV} introduced the EgoVQA dataset with questions related to actions (of the camera wearer and third persons), interactions and relative positions, counting, and colours. 
More precisely, it contains 600 question-answer pairs with visual content across 5K frames from 16 first-person videos,  with each video clip lasting from 20 to 100 seconds. This dataset has been mainly used for multiple-choice QA with (five-way classification).
\cite{Barmann_2022_CVPR} defined the QAEGO4D dataset that contains 1325 egocentric videos, each of 8 minutes on average, and 4837 unique answers. 
The authors provided both target moment annotations as well as answer confidence estimations, which can both serve as an additional source of (weak) supervision. The task on this data is cast as open-ended generative QA.

\cite{Datta_2022_CVPR} introduced the Episodic Memory EMQA dataset built by exploiting a 3D simulator to create egocentric RGB-D maps covering indoor paths. 
It contains 9.7K spatial and spatio-temporal localization questions about 12 object categories and the ground truth is provided as a binary segmentation map (``answer'' vs ``background'' pixels). 
The EgoTaskQA benchmark proposed by \cite{jia2022egotaskqa} contains 40K questions balanced with a 1:2 ratio of binary and open-answer. They were procedurally generated within four types of questions (descriptive, predictive, explanatory and counterfactual) to systematically test models' capabilities over spatial, temporal and causal domains of goal-oriented task understanding. The corresponding videos are reasonably long with an average of five actions per clip to cover sufficient information for action dependency inference and future prediction. 
The QA task is formulated as a classification problem over the whole answer vocabulary.
The AQTC benchmark proposed by \cite{assistQ} was created with a  close focus on task completion and affordances. It contains 100 instructional videos with an average duration of 115 seconds and involves 25 common household appliances, with 531 multiple-choice question-answer samples. The task associated with the dataset is particularly challenging as most of the answers require a sequence of more than two multi-modal steps to guide the user in operating the observed device.

\paragraph{For the future}
Egocentric VQA is a key enabler for a wide range of assistive applications in daily life and in working environments, but the state-of-the-art is in the early proof-of-concept phase and several challenges still need to be tackled.
Starting from the data, all the existing egocentric VQA testbeds focus on indoor scenes which limits the model's applicability. Moving to outdoor environments implies managing a shift in the video features and in the questions' semantics. 
Multi-modality is one crucial aspect of the task, but questions and answers are interpreted as text while speech, and more in general sound, are important cues that are currently less investigated. 

Regarding the emergence of powerful vision-language models, their potential for application in egocentric VQA has only begun to be explored.

Recently LLaVA~\citep{liu2023llava} extended VQA to in-the-wild conditions where the answers require extensive knowledge coverage and multilingual understanding capabilities. The obtained results showed the limitations of existing models in grasping complex semantics.
A relevant aspect to consider is also the length of the required video to answer the question: \cite{mangalam2023egoschema} introduced EgoSchema, a very long-form video question-answering dataset that can serve as a valuable probe to assess the understanding capabilities of modern vision and language systems.
When benchmarking several methods, the authors showed that even models with several billions of parameters achieve QA accuracy of less than 33\% on the EgoSchema multi-choice question answering task, while humans achieve about 76\% accuracy.
\cite{jia2022egotaskqa} indicated that for the most challenging reasoning questions, the performance of large pretrained vision-language models may show a drop and that the development of tailored prompting strategies for those cases is an interesting problem to be solved. 
In this work the best accuracy in predicting the correct answer in the open-ended setting is 30\%, while humans get 82\%.

\subsubsection{Ego-Language Models}
\label{subsec:dialogue}
To engage in a dialogue with \egox, it should be endowed with language abilities that go beyond that of answering visual questions. A general conversation may include descriptions, explanations, and instructions as well as narratives, summaries, and comparisons. 
The most recent research products in this context are Large Language Models (LLMs), trained with a huge amount of textual data and capable of chatting with a user~\citep{touvron2023llama,NEURIPS2020_1457c0d6}.   
The computer vision research community is now focusing on the development of large multi-modal models that integrate vision and language by building on LLMs. In particular, this topic has gained momentum in the egocentric literature. We discuss these advances in this section

\paragraph{Seminal works}
An essential skill to unlock linguistic interactions based on visual information is that of translating semantic content seamlessly between the two modalities: from video to text and from text to video. One approach to achieving that is to learning a representation space shared by the two modalities. Once trained, these embeddings may be fine-tuned for a range of downstream tasks.
Motivated by this, \cite{lin2022egocentric} explored approaches for Video-Language Pretraning (VLP) and proposed a novel video-text contrastive objective, EgoNCE for egocentric videos.
It adjusts the InfoNCE objective~\citep{Oord2018RepresentationLW} by performing action-aware positive sampling and scene-aware negative sampling. This makes NCE specific to long egocentric videos with multiple actions.

In \cite{Suris2022}, the task of relating novel words, outside the learnt vocabulary, to visual objects is explored. The authors created 
an episode of examples consisting of video-text pairs and the model is tasked with completing the masked word from the target example using the reference set from the episode. However, the model is only allowed to fill in the masked word by copying and pasting words from within the episode. This is how the framework learns a policy for word acquisition.

Other techniques have been developed to leverage LLMs without creating new embeddings.
Lavila \citep{zhao2023learning} focuses on automatic video narration and utilises two LLMs: a \textit{narrator} and a \textit{rephraser}. The \textit{narrator} is a visually conditioned auto-regressive language model that provides pseudo labels for existing and new clips with narrations. 
The rephraser, on the other hand, paraphrases the output of the narrator by changing word order or replacing common nouns and verbs. 
The results are reliable and diverse captions, providing temporally synced dense coverage for long videos.

\paragraph{State-of-the-art papers}
For Video-Language Pre-training, \cite{Pramanick_2023_ICCV} proposed EgoVLPv2 that incorporates cross-modal fusion directly into the video and language backbones. The network design keeps the encoders of the two modalities separated but the cross-modal attention modules combine their information and can be reused for downstream tasks with an advantage both in performance and efficiency.
In particular, the EgoVLPv2 pre-trained encoders can be leveraged both for fast retrieval and grounding tasks, which require dual and fusion encoders, respectively.

\paragraph{Datasets}
Two of the largest egocentric videos datasets, EPIC-KITCHENS~\citep{Damen2021rescaling} and Ego4D~\citep{grauman2022ego4d} provide dense free-form text descriptions, also known as narrations. They were collected through a stop-and-narrate approach where the subject watches their video and notes about what is happening there. 
For EPIC-KITCHENS, the authors record audio narrations from the camera wearers themselves, in their native language. 
They then transcribe and translate these timestamped narrations.
For Ego4D, the authors hire annotators who watch the videos and write free-form descriptions of what is happening roughly every four seconds. For each video, two descriptions are collected from different annotators. \cite{lin2022egocentric} utilises Ego4D to curate the EgoCLIP dataset for pre-training models on video-text pairs from egocentric videos. EgoCLIP consists of 3.8M clean egocentric clip-text pairs. For selecting the clips, videos with missing narrations from Ego4D are filtered, validation and test videos are excluded, and narrations from both runs in Ego4D are used to maintain the diversity. As EgoCLIP is derived from Ego4D, the dataset contains diverse human activities. \cite{lin2022egocentric} further propose the EgoMCQ benchmark that aims to evaluate video-text alignment and consists of 39K questions.

\paragraph{For the future}
Due to the inherent complexity posed by egocentric videos like head motion, occlusion, and limited field of view, standard VLP methods based on CLIP \citep{CLIP_icml2021} fall short in generalising well. So future works should improve the adaptation of existing VLMs to egocentric data and might also introduce new tailored tasks to pave the way towards \egox sustaining dialogues with the user.

For instance, \cite{Wang_2023_ICCV} target the development of an interactive AI assistant that can perceive, reason, and collaborate with humans in the real world. The proposed HoloAssist dataset consists of 166 hours of data captured by 222 participants. While capturing the data, there is a \textit{performer} and an \textit{instructor}. The performer works on the task while wearing the AR headset and the instructor watches the performer in real-time and verbally guides the performer. This data collection procedure allows to ground the mistakes and correct the action towards task completion.
\subsection{Privacy}
\label{sec:privacy}
Since the appearance of the first mass consumer wearable cameras in the late 2000s, the research community has been aware of the increased privacy risks related to their use. Such risks are primarily due to the intrinsic mobility of wearable cameras, which allows users to operate them in an ``always on'' mode, thus potentially capturing, transferring and processing sensitive information about themselves and bystanders. Hence, addressing privacy issues in egocentric vision brings specific challenges, as compared to fixed cameras.

While the community has a general understanding of the aforementioned risks, privacy in egocentric vision has not been systematically investigated, which is probably due to the fact that wearable devices equipped with cameras are not yet mainstream technology.
Rather, a range of seminal and exploratory works have been proposed in the last decade.
In this section, we provide a comprehensive discussion of the most relevant investigations on the topic.
In particular, previous research has explored privacy considerations related to wearable cameras through three distinct perspectives: studies aimed to assess the degree to which the use of wearable cameras can affect \textit{individuals’ privacy} (Section \ref{sec:privacy1}), endeavours to \textit{redact sensitive content} captured by wearable devices (Section \ref{sec:privacy2}), and  advancements in \textit{privacy-preserving} computer vision techniques (Section \ref{sec:privacy3}).

\subsubsection{Users' studies on individual privacy}
\label{sec:privacy1}
A line of works has studied how people perceive their privacy and that of bystanders while operating wearable cameras.
In particular, \cite{10.1145/2632048.2632079} performed a user study of $36$ people wearing a life-logging camera for one week, discovering that subjects prefer to be given the option to remove sensitive images in-situ, during the image collection.
Moreover, factors such as time, location, objects and people appearing in a photo determine its sensitivity, and camera wearers are generally concerned about the privacy of bystanders. 
Similar findings are reported in the work of~\cite{price2017logging}, which studied the perception of privacy by different groups of life-loggers. Complementarily, \cite{10.1145/2556288.2557352} investigated the reactions of bystanders to users wearing AR devices.
The study highlights that AR devices change the bystander experience due to subtlety and ease of recording by the camera wearer, with recordings considered more or less acceptable depending on when and where they are being taken.
\cite{10.1145/2702123.2702183} analyses photos taken during life-logging sessions, asking camera wearers to provide motivations on whether a given image should be shared or not: impression management and respect for others’ privacy were the main reasons for keeping images private.

Other works have studied the privacy implications of systems exploiting egocentric images and videos.
\cite{10.1145/2580723.2580730} analysed the security and privacy risks of AR applications in mobile and wearable devices. The study emphasised the necessity of dedicated protocols for wearable applications to ensure the safe usage of sensitive user information such as location data from visual signals or an accurate 3D model of an indoor environment~\citep{Templeman2012PlaceRaiderVT}.
In turn, some works have investigated how egocentric video can contain subtle information which may disclose the identity of the camera wearer, as discussed in Section \ref{sec:personid}. 

\subsubsection{Redacting sensitive information}
\label{sec:privacy2}
Based on the findings of previous studies~\citep{10.1145/2632048.2632079,10.1145/2580723.2580730,10.1145/2702123.2702183}, some works suggested to prevent sharing of egocentric images based on the presence of specific objects or persons. 
For instance, \cite{Templeman2014PlaceAvoiderSF} proposed PlaceAvoider, a system able to recognise whether a given egocentric image has been acquired in a sensitive location specified by the user, such as a bathroom or one’s bedroom, in order to prevent sharing of such images. 
\cite{10.1145/2858036.2858417} proposed to prevent sharing egocentric images based on the presence of detected screens (e.g., the ones of smartphones or laptops) which are likely to include personal information such as credit card numbers or addresses. 
\cite{9152778} investigated approaches to detect bystanders in egocentric images based on cues such as intentionally posing for a photo. 
Images with detected bystanders can then be submitted for review to the camera wearer before proceeding to share them. 

Other works investigated how egocentric images and videos can be transformed to protect privacy, while still allowing for downstream tasks to be performed. 
In particular, \cite{8014909} proposed a “cartoon transform” which alters the low-level properties of the image and replaces objects with aligned clip art. 
\cite{10.1145/3161190} studied how intentionally degrading image quality by blurring can improve bystanders' perception of privacy, while still allowing to perform downstream tasks such as activity recognition.  
Finally, \cite{9710988} proposed to add subtle perturbations to egocentric video that do not affect tasks like object detection or action/activity recognition but are strong enough to prevent the identification of the camera wearer from head motion analysis.

\subsubsection{Privacy preserving by design}
\label{sec:privacy3}
This line of work investigates systems and algorithms to tackle egocentric computer vision tasks while guaranteeing that privacy-sensitive content is correctly processed, by avoiding to collect it, store it, or make it available to untrusted applications.
\cite{Jana2013EnablingFP} introduced the idea of ``recognisers'' as a software layer to provide AR applications with only the necessary high-level, anonymous information, rather than giving direct access to raw sensor data such as RGB images, which may contain private content.
A similar concept is explored in~\cite{6547120}, where a privacy protection layer based on OpenCV is used to mediate sensor input (e.g., applying sketching transform) before making it available to possibly untrusted AR applications.
\cite{Ryoo2016PrivacyPreservingHA} proposed a privacy-conscious approach which learns how to extremely subsample egocentric video resolution to preserve privacy while still allowing to perform the downstream task of activity recognition.
Following~\cite{pittaluga2019revealing}, who showed how scenes can be revealed inverting structure from motion reconstructions, a line of research has investigated approaches for 6-DoF localisation that safeguards against reconstructing the original image from the information stored to support localisation~\citep{8953346,pietrantoni2023segloc,chelani2023privacy,chelani2021privacy,dusmanu2021privacy,ng2022ninjadesc}.
\cite{10.1145/3314111.3319913} presented a system which shuts off the video stream when sensitive visual content is detected, only to reactivate it based on the analysis of eye movements recorded by additional eye tracking cameras. 
Along the same lines, \cite{10.1145/3450614.3461684} proposed a deep learning based device which detects user-customised privacy-sensitive content such as objects and faces of specific people in order to serve as a privacy filter, blocking images which do not satisfy the established privacy constraints.
\cite{10059226} investigated how converting images into rich text descriptions can serve as an effective privacy-preserving approach for passive dietary intake monitoring from egocentric images, as compared to directly storing the input images.

\paragraph{Datasets}
A significant portion of the studies on privacy involves the collection of in-situ data and the administration of surveys to participants~\citep{10.1145/2556288.2557352,10.1145/2632048.2632079,price2017logging,10.1145/2858036.2858417}. These works primarily focus on examining participants' attitudes towards lifelogging through the utilisation of AR glasses. In addition to analysing the gathered survey data, \cite{10.1145/2702123.2702183} also examines the images generated during the data collection process. Another group of studies employs custom hardware setups to capture images in an in-situ context~\citep{10.1145/2493432.2493509,Templeman2014PlaceAvoiderSF,8953346}.

The First Person Social Interaction Dataset (FPSI) by ~\cite{fathi2012social}, described in Section \ref{sec:datasets}, has been used for privacy analysis.
Together with other data collections originally created for person identification (IITMD already covered by Section \ref{sec:personid}), it has been adopted to investigate potential biometric signature leakage in egocentric videos. 

\paragraph{For the future}
The discussed seminal works exemplify the efforts of the community in pursuing privacy-aware technologies. However, the investigations are not yet systematic 
mainly due to the limited adoption of wearable cameras by the general public.

Previous works mainly surveyed individuals but it is important to identify issues with existing datasets, as the covered demographic and number of subjects (<50) are generally limited. 
As devices are rapidly evolving, it will be crucial to evaluate the effect on privacy when moving from mobile phones and GoPros to wearable devices. This appears particularly important considering that the majority of the studies in the current literature used custom devices instead of commercially available ones  \citep{10.1145/2493432.2493509,Templeman2014PlaceAvoiderSF,7299183,8953346}. 

Finally, several works focused on analysing recorded data to assess privacy breaches, but only a few proposed solutions for privacy preservation or privacy-aware processing and more tailored strategies are needed.

\subsection{Beyond individual tasks}
The previous sections revised 12 distinct computer vision research tasks, which are fundamental for the future of egocentric vision. 
Few efforts have attempted to combine these tasks to get close to the abilities of our futuristic \egox{}.
For example, an approach that combines action recognition with a dynamic memory, so as to remind Sam about the bread in the toaster is beyond the reach of current methods.
Similarly, combining person Re-ID with trajectory forecasting towards assisting Judy in locating a suspect along her path has also not been explored before.
Assessing performance in daily tasks, whether to advise Sam about the amount of spice in his soup or qualitatively and quantitatively assess Marco's daily performance are still futuristic tasks not yet explored in egocentric vision.
The understanding of one's surroundings, actions and intentions, both independently as well as jointly with others, is key to the integration of \egox{} in our daily lives.
We hope more works will expand beyond the individual research tasks to get an effective assistive device for the wearer. 

We next review general datasets, suitable for multiple tasks, which can pave the way to such holistic understanding in egocentric vision.
\section{General Datasets}
\label{sec:datasets}
\begin{table*}[ht]
\caption{General Egocentric Datasets - Collection Characteristics. $^\dagger$: For EGTEA, Audio was collected but not made public. $^\star$:~For Ego4D, apart from RGB, the other modalities are present for subsets of the data.}
\label{tab:datasets_comparison1}
\resizebox{\textwidth}{!}{%
\begin{tabular}{llcrrrrr}
\hline
\multicolumn{1}{c}{\textbf{Dataset}} & \textbf{Settings} & \multicolumn{1}{c}{\textbf{Signals}} & \textbf{Hours} & \textbf{Sequences} & \textbf{AVG. video duration} & \textbf{Participants}\\ \hline
MECCANO~\citep{ragusa2023meccano} & Industrial & RGB, depth, gaze & 6.9 & 20 & 20.79 min & 20 \vspace*{6pt}\\ 
ADL~\citep{pirsiavash2012detecting} & Daily activities & RGB & 10.0 & 20 & 30.00 min & 20 \vspace*{6pt}\\ 
HOI4D~\citep{HOI4D_Liu} & Table-Top & RGB, depth & 22.2 & 4000 & 0.33 min & 9 \vspace*{6pt}\\ 
EGTEA Gaze+$^\dagger$~\citep{li2021eye} & Kitchen & RGB, gaze & 27.9 & 86 & 19.53 min & 32 \vspace*{6pt}\\ 
UTE~\citep{lee2012discovering} & Daily Activities & RGB & 37.0 & 10 & 222.00 min & 4 \vspace*{6pt}\\ 
EGO-CH~\citep{ragusa2020ego} & Cultural Sites & RGB & 37.1 & 180 & 12.37 min & 70 \vspace*{6pt}\\
FPSI~\citep{fathi2012social} & Recreational Site & RGB & 42.0 & 8 & 315.00 min & 8 \vspace*{6pt}\\
KrishnaCam~\citep{krishna-wacv2016} & Daily Routine & \begin{tabular}[c]{@{}l@{}}RGB, GPS, acc\end{tabular} & 69.9 & 460 & 9.13 min & 1 \vspace*{6pt}\\ 
EPIC-KITCHENS-100~\citep{Damen2021rescaling} & Kitchens & RGB, audio & 100.0 & 700 & 8.57 min & 37 \vspace*{6pt}\\ 
Assembly101~\citep{sener2022assembly101} & Industrial & RGB, multi-view & 167.0 & 1425 & 7.10 min & 53 \vspace*{6pt}\\
Ego4D$^\star$~\citep{grauman2022ego4d} & Multi Domain & RGB, Audio, 3D, gaze, IMU, multi & 3670.0 & 9650 & 24.11 min & 931 \vspace*{6pt}\\ \hline
\end{tabular}
}
\end{table*}

\begin{table*}[ht]
\caption{General Egocentric Datasets - Current set of annotations. $^\star$: For Ego4D, apart from narrations, the remaining annotations are only available for subsets of the dataset depending on the benchmark}
\label{tab:datasets_comparison2}
\resizebox{\textwidth}{!}{%
\begin{tabular}{ll}
\hline
\multicolumn{1}{c}{\textbf{Dataset}} & \multicolumn{1}{c}{\textbf{Annotations}} \\ \hline

MECCANO~\citep{ragusa2023meccano} & \begin{tabular}[c]{@{}l@{}}Temporal action segments, hand \& object bounding boxes, hand-object interactions, next-active object\end{tabular} \vspace*{7pt}\\

\vspace*{4pt}
ADL~\citep{pirsiavash2012detecting} & \begin{tabular}[c]{@{}l@{}}Temporal action segments, objects bounding boxes, hand-object interactions\end{tabular} \\

\vspace*{4pt}
HOI4D~\citep{HOI4D_Liu} & \begin{tabular}[c]{@{}l@{}}Temporal action segments, 3D hand poses and object poses, panoptic and motion segmentation, object \\meshes, scene point clouds\end{tabular} \\

\vspace*{9pt}
EGTEA Gaze+~\citep{li2021eye} & Temporal action segments, hand masks, gaze\\

\vspace*{9pt}
UTE~\citep{lee2012discovering} & Text descriptions, object segmentations\\

\vspace*{9pt} 
EGO-CH~\citep{ragusa2020ego} & \begin{tabular}[c]{@{}l@{}}Temporal locations, object bounding boxes, surveys, object masks\end{tabular}\\

\vspace*{9pt}
FPSI~\citep{fathi2012social} & Temporal social interaction segments\\ 

\vspace*{6pt}
KrishnaCam~\citep{krishna-wacv2016} & Motion classes, virtual webcams, popular locations \\

\vspace*{5pt}
EPIC-KITCHENS-100~\citep{Damen2021rescaling} & \begin{tabular}[c]{@{}l@{}}Temporal action video segments, Temporal audio segments, narrations, hand and objects masks,\\ hand-object interactions, camera poses\end{tabular}\\

\vspace*{5pt}
Assembly101~\citep{sener2022assembly101} & Temporal action segments, 3D hand poses\\

\vspace*{5pt}
Ego4D$^\star$~\citep{grauman2022ego4d} & \begin{tabular}[c]{@{}l@{}}Narrations, Temporal action segments, moment queries, speaker labels, diarisation, hand bounding boxes,\\ time to contact, active objects bounding boxes, trajectories, next-active objects bounding boxes\end{tabular} \\ \hline

\end{tabular}
}
\end{table*}

\begin{table*}[ht]
\caption{General Egocentric Datasets - Current set of tasks: \textbf{4.1} Localisation, \textbf{4.2} 3D Scene Understanding, \textbf{4.3} Recognition, \textbf{4.4} Anticipation, \textbf{4.5} Gaze Understanding and Prediction, \textbf{4.6} Social Behaviour Understanding, \textbf{4.7} Full-body Pose Estimation, \textbf{4.8} Hand and Hand-Object Interactions, \textbf{4.9} Person Identification, \textbf{4.10} Summarisation, \textbf{4.11} Dialogue, \textbf{4.12} Privacy.}
\label{tab:datasets_comparison3}
\resizebox{\textwidth}{!}{%
\begin{tabular}{lcccccccccccc}
\backslashbox[40mm]{\textbf{Dataset}}{\textbf{Task}}
 & \textbf{4.1} & \textbf{4.2} & \textbf{4.3} & \textbf{4.4} & \textbf{4.5} & \textbf{4.6} & \textbf{4.7} & \textbf{4.8} & \textbf{4.9} & \textbf{4.10} & \textbf{4.11} & \textbf{4.12} \\ \hline
MECCANO~\citep{ragusa2023meccano} &  &  & \checkmark & \checkmark & \checkmark &  &  & \checkmark &  &  &  &  \\
ADL~\citep{pirsiavash2012detecting} &  &  & \checkmark & \checkmark &  &  &  &  &  &  \checkmark & & \\ 
HOI4D~\citep{HOI4D_Liu} &  &  &  &  &  &  &  & \checkmark &  &  &  & \\ 
EGTEA Gaze+~\citep{li2021eye} &  &  & \checkmark & \checkmark & \checkmark &  &  & \checkmark &  &  &  &  \\ 
UTE~\citep{lee2012discovering} &  &  &  &  &  &  &  &\checkmark  &  &  \checkmark & & \\ 
EGO-CH~\citep{ragusa2020ego} & \checkmark &  &  &  &  &  &  &  &  &  &  &  \\ 
FPSI~\citep{fathi2012social} &  &  &  &  &  & \checkmark  &  &  &  & \checkmark & & \checkmark  \\ 
KrishnaCam~\citep{krishna-wacv2016} &  &  & & \checkmark &  &  &  &  &  &  &  &  \\
EPIC-KITCHENS-100~\citep{Damen2021rescaling} &  & \checkmark & \checkmark & \checkmark &  &  &  & \checkmark &  &  & \checkmark & \checkmark\\
Assembly101~\citep{sener2022assembly101} &  &  &  \checkmark & &  &  &  & \checkmark &  &  &  & \\
Ego4D~\citep{grauman2022ego4d} &  &  & \checkmark & \checkmark & \checkmark & \checkmark &  & \checkmark &  &\checkmark  & \checkmark &\\ \hline
\end{tabular}
}
\end{table*}

Datasets have become the fuel of computer vision research. They offer the starting point for studying new research problems and developing artificial intelligence that can successfully support humans.  
The more realistic a dataset is, the higher its value in transforming our future and often the higher its challenge. Such datasets usually require increased research efforts to achieve good performance on their various metrics.

In particular, the availability of these datasets has crucially contributed to advancements in egocentric vision research. In fact, as wearable cameras are still relatively new, the videos available online are not taken from the egocentric perspective. 
In the previous sections, we reported datasets that were designed for one task.
In this section, we review general datasets that are suitable for a variety of tasks, and present their characteristics.
We compare in Table~\ref{tab:datasets_comparison1} the most popular publicly available egocentric datasets in terms of domains, size and modalities. We then detail available annotations to date in Table~\ref{tab:datasets_comparison2} and in Table~\ref{tab:datasets_comparison3} we relate these datasets to the tasks reviewed in Section~\ref{sec:tasks}. 
Next, we provide a narrative for these general datasets.

Activity of Daily Living (ADL) by~\cite{pirsiavash2012detecting} was one of the first egocentric datasets. It consists of one million frames captured in people's homes. The dataset is only scripted at  a high-level by asking the camera wearers to carry out specific tasks such as watching TV or doing laundry. It is annotated with object tracks, hand positions, and interaction events. ADL has found its use in various tasks like action~\citep{vondrick2016anticipating} and region anticipation~\citep{furnari2017next}, action recognition~\citep{pirsiavash2012detecting} and video summarisation~\citep{lu2013story}. 
Similarly, the UTE dataset~\citep{lee2012discovering} is composed by videos of 4 participants involved in various activities such as eating, shopping, attending a lecture, driving and cooking. One notable difference with respect to ADL is in the video length:  
the average duration of a video in UTE is 3.7 hours (222 mins) compared to 30 mins for ADL. 
Regarding the annotations, the UTE dataset provides a paragraph summary of the videos and polygon annotations around the subjects based on the summary, which makes it suitable for studying video summarisation~\citep{lee2012discovering,lu2013story}.
Further expanding the covered time range, 
the KrishnaCam dataset by~\cite{krishna-wacv2016} includes nine months of one student's daily activities. It consists of 7.6 million frames, spanning 70 hours of video, accompanied by GPS position, acceleration, and body orientation data. Thanks to its time evolution capture of nine months, KrishnaCam can be used to study tasks such as trajectory prediction, detecting popular places and scene changes. The dataset has been also used to address online object detection~\citep{Wang2021WanderlustOC} and for self-supervised representation learning~\citep{Purushwalkam2022TheCO}.

Different in terms of domains and captured signals, the GTEA Gaze dataset~\citep{fathi2012learning} and its extension EGTEA Gaze+~\citep{li2021eye} cover recipe preparation with the gaze signal in a single kitchen. The GTEA Gaze dataset by~\cite{fathi2012learning} focuses on action recognition and gaze prediction and involves the use of eye-tracking glasses equipped with an infrared inward-facing gaze sensing camera to track the 2D location of the subjects' eye gaze during meal preparation activities. The dataset includes 17 sequences performed by 14 subjects making pre-specified meal recipes. It has been annotated with 25 frequently occurring actions, such as ``take'', ``pour'', and ``spread'' indicating their starting and ending frames. This dataset was later extended as EGTEA Gaze+ by~\cite{li2021eye} with 28 hours of cooking activities, including video, gaze tracking data, and action annotations of 106 actions, along with pixel-level hand masks. The dataset has been used to address different tasks such as anticipation~\citep{furnari2019would, girdhar2021anticipative, zhong2023anticipative}, action recognition~\citep{kazakos2021MTCN,fathi2012learning}, procedural learning~\citep{EgoProceLECCV2022}, and future hand masks prediction~\citep{jia2022generative}.

A few datasets targeted multi-person egocentric social interactions.
The First Person Social Interaction (FPSI) dataset by~\cite{fathi2012social} was collected over three days, by a group of 8 individuals that visited Disney theme parks, recording over 42 hours of multi-person videos using head-mounted cameras. The group often splits into smaller sub-groups during the day, resulting in unique experiences in each video. The dataset consists of over two million images,  manually labelled for six types of social interactions: dialogue, discussion, monologue, walk dialogue, walk discussion, and background activities.
The dataset has proven useful for video summarisation~\citep{nagar2021generating,10.1145/3343031.3350880,7299109} and privacy preservation~\citep{fathi2012social,10.1007/978-3-030-58520-4_24}.

\cite{ragusa2020ego} proposed the EGO-CH dataset to study 
visits to cultural heritage sites. It includes 27 hours of video recorded from 70 subjects. Annotations are provided for 26 environments and over 200 Points of Interest (POIs), featuring temporal labels indicating the environment in which the visitor is located and the currently observed PoI with bounding box annotations. 
It has been used by the authors to tackle room-based localisation, PoI recognition, image retrieval and survey generation -- i.e. predicting the responses in the survey from the egocentric video. Furthermore, the dataset has been used to address object detection~\citep{pasqualino2022unsupervised,pasqualino2022multi}, image-based localisation~\citep{Orlando2020VirtualTR} and semantic object segmentation~\citep{ragusa_ICPR2020}.

While these datasets explore various aspects of egocentric vision, their small scale and focus on a single environment or a handful of individuals poses challenges when training deep learning models or attempting to generalise to other locations or subjects. To this end, the EPIC-KITCHENS dataset by~\cite{damen2018scaling} was proposed as a significantly larger egocentric video dataset introduced in 2018 and subsequently extended with the latest version EPIC-KITCHENS-100 by~\cite{Damen2021rescaling}. The dataset comprises 100 hours of unscripted video recordings captured by 37 participants from 4 countries in their own kitchens. It is unique in its instructions to participants, so as to start recording before entering the kitchen and only to pause when stepping out. This offered the first unscripted nature where participants go around their environments unhindered forming their own goals.
The dataset has been annotated temporally with action segments.
It consists of 90K action segments, 20K unique narrations, 97 verb classes, and 300 noun classes. 
It has since been extended with three additional annotations.
First, EPIC-KITCHENS Video Object Segmentations and Relations (VISOR)~\citep{VISOR2022} provided pixel-level annotations focusing on hands, objects and hand-object interaction labels.
VISOR offers 272K manual semantic masks of 257 object classes, 9.9M interpolated dense masks, 67K hand-object relations.
Second, EPIC-SOUNDS~\citep{EPICSOUNDS2023} annotates the temporally distinguishable audio segments, purely from the audio stream of videos in EPIC-KITCHENS.
It includes 78.4k categorised segments of audible events and actions, distributed across 44 classes as well as 39.2k non-categorised segments.
Third, EPIC Fields~\citep{EPICFields2023} successfully registered and provided camera poses for 99 out of the 100 hours of EPIC-KITCHENS. 
This is achieved through a proposed pipeline of frame filtering so as to attend to transitions between hotspots.
The camera poses offer the chance for combining all aforementioned annotations with 3D understanding and are likely to unlock new potential on this dataset.

Since its introduction, EPIC-KITCHENS has become the de facto dataset for egocentric action recognition~\citep{kazakos2019TBN,Xiong2022m,yan2022multiview,girdhar2022omnivore}, privacy~\citep{10.1145/3394171.3413654}, and anticipation~\citep{furnari2019would,girdhar2021anticipative,gu2021transaction,roy2022action,liu2020forecasting,jia2022generative,pasca2023summarize,zhong2023anticipative}.
New tasks have also been defined around EPIC-KITCHENS, particularly related to domain adaptations with its capture in multiple locations and over time~\citep{munro2020multi,Kim_2021_ICCV,NEURIPS2021_c47e9374}, video retrieval~\citep{zhao2023learning,lin2022egocentric}, manipulations~\citep{pmlr-v205-shaw23a} as well as niche topics like object-level reasoning~\citep{Baradel_2018_ECCV} and learning words in other languages from visual representations~\citep{Suris2022}.

A couple of datasets focus on industry-like setting. MECCANO by~\cite{ragusa2021meccano,ragusa2023meccano} is an egocentric procedural dataset capturing subjects building a toy motorbike model. 
The dataset includes synchronised gaze, depth and RGB. Its 20 object classes cover components, tools, instructions booklet. It has been used to address tasks like action recognition~\citep{deng2023BEAR}, active object detection~\citep{Fu_2022_CVPR}, hand-object interactions~\citep{Tango2022BackgroundMD} and procedural learning~\citep{EgoProceLECCV2022}.
Similarly, Assembly101~\citep{sener2022assembly101} is a procedural activity dataset with 4321 videos of individuals assembling and disassembling 101 ``take-apart'' toy vehicles. The dataset showcases diverse variations in action orders, mistakes, and corrections. It contains over 100K coarse and 1M fine-grained action segments, along with 18M 3D hand poses. This dataset has found its use in action recognition~\citep{wen2023interactive}, anticipation~\citep{Zatsarynna2023ActionAW} and hand pose estimation~\citep{zheng2023hand,Ohkawa_2023_CVPR}.
Additionally, HOI4D dataset by~\cite{HOI4D_Liu} consists of 2.4M RGB-D video frames and 4000 sequences, featuring 9 participants interacting with 800 object instances from 16 categories. The dataset provides annotations for panoptic and motion segmentation, 3D hand pose, category-level object pose and includes reconstructed object meshes and scene point clouds. This dataset has proven useful for object segmentation and shape reconstruction~\citep{Liu2023SelfSupervisedCA,Zhang_2023_CVPR, Hao_point_eccv22}, action segmentation~\citep{Reza2023EnhancingTB, Zhang_2023_CVPR}, hand-object manipulation synthesis~\citep{Zheng_2023_CVPR,ye2023affordance}, hand action detection~\citep{HungCuong2023YOLOSF} and 3D hand pose estimation~\citep{ye2023affordance}.

The most impressive and massive-scale dataset to date is Ego4D by~\cite{grauman2022ego4d}, with 3,670 hours of daily-life activity videos spanning hundreds of unscripted scenarios (household, outdoor, workplace, leisure, etc.) captured by 931 unique camera wearers from 74 worldwide locations and 9 countries. It primarily comprises videos, with subsets of the dataset containing audio, eye gaze, and 3D meshes of the environment.
The dataset was released with a set of benchmarks and train/val/test split annotations that focus on the past (querying an episodic memory), the present (hand-object manipulation, audio-visual conversation, and social interactions), as well as the future (forecasting activities and trajectories).

Due to the massive-scale and unconstrained nature of Ego4D, it has proved to be useful for various tasks including action recognition~\citep{liu2022egocentric, DeLange2023EgoAdaptAM}, action detection~\citep{Wang2023EgoOnlyEA}, visual question answering~\citep{Barmann_2022_CVPR}, active speaker detection~\citep{Wang2023LoCoNetLC}, natural language localisation~\citep{Liu2023AnchorBasedDF}, natural language queries~\citep{ramakrishnan2023naq}, gaze estimation~\citep{lai2022eye}, persuasion modelling for conversational agents~\citep{lai2023werewolf}, audio visual object localisation~\citep{Huang_2023_CVPR}, hand-object segmentation~\citep{Zhang2022FineGrainedEH} and action anticipation~\citep{ragusa2023stillfast,pasca2023summarize,mascaro2023intention}.
New tasks have also been introduced thanks to the diversity of Ego4D, e.g. modality binding~\citep{Girdhar_2023_CVPR}, part-based segmentation~\citep{Ramanathan_2023_CVPR}, long-term object tracking~\citep{tang2023egotracks}, relational queries~\citep{Yang_2023_CVPR} and action generalisation over scenarios~\citep{plizzari2023can}.
Additionally, due to its unprecedented scale, it has broken grounds in training robot models with a series of publications \citep{pmlr-v205-nair23a,pmlr-v205-radosavovic23a,ma2023vip}, transforming the field of learning from demonstrations. 
The potential for the Ego4D dataset is yet to be fully explored.

As noted at the start of this section, egocentric datasets have a key role in research advancement. 
By reviewing our initial forecast of the future in Section~\ref{sec:stories}, some scenarios received more attention than others in producing large-scale datasets. 
EGO-Home (Section~\ref{sec:ego-home}) is partially overlapping with datasets such as EPIC-KITCHENS-100, Ego4D and EGTEA Gaze+.
However, these datasets mostly focus on the home activities of cooking, cleaning and playing games. 
They do not cover parts related to down time (i.e. rest), or grooming or personal health, mostly due to privacy concerns.
EGO-Worker (Section~\ref{sec:ego-worker}) is related to datasets such as MECCANO, Assembly-101 and HOI4D. 
However, these do not cover the holistic aspects of a worker's daily activities and are yet to explore the critical aspects of safety and feedback.
EGO-Tourist (Section~\ref{sec:ego-tourist}) is related to the EGO-CH dataset of visitors in heritage sites. However, the scale remains very small despite the presence of large-scale touring videos on YouTube that could be utilise for city-wide touring.

EGO-Police (Section~\ref{sec:ego-police}) does not correspond to any publicly available datasets.
Despite the wide usage of chest-mounted cameras within the police forces worldwide, such data is very sensitive, particularly across boarders. Relevant datasets currently are far from being utilised for advancing research in egocentric vision.
Finally, we call for more datasets in egocentric understanding from the entertainment industry, given the huge potential of transforming this domain as noted by the Ego-Designer scenario we presented (Section~\ref{sec:ego-designer}).

\section{Conclusion}
\label{sec:conclusion}
This paper aimed to provide a future-to-present perspective into egocentric vision. 
Looking ahead, we envisage a wearable device that we call \egox{}, holding the potential to redefine our daily lives. 
We showcased its seamless integration into our everyday existence through character-based futuristic scenarios, indoors and outdoors, at work, at home and even during holidays. 

We demonstrated the need for this device to be \textit{multi-sensored} and \textit{multi-tasked}. While our focus is on cameras and visual cues, the future clearly requires it to be multimodal in its capabilities, whether for perceiving the surroundings and understanding what is happening in the observed scene, or interacting with the camera wearer.
At the same time, without the ability to solve multiple fundamental vision tasks it will not be possible to get a competent egocentric assistant. 

Additionally, we believe that further developing generative tasks in egocentric vision will play a pivotal role towards building \egox{}. Consider, for instance, scenarios where Marco could benefit from a device guiding him through his work procedure by illustrating the sequential steps within his environment. Similarly, \egox{} could spark Stanley's creativity by proposing diverse scenographies projected onto his current surroundings. Current egocentric methods employ generative approaches in limited contexts, ranging from predicting future head motion \citep{jia2022generative} to anticipating gaze \citep{zhang2017deep} and modelling hand-object interactions \citep{ye2023affordance}. Only a handful of works explore cross-view third-to-first-person image \citep{liu2020exocentric} and video \citep{liu2021cross} synthesis. One recent work that closely aligns with our use cases is from \cite{yang2023learning}. They introduced a universal video generator that predicts future frames based on both low- and high-level textual action prompts. When run sequentially it can also effectively simulates long-horizon interactions. This quality makes it well-suited for generating a visual representation of work procedures tailored to Marco's needs.

In this paper, we reviewed 12 research tasks in egocentric vision: \textit{localisation}, \textit{3D scene understanding}, \textit{recognition}, \textit{anticipation}, \textit{gaze understanding and prediction}, \textit{social behaviour understanding}, \textit{full-body pose estimation}, \textit{hand and hand-object interactions}, \textit{person identification}, \textit{summarisation}, \textit{dialogue} and \textit{privacy}.
For each task, we presented an overview considering appreciated seminal works that set the research path, 
and we provided an insight into the current state-of-the-art methods, publicly available datasets and directions for future innovations. While the literature builds on previous research based on fixed cameras, each of these tasks present challenges which are unique to egocentric vision, and in particular to the mobile nature of wearable cameras and the need for a user-specific understanding of the scene. On the other hand, egocentric vision brings new opportunities for human-centric applications as discussed at large in this paper. We anticipate that future research will focus on bridging the gap between egocentric approaches and those based on third-person vision in the spirit of convergence towards a unified technology. Towards achieving this goal, the newly introduced Ego-Exo4D dataset recorded using both egocentric and up to 4 exocentric cameras has recently been introduced \citep{grauman2023ego}.

We highlight that these tasks cannot exist independently -- i.e. it is infeasible that we will be learning one deep model per task. This is not only because a model-per-task is extremely inefficient, but because these tasks are co-dependent and the prediction of one task would inform plausible predictions of another. 
We encourage researchers to explore the taxonomy of research tasks in egocentric vision. Moreover, future works should also consider open set settings so that each task is able to manage novelty to avoid relying too much on pre-defined label sets and enhance model trustworthiness.

Regarding efficiency, this survey stopped short of exploring the need for real-time sensing and learning, although it is evident that we need to build models capable of performing all the mentioned tasks in real-time or with very minimal latency. Ideally, future egocentric devices should be always connected online, while respecting all privacy and protection concerns. 
We encourage other researchers to analyse these aspects as without efficient sensing, efficient computing and real-time interactions, the future would remain a dream in fiction novels and sci-fi films. At the same time, without privacy-aware models, sensors and systems, the future would fail to deliver on its users expectations.

We hope this paper offers a stepping stone for researchers to make the future of egocentric vision a reality.
We are seeking input from researchers in the field to strengthen and complete this survey, so it can serve as a useful reference to incoming researchers who wish to explore and contribute to egocentric vision.

\section*{Acknowledgements}
We thank Fritz J. Rustan, Illustrator in 99designs, for the fruitful and close collaboration to produce the \egox{} illustrations~(Fig~\ref{fig:home} - Fig~\ref{fig:artist}). 

We thank Mirco Planamente for early discussions on this survey and initial collection of relevant papers.

Research at the University of Bristol is supported by EPSRC Program Grant Visual AI EP/T028572/1. D. Damen is supported by EPSRC Fellowship UMPIRE EP/T004991/1. 

Research at University of Catania has been supported by the project Future Artificial Intelligence Research (FAIR) – PNRR MUR Cod. PE0000013 - CUP: E63C22001940006.

T. Tommasi is supported by the project FAIR - Future Artificial Intelligence Research and received funding from the European Union Next-GenerationEU (PIANO NAZIONALE DI RIPRESA E RESILIENZA (PNRR) – MISSIONE 4 COMPONENTE 2, INVESTIMENTO 1.3 – D.D. 1555 11/10/2022, PE00000013).
C. Plizzari and G. Goletto acknowledge travel support from ELISE (GA no 951847).
G. Goletto is supported by PON “Ricerca e Innovazione” 2014-2020 – DM 1061/2021 funds.
\bibliography{sn-bibliography}
\end{document}